\documentclass[10pt]{article} %
\usepackage[accepted]{tmlr}

\usepackage{amsmath,amsfonts,bm}

\def\eqref#1{equation~\ref{#1}}

\def\1{\bm{1}}

\DeclareMathAlphabet{\mathsfit}{\encodingdefault}{\sfdefault}{m}{sl}
\SetMathAlphabet{\mathsfit}{bold}{\encodingdefault}{\sfdefault}{bx}{n}

\usepackage{hyperref}
\usepackage{url}
\usepackage{amsmath}
\usepackage{comment}
\usepackage{graphicx}
\usepackage{bm}
\usepackage{todonotes}
\usepackage{booktabs}

\title{Why is constrained neural language generation particularly challenging?}

\author{\name Cristina G\^arbacea \email garbacea@uchicago.edu \\
      \addr Data Science Institute\\
      University of Chicago
      \AND
      \name Qiaozhu Mei \email qmei@umich.edu \\
      \addr School of Information, Department of EECS\\
      University of Michigan}

\def\month{02}  %
\def\year{2025} %
\def\openreview{\url{https://openreview.net/forum?id=Vwgjk5ysWn}} %

\begin{document}

\maketitle

\begin{abstract}
Recent advances in deep neural language models combined with the capacity of large scale datasets have accelerated the development of natural language generation systems that produce fluent and coherent texts (to various degrees of success) in a multitude of tasks and application contexts. However, controlling the output of these models for specific user and task needs is still an open challenge. This is crucial not only to customizing the content and style of the generated language, but also to their safe and reliable deployment in the real world.
We present an extensive survey on the emerging topic of \textit{constrained} neural language generation in which we formally define and categorize the problems of natural language generation by distinguishing between \textit{conditions} and \textit{constraints} (the latter being testable conditions on the output text instead of the input), present constrained text generation tasks,
and review existing methods and evaluation metrics for constrained text generation. Our aim is to highlight recent progress and trends in this emerging field, informing on the most promising directions and limitations towards advancing the state-of-the-art of constrained neural language generation research.
\end{abstract}

\section{Introduction}

Recent advances in the field of natural language generation (NLG) %
\citep{gatt2018survey} have resulted in models able to produce realistic, coherent, and fluent texts in a multitude of natural language processing tasks. Powerful large scale language models can be readily used to perform unconditional language generation, however these models provide little control over attributes of the \textit{generated} texts. Unlike conventional methods which were able to provide fine-grained control over many aspects of the system output including incorporating domain-specific dictionaries, terminology or certain words in the generated output, neural end-to-end approaches remove many of these knobs and switches \citep{post2018fast}. However, imposing constraints on the output generated by these models is crucial for achieving useful and safe language generation in a multitude of real world application scenarios. For example, it can help avoid generic and meaningless responses in dialogue systems \citep{see2019makes}, personalize dialogue agents based on user features that lead to more engaging and meaningful conversations \citep{zhang2018personalizing}, ensure non-offensive sentence completion and friendly communication \citep{liu2019rhetorically}, intervene on the system output in interactive scenarios where domain specific terminology must be included in the generated texts \citep{crego2016systran}, or aid in creative applications such as poetry generation or assisted story writing \citep{peng2018towards}. Moreover, controlling a generic pretrained language model in
order to satisfy certain desiderata helps avoid generating toxic content, prevents demographic biases, can steer generations towards  a desired topic or style \citep{khalifa2020distributional}, and helps communicate intentions in suitable manners for different situations, target audiences and environments \citep{lample2018multiple, li2018delete}. Incorporating prior knowledge and target side constraints in text generative models has numerous applications in many natural language processing areas, including dialogue systems, machine translation, question answering, text summarization, text simplification, image captioning, etc.  Unquestionably, constrained text generation is important in many real-world applications, but compared to other instances of natural language generation, constrained text generation using neural networks remains an open challenge.

We identify the following reasons why constrained neural text generation represents a much harder problem compared to other instances of neural text generation: \textit{i) lack of model expressiveness}: current models are not expressive enough to incorporate arbitrary constraints, defined as testable conditions on the \textit{output} text, into the objective function at training time; \textit{ii) lack of suitable evaluation metrics}: while one can verify whether an output satisfies a constraint or not, it is usually hard to measure \textit{to what extent} an output satisfies a constraint, and it is even harder to jointly evaluate this with other properties of the generated text (such as relevance or coherence); \textit{iii) difficulty in constrained optimization}: even if constraints can be expressed and added to the objective function, they are usually non-differentiable, especially at the token level, which is bad since most methods model and generate text as a sequence of tokens; 
\textit{iv) lack of constrained text generation datasets} that are diverse and representative enough of the variety of practical constraints. %

Commonly used sequential text generation methods and architectures assume a rigid modeling of the output sequence based on word ordering, with tokens  generated progressively one at a time in a standard left-to-right manner \citep{chan2019kermit}.  %
Such autoregressive models cannot easily express constraints at arbitrary positions in the generated sequence or satisfy constraints involving multiple input objects. In addition to these issues, it is generally more challenging to incorporate multiple and heterogeneous constraints, which conform to given rules, topics, sentiments, lexical constraints, or pre-defined stylistic and content attributes.

Our work focuses on the emerging problem of neural natural language generation with constraints. We first define the problem and differentiate between the ambiguous use of \textit{conditions} and \textit{constraints} in natural language generation, including examples that represent instantiations of the constrained neural text generation problem. We then survey approaches, learning methodologies and model architectures employed for generating texts with desirable attributes, and corresponding evaluation metrics. We conclude with open research problems and limitations of current models. Our work aims to draw clear boundaries between the confusing terminology used in the neural language generation literature, highlight the main approaches and discuss how they suffer from the general challenges of constrained text generation, and serve as an informative guide for solving these general challenges and advancing meaningful, useful, and safe constrained NLG research. 

\section{Problem Definitions}
\label{problem_definition}

We formally define the problem of natural language generation, accounting for context, conditions, and constraints placed on text generative models. First, we aim to articulate the key difference between condition and constraint since the distinction between these concepts is rather blurred in the natural language processing literature. Given a text generation task defined as $g(X) \rightarrow X’$, we define \textit{condition} as a testable statement of the \textit{input} $X$, and \textit{constraint} as a testable statement of the \textit{output} $X'$.

Accounting for the distinction above, we divide the text generation problem into three categories: \textit{i) generic or free-text generation} which we present in Section \ref{generic_generation}, \textit{ii) conditional text generation} which we introduce in Section \ref{conditional_generation}, and \textit{iii) constrained text generation} which we outline in Section \ref{constrained_generation}. The focus of our work is on the particular problem of \textit{constrained} text-to-text generation, leaving aside text generation tasks from other types of inputs such as data-to-text generation or image-to-text generation which are \textit{conditional} in nature  according to our definitions.

\subsection{Generic/Free-Text Generation}
\label{generic_generation}

The problem of generic text generation considers the intrinsic history of words generated until the current timestep in the sequence as context, and does not place any external user-defined conditions or constraints on the model output. We formally define it in what follows.

Given a discrete sequence of text tokens $\bm{x}=(x_1, x_2, \ldots, x_n)$ as input where each $x_i$ is drawn from a fixed set of symbols, generic text generation aims to learn the unconditional probability distribution $p(\bm{x})$ of sequence $\bm{x}$. This distribution can be auto-regressively factorized using the chain rule of probability \citep{bengio2003neural} into a product of conditional probabilities $p(\bm{x}) = \prod_{i=1}^{n} p(x_i | x_{<i})$ to perform density estimation and generation of text data. When $p(\bm{x})$ is modeled by a neural network with parameters $\theta$, the neural network is trained to minimize the negative log-likelihood $\mathcal{L}(D)=-\sum_{k=1}^{|D|}\sum_{i=1}^{N_k}\log p_{\theta}(x_i^k|\bm{x}_{<i}^{k})$ over a collection of samples $D=\{\bm{x}^1, \ldots, \bm{x}^{|D|}\}$, where $N_k$ is the length of $\bm{x}_k$. To generate new samples, each token $x_i$ is iteratively sampled from $p_{\theta}(x_i | \bm{x}_{<i})$ and is fed back into the model as the input for the next timestep.

Large scale models for generic text generation show promising abilities to imitate the distribution of
natural language and generate long-term realistic and coherent texts, however such free-text generation models place a lot of burden on the generative model to capture complex semantic and structural features underlying the data distribution; this can often result in repetitive, contradictory, and largely randomized generated texts \citep{holtzman2019curious}. Notably, the content generated by free-text generative models cannot be controlled with respect to particular attributes and modes of the data distribution. This inability to control which regions of the data distribution are generated is particularly problematic considering there is significant toxicity, hate, bias, and negativity present in the large-scale web crawled datasets text generation models are commonly trained on. Imposing conditions or constraints on the generation process results in safer and more useful generated texts for downstream application tasks \citep{krause2020gedi}. 

\subsection{Conditional Text Generation}
\label{conditional_generation}

Conditional text generation manipulates attributes of the generated content depending on specific contexts or user needs, and it allows the data generation process to focus on specific modes of the data. Conditioning the generative model on additional information makes it possible to generate texts which satisfy given \textit{input} conditions and meet desired attributes. In the literature conditional text generation is sometimes referred to as context-dependent text generation. While the word context may carry different semantics for different readers, in this survey we consider as context only attributes which are inherently external to the model itself; model intrinsic attributes such as for example, the history of past generated words, is already included in the formulation of generic text generation. For example, context attributes used for conditioning generated texts are the \textit{source sentence} in machine translation, the \textit{conversational history} in dialogue systems, the \textit{input document} in text summarization and text simplification, the \textit{input question} in question answering systems, the \textit{prompt} given to a large language model, or contextual information such as \textit{product, time}, and \textit{location} in review generation.Essentially, conditional text generation is a form of sequence-to-sequence generation, given the input condition is a text sequence and the goal is to generate another output sequence.

Conditional text generation models add a contextual variable or attribute code $c$ to the probabilistic model $p(\bm{x})$ transforming it into a conditional probability model $p(\bm{x}|c)$, which can
be auto-regressively decomposed using the chain rule of probability $p(\bm{x}|c)=\prod_{i=1}^{n}p(x_i|\bm{x}_{<i}, c)$. When $p(\bm{x}|c)$ is modeled by a neural network with parameters $\theta$, the model minimizes the negative log-likelihood loss function accounting for the attribute code $c$: $ \mathcal{L}(D)=-\sum_{k=1}^{|D|}\sum_{i=1}^{N_k}\log p_{\theta}(x_i^k|\bm{x}_{<i}^{k}, c^{k})$. Besides generation, conditional models can also be used as generative classifiers to compute $p(c|x_{<i})$ by applying Bayes rule.

\subsection{Constrained Text Generation}
\label{constrained_generation}

The problem of constrained text generation is focusing on generating coherent and logical texts that do (not) cover lexical concepts (for eg., pre-defined nouns, verbs, entities, phrases or sentence fragments) desired to be (not) present in the \textit{output}, as well as generate outputs that abide to specific format, semantic, syntactic or utility rules to reflect the particular interests of the system user. Constraints impose restrictions on the generative model that must be satisfied by any solution to the optimization problem and their fulfillment can be tested accordingly.
In the literature the distinction between conditional, controlled, and constrained text generation is not clearly defined, and these terms are often used interchangeably. In fact, the first work that proposed generating constrained text is actually referring to the task as ``controlled'' generation \citep{hu2017toward}. In what follows we formally define the problem of constrained text generation. 

Let us consider we are (optionally) given an unordered or ordered set of $n$ concepts $c=\{c_1, c_2, \ldots, c_n\} \in \mathcal{C}$,  where $\mathcal{C}$ denotes the space of all concepts, and $c_i \in C$ is a concept belonging to the concept vocabulary. In addition, let us assume we are also (optionally) given a set of $m$ rules $z=\{z_1, z_2, \ldots, z_m\} \in \mathcal{Z}$, with $z_i \in \mathcal{R}$, where $\mathcal{R}$ denotes the space of all rules, and each $z_i$ is a text generation constraint expressed in logical form. We formulate constrained text generation as learning the structured predictive function $f : \mathcal{C} \cup \mathcal{Z} \rightarrow \mathcal{X}$, where $\mathcal{C} \cup \mathcal{Z} \ne \phi$ which maps a set of concepts and/or constraint rules to a generated sentence. Therefore, constrained text generation methods impose constraints on the generated sentences and produce output in the form of grammatical sentence $y \in \mathcal{Y}$ which contains all concepts present in $c$ and all constraint rules specified in $z$. The probability $p(y|f)$ can still be modeled autoregressively $p(y|f)=\prod_{i=1}^{n}p(y_i|\bm{y}_{<i}, f)$; when $p(y|f)$ is modeled by a neural network with parameters $\theta$, the negative log likelihood function can be minimized while leveraging $f$ for constraint satisfaction $\mathcal{L}(D)=-\sum_{k=1}^{|D|}\sum_{i=1}^{N_k}\log p_{\theta}(y_i^k|\bm{y}_{<i}^k, f)$.

The matching function $f$ manipulates the probability distribution and indicates to which extent the constraints are satisfied. In the literature, constrained text generation methods can be either \textit{i) Soft-constrained (priming)}, when the matching function $f$ is a soft measure of semantic similarity and only requires the generated sentences to be semantically related to the given constraints, or 
\textit{ii) Hard-constrained}, when the matching function $f$ is a binary indicator which rules out the possibility of generating infeasible sentences that do not meet the given constraints.  
Hard-constrained text generation is notably a more challenging task compared to soft-constrained text generation, and it requires specialized approaches and architectures to ensure the constraints in the output sentence. In contrast, soft-constrained text generation models are usually easier to design, e.g., with the use of existing copy and attention mechanisms for soft enforcing constraints and annotated keyword-text pairs; nevertheless, even soft constraints are likely to be lost during generation, especially if multiple weakly correlated (lexical) constraints must be included \citep{zhang2020pointer}. 

Compared to generic text generation which assumes no conditions on input or output other than existing context, and compared to conditional text generation which places conditions on the input which can be considered at training time, constrained text generation places conditions on the output which is a considerably more difficult and challenging problem to solve. Unlike input conditions, output conditions cannot be considered at training time and their satisfaction is assessed after training has completed by sampling and inspecting the generated outputs. In addition, standard sequence generation architectures are not designed to easily accommodate or incorporate output constraints. Given the model structure itself cannot express output conditions, it becomes challenging to evaluate the extent to which constraints are satisfied by a model, objectively compare and contrast the performance of different models, and measure overall success to inform on progress in constrained natural language generation. Due to these limitations, current methods proposed to address constrained text generation are neither satisfactory nor sufficient. The main machine learning challenge is that it is hard to evaluate the objective function for constrained text generation, and very few studies have approached the problem from the prism of editing the objective function to incorporate constraints at training time. Even if constraints were to be added to the objective function itself, constrained optimization would be another challenge. 
In general, reinforcement learning approaches are used in the context of text generation to optimize non-differentiable reward functions computed at the token level, for eg.,  BLEU in machine translation or ROUGE in text summarization. However, optimizing such automatic measures that focus on local n-gram patterns often results in deteriorated textual outputs despite increased automatic scores \citep{bosselut2018discourse, pasunuru2020dorb}. Moreover, applying reinforcement learning to text generation at the word level leads to difficulty in proper temporal credit assignment for long-term textual rewards \citep{saleh2019hierarchical}. Given that the environment provides only delayed rewards as the agent executes a sequence of actions, it is often impossible to know whether the agent succeeds in achieving a task until the end of the episode, at which point the agent needs to determine which of the actions in the sequence are to be credited with producing the resulting reward \citep{gao2019neural}. Adding constraints on top of existing reinforcement learning issues would be detrimental to the learning process, if not make learning close to impossible: the objective function would be even harder to optimize, rewards would be delayed, sparse and non-informative. Despite these open problems and limitations, we argue neural constrained text generation is an important research area which deserves a lot more attention.

Constrained text generation is useful in many scenarios, such as incorporating in-domain terminology in machine translation \citep{post2018fast}, improving semantic corectness \citep{balakrishnan2019constrained}, avoiding generic and meaningless responses in dialogue systems using grounding facts \citep{mou2016sequence}, paraphrase generation in monolingual text rewriting \citep{hu2019improved, kajiwara2019negative}, incorporating ground-truth text fragments (such as semantic attributes, object annotations) in image caption generation \citep{anderson2017guided}, creating a story \citep{fan2018hierarchical} or poem \citep{ghazvininejad2017hafez} using a pre-defined set of keywords, or re-writing a user search query as a fluent sentence. %
Typical attributes used to generate constrained natural language are the \textit{tense} and the \textit{length} of the summaries in text summarization \citep{fan2018controllable}, the \textit{sentiment} of the generated content in review generation \citep{mueller2017sequence}, \textit{language complexity} in text simplification or the \textit{style} in text style transfer applications. In addition, constrained text generation is used to overcome limitations of neural text generation models for dialogue such as genericness and repetitiveness of responses \citep{see2019makes, serban2016building}. 

Nevertheless, generating text under specific lexical constraints is challenging. Common models and architectures employed for natural language generation are autoregressive in nature, generating tokens one by one in a sequential manner from left to right; by design, these models lack fine control over the generated sequence and cannot easily support constraints at arbitrary positions in the output or constraints involving multiple input objects \citep{zhang2020pointer, hsieh2021enconter}. While for humans it is straightforward to generate sentences that cover a given set of concepts or abide to pre-defined rules by making use of their commonsense reasoning ability, generative commonsense reasoning with a constrained text generation task is more challenging for machine learning models \citep{lin2019commongen}.

\section{NLG Constraints}
\label{nlg_constraints}

Natural language generation models place restrictions on the generated output to produce texts that reflect certain user preferences. In Table \ref{table_tasks} we present NLG tasks distinguishing between conditions and constraints. We broadly group existing constraints into the following categories:

\begin{table*}[!h]
\caption{Overview of constrained NLG tasks, differentiating between conditions and constraints.}
\begin{center}
\scalebox{0.8}{
\begin{tabular}{l l l l l l l}
\multicolumn{1}{c}{\bf Task} & \multicolumn{1}{c}{\bf Condition} & \multicolumn{5}{c}{\bf Constraint} \\
& & \textit{Lexical } & \textit{Format} & \textit{Semantic} & \textit{Syntactic} & \textit{Utility} \\
\toprule
\bf Machine Translation & source input & words & -- & topic & paraphrase & target language \\
 &  & phrases & & sentiment & tense & politeness \\
  &  & entities & & & gender pronouns & factuality/faithfulness \\
& &  & & &  & helpfulness, harmlessness\\
\noalign{\vskip 2mm} 
\midrule
\noalign{\vskip 2mm} 
\bf Dialogue Generation & past utterance(s) & words & length & topic & paraphrase & politeness \\
& & phrases & verbosity & sentiment & gender pronouns & personality traits\\
& & entities & & toxicity & & factuality/faithfulness \\
& &  & &  & & helpfulness, harmlessness \\
\noalign{\vskip 2mm} 
\midrule
\noalign{\vskip 2mm}
\bf Text Summarization & input document(s) & words & length & topic & paraphrase & factuality/faithfulness \\
& & phrases & & & & helpfulness, harmlessness\\
& & entities & \\
\noalign{\vskip 2mm} 
\midrule
\noalign{\vskip 2mm}
\bf Text Simplification & input text & words & length & topic & paraphrase & simpler vocabulary\\
& & phrases & & & & readability \\
& & entities & & & & factuality/faithfulness \\
& &  & & & & helpfulness, harmlessness \\
\noalign{\vskip 2mm} 
\midrule
\noalign{\vskip 2mm}
\bf Text Style Transfer & source text & words & length & topic & paraphrase & style \\
& & phrases & & sentiment & tense &  factuality/faithfulness \\
& & entities & & & gender pronouns  \\
\noalign{\vskip 2mm} 
\midrule
\noalign{\vskip 2mm}
\bf Question Answering & input question & words & length & topic & paraphrase & factuality/faithfulness\\
& & phrases & & & tense & politeness \\
& & entities & & & gender pronouns & helpfulness, harmlessness \\
\noalign{\vskip 2mm} 
\midrule
\noalign{\vskip 2mm}
\bf Narrative Generation/ & -- & words & length & topic & paraphrase & readability \\
\bf Story telling & & phrases & & sentiment & tense & factuality/faithfulness \\
& & entities & & &  & helpfulness, harmlessness\\
& & entities & & & gender pronouns & style\\
\noalign{\vskip 2mm} 
\midrule
\noalign{\vskip 2mm}
\bf Poetry Generation & -- & words & length &  topic & paraphrase & readability \\
& & phrases & rhyme & sentiment & tense & factuality/faithfulness \\
& & entities & rhythm & & gender pronouns & style \\
\noalign{\vskip 2mm} 
\bottomrule
\end{tabular}
}
\label{table_tasks}
\end{center}
\end{table*}

\paragraph{\textbf{Lexical constraints}} Lexical constraints allow to include specific keywords, phrases or entities at arbitrary positions in the  output, and can be specified as a word (a single token) or phrasal constraint (a multi-word phrase). They are useful in tasks such as \textit{dialogue/poetry generation}, \textit{machine translation}, \textit{story telling}, etc. 

\paragraph{\textbf{Format constraints}} Format constraints such as number of sentences, length of sentences, order of words, number of syllables, etc. serve to denote preferences on the form and appearance of the generated output. Format constraints are particularly useful in tasks such as  \textit{poetry generation} to specify the form of the generated poem, for example quatrain or regulated verse, length of the poem, rhyme and rhythm. In \textit{text summarization} or \textit{text simplification}, length constraints define the length of the generated output to be strictly less  than the length of the input document, while in \textit{dialogue generation} they help define the level of verbosity of the dialogue agent.

\paragraph{\textbf{Semantic constraints}} Semantic constraints are used to define the topic and sentiment of the generated content, or control fine-grained aspects such as removing toxicity. Topic constraints are particularly useful in \textit{dialogue generation}, where the goal is to generate on-topic responses that are safe, non-harmful, unbiased, relevant to the dialogue context and particular user needs; in \textit{story telling} or \textit{poetry generation}, topic constraints help define the theme.
Generating language that conveys particular positive, neutral or negative sentiment aims to endow artificial agents with human-like traits such as compassion, empathy, and enables agents to react with appropriate emotion in diverse social situations; constraining on a specific sentiment is important in many tasks such as \textit{dialogue generation}, \textit{review generation}, \textit{story telling}, \textit{poetry generation} or \textit{text style transfer}. Furthermore, increasing politeness of a dialogue system or reducing toxicity of generated language are important aspects with respect
to human-centered metrics of conversation quality.

\paragraph{\textbf{Syntactic constraints}} Syntactically constrained text generation produces sentences with desired syntax by incorporating syntactic templates and rules in the training of the text generative model. Syntactic constraints are useful in paraphrase generation, where given a sentence and a target syntactic form (e.g., a constituency parse), a system must produce a paraphrase of the sentence whose syntax conforms to the target \citep{iyyer2018adversarial}. Generating texts that convey the same meaning but with different expressions has numerous applications in many natural language generation tasks, including monolingual transduction tasks such as \textit{text simplification}, \textit{text compression}, or \textit{text style transfer}, as well as in tasks like \textit{text summarization}, \textit{machine translation} or \textit{question answering} where alternative ways of expressing the same information  capture the inherent language variations. 

\paragraph{\textbf{Utility constraints}} Utility constraints capture holistic properties of the generated output, for example, stylistic, readability, faithfulness, helpfulness, harmlessness or politeness aspects. Preserving the information content of texts while manipulating attributes such as style, readability level, personality traits of the user or specific gender pronouns allows to customize generated texts to different audiences and make them relevant in a wide variety of end-user applications. Stylistic constraints are immediately relevant to the task of \textit{text style transfer}, with applicability in many tasks, including \textit{dialogue generation}, \textit{machine translation}, \textit{text simplification}, \textit{story telling}, \textit{poetry generation}, \textit{review generation}.

Constraining text generation on attributes such as readability and level of text complexity serves to adapt the generated output to users of different age, backgrounds and educational levels. Reducing complexity of texts while preserving the information content is the main goal of \textit{text simplification}; in addition, in tasks such as \textit{dialogue generation}, \textit{text summarization}, \textit{story telling}, \textit{poetry generation}, \textit{question answering} it is important to customize texts for various literacy levels.

In many languages the degree of politeness is an important aspect of inter-personal communication, and honorifics are used to express courtesy, social distance, or the relative social status between the speaker and their addressee(s) \citep{sennrich2016controlling}. Politeness constraints on the output are used in \textit{machine translation}, \textit{dialogue generation}, \textit{story telling}, and \textit{text style transfer}.

Faithfulness constraints enforce similarity between a generated text sequence and its corresponding input, requiring models to generate texts that are faithful, factual and preserve the original information content. Such constraints are important in many tasks, including \textit{text summarization}, \textit{machine translation}, \textit{text simplification} or \textit{dialogue generation}, where models are vulnerable to producing hallucinated content.

Language constraints are useful when translating texts between different languages such as in \textit{machine translation}, or from complex language into simple language such as in \textit{text simplification}.

\section{Constrained Natural Language Tasks}

In what follows we briefly describe major NLG tasks and differentiate between the roles of conditions and constraints in these tasks.

\paragraph{\textbf{Machine Translation}}
Machine translation is focusing on the automatic translation of textual content from one language into another language,
and is a typical example of both \textit{conditional and constrained text generation}, as it conditions on the input text in the source language and constraints the model to generate fluent and faithful output in the target language.  
Additionally, constraints can be placed on the degree of formality and politeness, the use of gender-specific pronouns, the inclusion in the target sentence of named entities or specific concepts from the source sentence.

\paragraph{\textbf{Dialogue Systems}}

A dialogue system, also known as a conversational agent, is a computer system designed to converse with humans using natural language. 
Dialogue generation is an instance of \textit{conditional text generation} where the system response is conditioned on the previous user utterance and frequently on the overall conversational context (e.g., a prompt given to Large Language Model based Chatbots). Dialogue generation can also be an instance of \textit{constrained text generation} - it is desirable that generated dialogues incorporate explicit personality traits \citep{zheng2019personalized}, control the sentiment \citep{kong2019adversarial}, topic, degree of formality and politeness of the generated response to resemble human-to-human conversations. In addition, dialogue responses may need to incorporate text excerpts from past dialogue history or entities such as locations, persons, institutions, etc.
From an application point of view, dialogue systems can be categorized %
into: \textit{i) task-oriented dialogue agents}, designed to help users complete a particular task, or 
\textit{ii) non-task oriented dialogue agents (chat-bots)} designed to carry entertaining conversations with their users on a wide range of open domains. 
A common problem in dialogue generation systems is that they tend to generate safe, universally relevant responses that carry little meaning \citep{serban2016building, li2016diversity, mou2016sequence}. Moreover, they can fail to take turns asking questions and balance specificity with genericness of the output \citep{see2019makes}.

\paragraph{\textbf{Text Summarization}}
\label{text_summarization}

Text summarization facilitates a quick grasp of the essence of a document and produces a condensed version of its content, by copy-pasting the relevant portions from the input as in extractive summarization \citep{nallapati2017summarunner}, or by generating novel content as in abstractive summarization \citep{rush2015neural, nallapati2016abstractive, see2017get}, or via hybrid approaches \citep{liu2018generating} that combine both techniques.
Text summarization is a \textit{conditional text generation task} where the condition is represented by the given document(s); additional conditions are used in remainder summarization to flexibly define which parts of the document(s) are of interest, for eg., remaining paragraphs the user has not read yet, or in source-specific summarization to condition summaries on the specific input source and style of writing, for eg., newspapers, books or news articles.
Text summarization is also a \textit{constrained text generation} task considering that the length of the  summary is fixed, pre-determined, and strictly less than the original document.
The goal of text summarization is to allow the user to digest information at different levels of granularity and detail according to personal needs, interests and time budget.
Moreover, constraints can be placed on specific concepts to include in the summary, such as named entities, or on explicitly picking sentences from the original document as in extractive summarization.

\paragraph{\textbf{Text Simplification}} Text simplification is designed to reduce the text complexity, while preserving its original meaning. %
In the literature, simplification has been addressed at multiple levels: \textit{i) lexical simplification} %
focused on replacing complex words
or phrases with simpler alternatives; \textit{ii) syntactic simplification} %
alters the syntactic structure of the sentence; \textit{iii) semantic simplification} %
paraphrases portions of the text into simpler and clearer variants. End-to-end models attempt to combine all these steps. Text simplification is both \textit{conditional and constrained text generation}; we are conditioning on the input complex text to generate a simpler version, accounting for constraints such as higher readability, simpler vocabulary, and shorter sentence length than the complex input.

\paragraph{\textbf{Text Style Transfer}}
Style transfer has its origins in computer vision applications for image-to-image translation %
and more recently has been used in natural language processing applications for machine translation, sentiment modification to change the sentiment of a sentence from positive to negative and vice versa, word substitution decipherment and word order recovery \citep{hu2017toward}. Text style transfer is designed to preserve the information content of a source sentence while altering the way it is delivered to meet desired presentation constraints. Textual content is disentangled from the style in which it is presented, and manipulating stylistic attributes can be done without parallel aligned data between source and target styles. Text style transfer is an instance of both \textit{conditional and constrained text generation} given that we condition on the given source text and constrain the transferred sentences to stylistically match target examples.

\paragraph{\textbf{Question Answering}}

Question answering systems are designed to find and integrate information from various sources to provide responses to user questions  \citep{fu2018natural}. While traditionally candidate answers consist of words, phrases or sentence snippets retrieved and ranked appropriately from knowledge bases and textual documents \citep{kratzwald2019rankqa}, answer generation aims to produce more natural answers by using neural models to generate the answer sentence. Question answering is both \textit{conditional and constrained text generation} task; the system conditions on the user question, and simultaneously ensures that concepts needed to answer the question are present in the generated output.
Diverse question answering systems are proposed in the literature addressing for eg., medical information needs \citep{wiese2017neural}, %
mathematical questions \citep{schubotz2018introducing}, quiz bowl questions \citep{iyyer2014neural}, cross-lingual and multi-lingual questions \citep{loginova2018towards}. Notably, in practical applications users are not only interested in learning the exact answer word or phrase, but also in how it relates to background information and to previously asked questions and answers \citep{fu2018natural}.

\paragraph{\textbf{Narrative Generation/Story Telling}} Neural narrative generation %
is an important step towards computational creativity \citep{gervas2009computational} %
and represents a long-form open-ended text generation task which simultaneously addresses the selection of appropriate content (\textit{``what to say''}) and the surface realization of the generation (\textit{``how to say it''})\citep{wiseman2017challenges}. Narrative generation is a \textit{constrained text generation} task that places explicit constraints on concepts to steer the narrative in particular topic directions and expands the few keywords specified as the story title, beginning or ending.

While existing models can generate stories with good local coherence,  generating long stories is challenging. Difficulties in coalescing individual phrases into coherent plots and in maintaining character consistency throughout the story lead to a rapid decrease in coherence as the output length increases \citep{van2019narrative}. Hierarchical models for story generation break down the generation process into multiple steps: first modelling the action sequence, then the story narrative, and finally entities such as story characters \citep{fan2019strategies}. Neural narrative generation that interactively combines story-writing with human collaboration improves story quality and human engagement \citep{goldfarb2019plan}. %

\paragraph{\textbf{Poetry Generation}}
The poem generator operates in an interactive context where the user supplies the model with a set of ordered concepts that reflect her writing intent, as well as the format of the poem, for eg. quatrain or regulated verse.
Poetry generation is a \textit{constrained text generation} problem since user defined concepts need to be included in the generated poem, and a \textit{conditional text generation} problem given the explicit conditioning on stylistic attributes. For a detailed overview of poetry generation see \citep{oliveira2017survey}.

\section{Constrained NLG Methods}

Accounting for the different types of constraints introduced in Section \ref{nlg_constraints}, we distinguish the following methodologies commonly employed in the constrained text generation literature: \textit{i)} decoding approaches, \textit{ii)} fine-tuning approaches, \textit{iii)} discriminative approaches, \textit{iv)} edit-based approaches, \textit{v)} adapting existing models and architectures to accommodate constraints on the generated output, and \textit{vi)} prompting large language models. In what follows we present each approach in detail, outlining the main associated challenges. %

\subsection{Decoding approaches}

The most popular approach to text generation in the literature has been supervised learning with task-specific datasets; nevertheless since many real-world applications require diverse and potentially evolving constraints, it is infeasible to annotate task-specific training data for every combination of constraints \citep{qin2022cold}. Even if collecting the data was not a bottleneck, re-training large language models that are extreme in scale for each new constraint or combination of constraints is undesirable. The alternative to fine-tuning language models with task-specific datasets is to enrich decoding algorithms so as to accommodate constraints on the fly. We present decoding approaches to constrained text generation below. 

\paragraph{Lexical constraints} \textbf{\textit{Lexically constrained (guided) decoding}} aims to restrict the search space at decoding time to sequences which contain pre-defined lexical constraints only. These lexical constraints can be specified in the form of a word constraint (a single token) or a phrasal constraint (a multi-word phrase, i.e. a sequence of two or more contiguous tokens). To this end, the beam search decoding algorithm is modified to enforce the inclusion of pre-specified words and phrases in the generated output by allowing the model distribution to not only account for the given lexical constraints, but also to generate parts of the output sequence not covered by the constraints. In general, the decoder can more easily place multiple sequential tokens in a phrasal constraint (where the permutation order is fixed) on the generated output as opposed to placing multiple separate, independent constraints. In addition, the lexically constrained decoding approach assumes lexical constraints are pre-determined, which may not always be the case; if so, the open question is where to get lexical constraints from.

Early work on constrained decoding in machine translation relies on the \textbf{\textit{placeholder approach}} designed to recognize identifiable elements (numbers and named entities) in the source sentence, temporarily replace these with corresponding placeholders during preprocessing, and then substitute the assigned placeholders with the original source-language strings during beam search decoding \citep{crego2016systran, iso2024autotemplate}. Nevertheless, such an approach is limited and unable to model the source tokens in target language specific terminology or the vocabulary from a new out-of-distribution domain. \textbf{\textit{Prefix decoding}} represents a modification of beam search to first ensure that a user defined target prefix is generated first, and only after build hypotheses for the suffix that maximize the coverage of the remaining source-side tokens.
As decoding progresses from left to right, the decoder transitions from a constrained prefix decoding mode to unconstrained beam search. For example, the start of the sentence symbol \textit{$<$s$>$} can be easily included as the first word of a constraint \citep{knowles2016neural, wuebker2016models}. In the context of text summarization, an essential property of a summarization system is the ability to generate a summary with desired length. \textbf{\textit{Grid beam search}} \citep{hokamp2017lexically} extends beam search decoding to allow for the inclusion of arbitrary target side hard lexical constraints at any position in the generated sequence. Given $C$ input constraints, the algorithm maintains $C+1$ separate beams $B_0, B_1, \ldots, B_c$ that group together hypotheses which meet the same number of satisfied constraints. Decoding runs similar to beam search, with an additional dimension added to keep track of how many constraints are met by each hypothesis at every timestep; the highest scoring hypothesis in beam $B_c$ is ultimately generated. However, grid beam search is impractical as decoding complexity is linear in the number of constraints, i.e. beam size increases proportionally to the amount of constraints and changes for every sentence.
\textbf{\textit{Constrained beam search}} \citep{anderson2017guided} guarantees the inclusion of input constraints in the generated sentences by extending beam search with a finite state machine whose states mark completed subsets of the input set of constraints; %
however, decoding complexity has an exponential cost in the number of constraints, making it infeasible in many applications. \textbf{\textit{Dynamic beam allocation}} \citep{post2018fast} improves upon the runtime complexity of grid beam search and constrained beam search by decoding with constant complexity $O(1)$ in the number of constraints. The algorithm still groups together hypotheses that have met the same number of constraints by using a single fixed-size beam which is dynamically divided at each time-step according to how many constraints have been met. Despite being more efficient, dynamic beam allocation does not necessarily outperform conventional beam search \citep{lin2019commongen}. In addition, the generation of hypotheses that only partially satisfy a phrasal constraint needs to be aborted to unwind to the tokens in the constraint. \textbf{\textit{Neurologic decoding}} \citep{lu2021neurologic} modifies beam search to enforce the satisfaction of lexical constraints expressed under predicate logic in conjunctive normal form (CNF). Given the intractability of exhaustive beam search to optimize CNF constraints, the algorithm searches for approximately-optimal output sequences in which all clauses are satisfied, including both positive and negative constraints (i.e. words that must be generated, respectively omitted in the output sequence). The method is applied to cooking recipe generation, where the task is to generate cooking instructions given a dish name and a list of ingredients, and to data-grounded dialogue response generation where a response is generated given a query and a list of facts to convey.

In general, lexically constrained decoding methods have high computational complexity and force the inclusion of specific words in the generated sentence at every timestep of the generation process with no prior examination of these specific words before generation begins \citep{latif2020backward}; this unnatural way of generating sentences can impact the quality and naturalness of the generated output \citep{liu2019bfgan, post2018fast}. In addition, there is a trade-off between the generated
text quality and hard constraint satisfaction \citep{iso2024autotemplate}. In a lack of suitable evaluation metrics, there is no commonly agreed criteria for objectively assessing the quality of the generated sentences and conducting comparisons across NLG models.

\paragraph{Format constraints} \textbf{\textit{Fixed length decoding}} \citep{kikuchi2016controlling} constrains the length of generated summaries in two ways: \textit{i)} by preventing the decoder from generating the end-of-sentence tag until the length of the generated sequence exceeds the desired length, and \textit{ii)} by defining the minimum and maximum length range of the sequence and discarding out-of-range sequences. \textit{\textbf{Non-monotonic decoding}} approaches allow tokens to be inserted at any position in the generated sequence during decoding, therefore accommodating flexible orderings of the output. Unlike left-to-right autoregressive generation that produces a single word at a time, non-monotonic decoding can satisfy lexical constraints at multiple locations in the output sequence allowing for highly parallel generation and faster decoding times. Nevertheless, such approaches assume the generated sequence length is known a priori, preventing it from being dynamically adjusted as generation proceeds. Moreover, such models assume conditional independence between output tokens, i.e. tokens are generated independently, may be inconsistent and agnostic to each other. Consequently, this approach may hurt the expressiveness of the model and lead to potential performance degradation, impacting the fluency and naturalness of the output. In addition, non-monotonic sequence decoding approaches can terminate prematurely before constraints are satisfied in the output sequence \citep{zhang2020pointer, hsieh2021enconter}. The main limitation of this approach is the lack of model expressiveness in accommodating constraints.

Insertion Transformer \citep{stern2019insertion} proposes a flexible sequence generation
framework based on repeated insertion operations into an initially empty output sequence until a termination
condition is met. The model adopts a progressive masking approach based on token importance in the original text and is trained to generate a missing token between every two tokens in the input.
 To this end, the original Transformer \citep{vaswani2017attention} decoder is modified to allow insertions not just at the end but anywhere in the output sequence. The model can decode sequences serially one token at a time, or it can decode sequences
in parallel with simultaneous insertions at multiple locations. A similar approach is considered in InDIGO \citep{gu2019insertion} which extends Transformer for insertion-based decoding with inferred generation order. Token generation order for the output sequence is modeled as a latent variable, and at
each decoding step the model predicts both the generated word and its position in the output sequence; nevertheless, strong conditional independence is assumed between the output tokens which hurts output quality. An iterative refinement step based on latent variables is added to the Transformer decoder to refine a target sequence gradually over multiple steps until a predefined stopping criterion is met \citep{lee2018deterministic}. Progressive Insertion Transformer \citep{zhang2020pointer} uses non-autoregressive modeling based on a top-down progressive structure for lexical hard-constrained text generation. Given lexical constraints as input, the model inserts tokens progressively according to word importance to generate the target sequence, as follows: first it generates high-level words in a sentence such as nouns, adjectives and verbs, then uses these as pivoting points to insert details of finer granularity and finally completes the sentence by adding connecting words which carry less information, such as pronouns and prepositions. Entity Constrained Insertion Transformer \citep{hsieh2021enconter} builds upon previous models considering hard lexical constraints in the form of entities in the output sequence. Similar approaches train the Transformer decoder to insert missing tokens in a partially complete sequence without relying on a pre-specified factorization of tokens \citep{chan2019kermit, gu2019levenshtein}; based on the information available in the sequence, the insertion-based generative model is able to dynamically infer the remaining parts irrespective of their arbitrary order. 

\paragraph{Syntactic and Semantic constraints} \textit{\textbf{Distributional constraints}} \citep{baheti2018generating} on topic and semantic
similarity are used to incorporate source side-information at decoding time in neural conversational systems and encourage the generation of more diverse responses. Moreover, constraints over topics and syntax are used to generate matching or semantically similar statements in response to the user input \citep{niu2018polite}. Lexically constrained decoding from pre-trained language models aims to steer language models in useful and safe directions so as to minimize the risks associated with these models generating biased, offensive and toxic content \citep{sheng2019woman, holtzman2019curious}. \textbf{Weighted decoding} methods employ a linear combination of output logits from multiple prompts (raw prompt vs prefix-prepended prompt) to vary the strength of desired target attributes in the output text \citep{pei2023preadd, zhang2022discup}. \textbf{Energy-based constrained decoding} allows the specification of style and lexical constraints through an energy function and performs differentiable reasoning through gradient-based sampling \citep{qin2022cold}, \citep{mireshghallah2022mix}. Constrained text generation is viewed as an optimization problem, where the goal is to iteratively seek text with
lower energy. The sampling process uses gradients of the energy function to update a continuous relaxation of text data, which is then mapped back to the discrete space of natural language via a discretization approach. For streering the generation towards desired constraints, biases are applied to the logits of the pre-trained model output layer, which is also found to improve the speed of the decoding process \citep{liu2023bolt}. Nevertheless, sampling from energy-based models requires many iterations to converge to plausible text.

\subsection{Fine-tuning approaches}

{\paragraph{Semantic and Utility constraints} Controlling the output of pre-trained language models is crucial in a wide-range of safety-critical applications, including mental health support chatbots, sentiment controlled text generation, language detoxification,  etc. To this end, fine-tuning approaches are used for fine-grained control over individual stylistic aspects (for eg., length, professional and descriptive style, tense, personal voice, gender) and content aspects (for eg., sentiment and topic) of the generated texts
\citep{ficler2017controlling}, \citep{lample2018multiple}. Typically, the pre-trained model is fine-tuned separately for each attribute of interest, which poses the challenge of how to learn disentangled latent representations of style and content in neural language models \citep{john2019disentangled} and isolate the desired attribute from the distribution shift between the generative model and the fine-tuned dataset. The lack of datasets that are diverse and representative of constrained criteria encountered in practice represents an open challenge for fine-tuning pre-trained models.

CTRL \citep{keskar2019ctrl} uses control codes  to generate texts that meet user-defined constraints on domain, style, topics, dates, entities, relationships between entities, plot points, and task-related behavior. These pre-defined codes are appended at the beginning of raw text sequences to define task-specific training data and create controllable task-specific behaviour at sampling time. Similarly, fine-grained semantic control codes are used to steer generation towards targeted attributes \citep{ross2022tailor}. Decoding Experts (DExperts) \citep{liu2021dexperts} is a decoding-time method for constrained text generation which combines a pre-trained language model with both an “expert” and “anti-expert” language model in a product of experts. The “expert” models desirable aspects of the generated text (for eg., positive sentiment), while the “anti-expert” plays the antagonistic role of modeling undesirable attributes to be avoided (for eg., toxicity); each one of the three language models is conditioned on the same user prompt. While the method highlights the promise of customizing decoding from pre-trained language models in safe and efficient ways, gathering large amounts of toxic data to model undesirable attributes may be challenging. In general, adding negativity to a positive prompt is a much easier task than adding a positive turn to a negative prompt \citep{madotto2020plug}. Augmenting the set of positive examples commonly used for LLM training with a set of negative examples, i.e. completions given a prompt that a model should not generate, helps reduce the likelihood of repetitive model generations and negative tokens occuring in the output \citep{adolphs2023cringe}. Similarly, token-level or sequence-level objectives are used to discourage LLM models from assigning
high probabilities to certain tokens or sequences \citep{lu2022quark, liu2021constrained}. Aligning LLM models with user preferences at inference time is important for personalizing LLM models to their users without the need to retrain the model for each new target attribute. SteerLLM \citep{dong2023steerlm} allows end-users to customize responses during inference by conditioning a supervised finetuned model on a multi-dimensional set of user attributes. 

Reinforcement learning from human feedback (RLHF) \citep{christiano2017deep, stiennon2020learning} is a key component in improving the instruction-following and generation abilities of LLM models. 
Aligning AI models with human preferences is considered crucial for safely deploying artificial systems in the real-world and ensuring they exhibit behaviors consistent with human
values \citep{ouyang2022training, chatgpt2022, ziegler2019fine, shen2023large, wang2024comprehensive}. RLHF algorithms further finetune under a KL-constrained RL objective a language model that has already undergone supervised fine-tuning (referred to as the reference model); this objective encourages the model to maximize the reward and simultaneously discourages high KL divergence between the language model and the reference model. The reward model is derived from human preferences on text continuations with positive sentiment or vividly descriptive language, while the KL constraint is used to prevent the fine-tuned policy from drifting too far from the reference policy. In dialogue systems, KL control has been used to retain prior information and penalize divergence from the pre-trained model during RL fine-tuning \citep{jaques2019way}. Despite the promise of RLHF in aligning LLMs to user preferences, fine-grained control over large language models remains a significant challenge \citep{wang2024arithmetic}. Since RLHF training can be difficult and unstable, alternative approaches finetune language models to mimic the Best-of-N (BoN) \citep{nakano2021webgpt, touvron2023llama} distribution  by minimizing the KL divergence between the language model and the BoN distribution, for example when generating movie reviews with
positive sentiment \citep{amini2024variational} or in text  summarization \citep{gui2024bonbon}. Nevertheless, KL does not capture attributes of generated text that humans judge to be salient, such as the length of the generated response \citep{singhal2023long}. The lack of suitable evaluation metrics that correlate with human judgements is a bottleneck in generating high quality outputs.  

Pre-trained OpenAI-GPT2 \citep{radford2019language} model is used to re-write a story through counterfactual reasoning and generate a narrative consistent with the imposed constraints \citep{qin2019counterfactual}. In abstractive summarization, OpenAI-GPT2 is used in a reinforcement learning setting which trains the summarization agent to maximize coverage and fluency of the generated content constrained on a pre-defined length \citep{laban2020summary}. RecipeGPT \citep{h2020recipegpt} fine-tunes the GPT-2 pre-trained language model for generating cooking instructions when hard constraints are placed on the recipe title and ingredients; the model can also generate the list of ingredients for a recipe when constrained on the recipe title and specific cooking instructions. Infilling by language modeling is used to complete variable length text spans (e.g. words, n-grams and sentences) by fine-tuning a pre-trained language model on sentence pairs that contain both artificially-masked text and the corresponding original text \citep{donahue2020enabling}. 

While fine-tuning models on task specific datasets has become the dominant paradigm for constrained text generation from pre-trained large language models, these models generally fail to reliably incorporate the underlying constraints in the generated texts even when supervised with large amounts of task-specific examples \citep{lu2021neurologic}.  Moreover, the superficial alignment hypothesis \citep{zhou2024lima} argues that that almost all knowledge in large language models is learned during pre-training and that alignment tuning only teaches the base LLM in which data format and language style to interact with its users. Further analysis of this hypothesis finds that fine-tuning approaches only influence a small number of tokens focused primarily on stylistic elements \citep{lin2023unlocking}. The lack of model expressivity to incorporate constraints in an important challenge for fine-grained constrained text generation.

\paragraph{Format constraints} Due to length biases in reward models, LLM models fine-tuned with RLHF tend to suffer from verbosity issues and generate long answers that are not necessarily of high quality \citep{singhal2023long, park2024disentangling, dubois2024alpacafarm, kabir2023answers, nakano2021webgpt, sun2023exploring, wu2024fine}. Offline preference optimization algorithms with implicit reward are known to exploit evaluator length biases and quickly increase the length of the generated text during training without capturing more complex
features of human preferences \citep{rafailov2024direct, li2023alpacaeval, dubois2024length}. To disentangle verbosity from quality, length-controlled direct preference alignment methods employ an additional regularization term in the loss function that governs the token length of the generated response \citep{park2024disentangling}. Online RLHF approaches that first train a reward model include a similar length regularization term in the reward modeling stage \citep{chen2024odin}. Open problems with current RLHF frameworks include their limited ability to capture and adapt to the complexity of user-dependent preferences in the real world \citep{wang2024arithmetic}. Since human preferences can change over time depending on users and their expectations, methods relying on multi-objective reward models are used to capture desirable aspects of the generated texts  (eg., verbosity, factuality, helpfulness,  harmlessness) \citep{pan2023rewards, rame2024rewarded, dong2023steerlm, bakker2022fine, wu2024fine}. To enhance the control and personalization of a single LLM model across desired range of attributes, different user preferences are encoded as unit vectors and embedded numerically into the system prompt \citep{wang2024arithmetic, yang2024rewards, dong2023steerlm}. The analysis of the alignment tuning process reveals that RLHF simply teaches the base LLM model to select a sub-distribution of data formats for interacting with the user \citep{lin2023unlocking}. Notably, most distribution shifts occur with stylistic tokens (transitional phrases, discourse markers, safety disclaimers) instead of content-bearing words, which supports the \textit{superficial alignment hypothesis} \citep{zhou2024lima}.

\subsection{Discriminative approaches}

\paragraph{Utility constraints} One of the early works proposing constrained text generation learns disentangled latent representations by combining variational auto-encoders with attribute discriminators \citep{hu2017toward}. Semantic structure is imposed on the latent codes by using global discriminators, one for each attribute, to guide the learning of the discrete text generator and force it to allocate one latent dimension per attribute code. The model is used to manipulate the sentiment and tense of the generated sentences.

Weighted decoding \citep{holtzman2018learning} relies on a mixture of discriminative models to guide a recurrent generator towards incorporating attributes that enhance the overall coherence, style, and information content of the generated text. The discriminators complement each other and their weighted contributions form the final decoding objective from the generator. Similarly, stylistic configurations are revised and polished for generated poems by adding additional weights during decoding to control the style of generated poem, including the repetition, alliteration, word length, cursing, sentiment, and
concreteness \citep{ghazvininejad2017hafez}. Nevertheless, modifying the scoring function used for generation as in weighted decoding often leads to sacrificing fluency and coherence of the generated text \citep{see2019makes}. Selective sampling \citep{wang2017steering} relies on a sample selector (multilayer perceptron for binary classification) which outputs whether the current sample should be accepted or rejected based on the presence of desired target words that define the output style and topic in the generated sequence. The robustness of evaluation metrics is directly correlated with model performance, therefore it is crucial to focus on developing metrics that capture diverse aspects of text quality during training and sampling time.

Generating texts with desirable attributes from a pre-trained unconditional language model $P(X)$ is a non-trivial task. Most approaches resort to either training from scratch a new conditional model $P(X|a)$ for desired attribute $a$, or fine-tuning $P(X)$ on additional data representative for the attribute $a$. Theoretically, rejection sampling could also be used to sample $P(X|a)$ from $P(x)$, but this approach is highly inefficient in practice. Fudge \citep{yang2021fudge} generates text conditioned on a desired attribute $a$ (for eg., topic control in language generation, degree of formality in machine translation) while only accessing the output probabilities $P(X)$ of generative model $G$. Given an incomplete sequence prefix, the model trains binary discriminative models for one or multiple desired attributes to predict whether the attribute(s) will be fulfilled in the future complete sequence, therefore evaluation is an important challenge. The output probabilities of the discriminator(s) are then multiplied with the output logits of the generator $G$ to adjust the original probabilities of $G$ accounting for desired attribute(s) $a$ and model $P(X|a)$ via a Bayesian decomposition.

PPLM \citep{dathathri2019plug} combines a pre-trained language model with attribute classifiers that guide generation towards specific topics and sentiment styles. These classifiers are trained on top of the last hidden layer of the pre-trained language model, and gradients from the classifiers are backpropagated to update the hidden representations of the language model and steer generation in desirable directions. While PPLM  achieves fine-grained control of content and style attributes via a simple gradient-based sampling mechanism, the approach is computationally intensive and inefficient as it requires multiple forward and backward passes for each generation step. Plug-and-play methods have been used to control large pre-trained conversational models such as GPT-2 \citep{radford2019language}  using a variety of styles (positive and negative sentiment) and topics (Question, Sport, Business, Finance)  \citep{madotto2020plug, liu2020data}. Plug-in Language Model (PiLM) \citep{yang2024plug} manipulates the latent state of the language model using a regression model to generate texts that adhere to specific topics or sentiment. More effort needs to be focused on collecting datasets for constrained text generation that capture real-world constraints.

GeDi \citep{krause2020gedi} guides language generation from large language models towards desired attributes by using generative discriminators to compute classification likelihoods for all candidate next tokens on the fly at generation time. Given a class-conditional language model conditioned both on a desired attribute $c^{+}$ and an undesired attribute $c^{-}$, GeDi-guided contrastive generation uses the two instances of the model as discriminative classifiers to contrast and filter out common attributes between the two classes $c^{+}$ and $c^{-}$; then aspects of the desired attribute $c^{+}$ are transferred across domains via weighted decoding and filtering. The contrast between a positive and a negative class conditional distribution is employed both at training and inference time to control the bias, toxicity and negativity of GPT-2 \citep{radford2019language} and GPT-3 \citep{brown2020language}.
 Recent approaches explore the role word embeddings can play in debiasing LLMs and steering their generations in particular sentiment directions \citep{subramani2022extracting, turner2023steering, li2021prefix}. LM-Steer \citep{han2023lm} leverages the fact that linear transformations in
output word embeddings are equivalent to style changes in LLM generation; the method linearly transforms output word embeddings at decoding time using learnt parameters representative for each target style.

\subsection{Edit based approaches}

\paragraph{Utility constraints} Edit based approaches rely on the key idea that changing only a few words or phrases which are indicative of a particular attribute are sufficient to alter the style of a given piece of text. For example, 
the sentiment of a sentence can be altered from negative to positive by first identifying negative attribute markers ("bad", "worst", "disappointed"), deleting these negative attributes while keeping other content words fixed, and then generating the final output via a recurrent decoder which conditions on the extracted content words and the target attribute  
\citep{li2018delete}. Leaving from the observation that humans write text in incremental passes with multiple revisions, a prototype-then-edit model  first samples a prototype
sentence from the training corpus and then edits it conditioned on an edit vector \citep{guu2018generating}. Noticeably, text generation based on editing a prototype is much easier compared to generating text from scratch. Also building upon the "Delete Retrieve Generate" framework, the Generative Style Transformer \citep{sudhakar2019transforming} incorporates a neural mechanism to delete style attributes from the source sentence based on the attention weights of a Transformer model (Delete Transformer), and then generates sentences in the desired target style by decoding with a pre-trained GPT-2 \citep{radford2018improving} model. 

Activation editing \citep{li2022emergent, hernandez2023measuring, li2024inference} methods discover directions in the representation space that correspond to encodings of specific attributes (such as sentiment, topic, style, factual information, truthfulness, etc). When these encodings are added to the internal representations of large language models, they act as knowledge editors by manipulating the generated output to be consistent with desired constraints \citep{kong2024aligning}. The advantage of representation edits is that they enable constrained generation without the need to rely on textual prompts, while enhancing LLM control and intepretability. Steering vectors \citep{subramani2022extracting, turner2023steering} are added to the hidden states of a language model to generate texts with desired style or sentiment. Other editing approaches manipulate model weights instead of representations by inserting updated factual knowledge directly into model specific parameters \citep{meng2022locating, meng2022mass, mitchell2021fast, dai2022knowledge, rawat2021modifying, ilharco2022editing, orgad2023editing}. Nevertheless, open challenges with these models include indeterminate/uncertain editing boundaries, failing to account for contextual information and entailed consequences of edited facts \citep{liu2024evedit, cohen2024evaluating, zhong2023mquake}. Evaluations focused on assessing the success of model edits mostly consider a few tokens generated after an input prompt, and do not measure the consistency of edits over a long generation of text; failure modes observed in long-form generation include topic drift,  lexical cohesion issues, gradual and catastrophic forgetting of previous edited facts - these aspects limit the usefulness of model editing methods at scale \citep{rosati2024long, gupta2024model, li2023evaluating}. 

\paragraph{Lexical constraints} Metropolis-Hastings sampling \citep{miao2019cgmh} first inserts all constraint keywords in a template in random order, then samples local edit operations (word replacement, deletion or insertion) to perform at specific positions for improving sentence fluency. The probability of each edit operation being accepted or rejected is determined by a language model, however individually sampling each token results in slow convergence. Instead of randomly sampling edit operations, the gradient of a differentiable objective function is used to determine where and how to edit \citep{sha2020gradient}. The majority of editing approaches model a single edit step, unlike humans who do iterative refinement and editing of the content.   Modeling the whole process of iteratively generating sequences leverages neural network models to describe the likelihood of multi-step edits, with improved performance over modeling single-order edits \citep{reid2022learning}. Self-correction \citep{welleck2023generating} incorporates a learnt mechanism to iteratively revise LLM model outputs and correct imperfect generations that violate lexical, mathematical reasoning and toxicity constraints.

\subsection{Adapting existing models and architectures to accommodate constraints}

It is non-trivial to impose constraints on existing deep learning models while maintaining high generation quality since their model architecture is designed to generate sentences sequentially from left to right. %
While current deep learning models are lacking the expressiveness to incorporate constraints at training time and at arbitrary positions in the generated sequence,
well known models and architectures are adapted to accommodate constraints through a set of custom engineered approaches. We present these methods below.

\paragraph{Lexical constraints} Current architectures used for language generation produce texts sequentially from the first word to the last word, and it is non-trivial to impose lexical constraints on left-to-right  generation while maintaining high output quality for natural and fluent texts. Current workarounds for hard lexically constrained text generation address this limitation by generating texts in a non-monotonic fashion when employing \textit{\textbf{forward-backward language models}}. The backward language model takes a lexical constraint as input and generates the first half of the sentence backwards conditioned on the topic word, while the forward language model takes as input the sequence generated by the backward generator and produces its sentence completion in normal order conditioned on the backward generated sequence. While the topic word can occur at any position in the sentence, this approach can only generate output constrained on at most one lexical constraint; generating sequences with multiple lexical constraints is an open research problem. These approaches adapt existing frameworks for constrained text generation by splitting a sentence into two parts, which is unnatural and also hurts fluency when generating half of the sequence in reverse order. 

Given a topic word at an arbitrary position in a scientific paper title, a recurrent language model is tasked with generating both past and future words in the title conditioned on the given topic \citep{mou2015backward}. Similarly, on-topic dialogue responses that satisfy hard lexical constraints are generated with a "sequence to backward and forward sequences" (seq2bf) model \citep{mou2016sequence} which first predicts a keyword noun that reflects the gist of the response, then decodes the response backward and forward starting from the given word.
BFGAN \citep{liu2019bfgan} employs GANs for lexically constrained text generation (product reviews, conversational responses). The model incorporates three modules, namely a backward generator and a forward generator which collaborate on generating lexically constrained sentences, and a discriminator which guides the joint training with policy gradient of the two generators. 

Generating a fluent sequence which simultaneously satisfies multiple lexical constraints employs a backward-forward LSTM language model to first  generate the sequence from a user-defined verb constraint and then satisfy other lexical constraints by word embedding substitution based on cosine similarity between generated tokens and desired constraints \citep{latif2020backward}. Nevertheless, the approach assumes a verb constraint is always specified in the set of lexical constraints.

\textbf{Semantic and Utility constraints} Steering neural models in specific directions is achieved by: \textit{i)} \textit{\textbf{adding special tokens at the beginning}} %
\textit{\textbf{or end}} %
\textit{\textbf{of the source text}}, \textit{ii)} \textit{\textbf{incorporating additional conditions into the decoder hidden states}}  %
and \textit{iii)} \textit{\textbf{connecting the conditions directly to the decoder output layer}}.
A topic aware sequence-to-sequence model is used to generate on-topic conversational responses by conditioning the decoder on specific topic words \citep{xing2016topic}. Imposing conversational goals on dialogue agents aims to guide the conversation towards a designated target subject by combining coarse-grained topic constraints with discourse-level rules \citep{tang2019target}. Generating emotional responses in neural conversational systems is achieved by feeding the emotion category embedding to a sequence-to-sequence decoder \citep{zhou2018emotional}. %
Personalized chit-chat dialogue agents that display consistent personalities, viewpoints and are configurable depending on attributes of the system user are used to produce more personal, specific and engaging dialogue responses \citep{wang2017steering, bosselut2018discourse, zhang2018personalizing}. 
Nevertheless, finding the proper balance between fluency, engagement, consistency and a persistent personality remains an open challenge for current dialogue models due to lack of a measurable objective function and correspondingly suitable evaluation metrics. While it is possible to judge whether or not an output satisfies one constraint, it is hard to judge the extent to which (“how much”) the constraint is satisfied; it is even harder to jointly model/measure multiple constraints. Moreover, accounting for repetition and diversity is important as these models often get stuck in an infinite loop of redundant, dull, generic and universally relevant responses that carry little meaning \citep{li2016deep, see2019makes, mou2016sequence}.

For integrating factual knowledge into open-ended conversational systems, factoid and entity-rich web documents are encoded altogether with the conversation history into the same representation which is passed to an attentional neural decoder that generates the response tokens. Similarly, speaker-level representations are integrated into seq2seq conversational models for generating personalized conversation responses  \citep{li2016persona}. Fact-guided sentence modification for dynamically rewriting, updating or correcting articles according to changing information is an instance of constrained text generation which presents the particular challenge that the rewritten sentence needs to be consistent with an input claim while at the same time preserving non-contradicting content \citep{shah2020automatic}. Given the claim and an old sentence, an updated sentence is produced by first identifying contradictory components in the input sentence, masking these, then using the residual sentence and the claim as input into a two encoder sequence-to-sequence model with copy attention to produce the update sentence consistent with the claim.  Syntactically controlled paraphrase generation produces paraphrases of an input sentence by constraining the system on the target syntactic form
\citep{iyyer2018adversarial}, however not many syntactically constrained datasets to learn from are available.

Controllable story generation with RNNs is used to influence the story ending valence (whether happy or sad) and the storyline (specified as a sequence of words) \citep{peng2018towards}. Story-telling methods commonly use a hierarchical approach to thematically consistent story generation, by first generating a prompt describing the topic for the story, and then constraining on the prompt for
generating the story content \citep{fan2018hierarchical}; %
additionally, constraints on the presence of entities are included as well \citep{clark2018neural}. Open-domain story generation requires composing coherent natural language
texts that describe plausible sequence of events and is more challenging compared to generating stories in a narrow domain given an existing plot.

Unsupervised machine translation methods are adapted for the task of text-style transfer by incorporating stylistic constraints in a neural seq2seq model with attention and \textbf{\textit{using a style classifier to guarantee the accuracy of style transfer}} \citep{zhang2018style}, or for control over multiple style attributes, including gender, sentiment or product type \citep{lample2018multiple}. In machine translation, honorifics constraints are important for producing socially appropriate forms of address and controling the level of courtesy \citep{sennrich2016controlling}; the system user defines the desired
level of politeness of the translation, however these user-defined constraints are only soft constraints and can be overridden by the attentional encoder-decoder machine translation system whenever the source
text provides strong politeness clues. 

For effective imposition of semantic structure in constrained text generation, latent space representations need to be disentangled  %
\citep{john2019disentangled}, 
such that varying an individual latent code will only change a single desired attribute. VAEs can achieve meaningful latent representations with designated semantics when combined with \textbf{\textit{attribute discriminators}} and optimized end-to-end with differentiable softmax approximation  \citep{hu2017toward}; this allows to generate sentences with constraints on sentiment and tense. Given an input sequence and a set of labels, sequence transduction with multi-space variational autoencoders  \citep{zhou2017multi} generates an output sequence that alters the content of the input sequence according to the constraints specified by the labels; the method is used for morphological inflection in multiple languages. In general, constrained text generation approaches assume that constraints need to be known a priori; however, this is not always possible, for eg., when suggesting alternative phrases for search queries in real-time, or when generating responses in dialogue systems according to the dynamics of the conversational context. Recent constrained text generation approaches control attributes of a generated sequence based on another sentence example: given two sentences $X$ and $Y$, the goal is to generate a new sentence $Z$ that follows the semantics of $X$ and the syntax of $Y$. To this end, a VAE model with two latent variables is used to achieve disentanglement in the continuous latent space between syntax and semantics \citep{chen2019controllable, bao2019generating}. 
Topic guided VAEs \citep{wang2019topic} use a Gaussian mixture model prior where each mixture component corresponds to a latent topic extracted from data as opposed to using pre-defined parameter settings which do not incorporate semantic meaning into the latent codes; the model is used for text summarization with designated topic guidance. Abstractive and extractive sentence compression with VAEs assumes the existence of a background language model from which a latent summary sentence is drawn first, and then the observed sentence is generated  conditioned on the latent summary \citep{miao2016language}; the model is able to balance copying a word from the source sentence with generating it from the background distribution. Iterative refinement of a sequence to transform it into another sequence with desired attributes exploits geometry of the latent space to produce incremental higher-quality revisions with theoretical guarantees in the combinatorial space of sequence elements \citep{mueller2017sequence, shen2017style}. 
Such latent variable manipulations can rewrite modern text in the language of Shakespeare, improve sentence positivity, address word substitution and word order recovery tasks without need for any revision examples. Constraints on the use of metaphor and personification in poems are incorporated in a conditional VAE with a rhetorically controlled decoder trained to emit meaningful and diverse rhetoric and overcome generic sentences \citep{liu2019rhetorically}. Variational neural machine translation \citep{zhang2016variational} incorporates a continuous latent variable to model the underlying semantics of sentence pairs. 
Nevertheless, efficiently performing posterior inference and large-scale training during the incorporation of latent variables remains an open challenge for constrained VAEs. Finding structure in the latent space that corresponds to particular sentiment makes it possible to steer the generation towards desired sentiment by adding the sentiment direction vector to the residual stream when generating sentence completions \citep{tigges2023linear}. 

Modifying textual attributes of sentences including sentiment, style, tense, voice, mood and negation is achieved by \textbf{\textit{incorporating conditioning information into a neural encoder-decoder model}}, and optimizing a reconstruction loss which interpolates between auto-encoding and back-translation components to encourage content compatibility, as well as an adversarial loss which encourages sentence-level stylistic attribute compatibility \citep{logeswaran2018content}. The model allows simultaneous conditioning on multiple textual attributes, however the extent to which the generated sentences match the conditioning information requires new objective evaluation metrics for attribute accuracy and content compatibility/preservation. 

Style transfer between scientific papers and newspapers is performed with \textbf{\textit{separate style decoders}}, or by generating both content and style from the same decoder \citep{fu2018style}. In poetry generation, it is common to impose hard constraints on rhyme, rhythm, and topic \citep{ghazvininejad2016generating,ghazvininejad2017hafez}. %
Given a user-supplied topic, the poetry generation algorithm first generates a large set of on-topic words and phrases, assigns rhyming words and phrases to specific lines, and then combines finite-state machinery with an RNN language model to score plausible poems that meet the desired constraints. While \textit{\textbf{augmenting an RNN with a working memory}} to explicitly maintain a limited history of generated topics and context, coherence in meaning and topics across the overall poem remains an important challenge \citep{zhang2014chinese}. Constrained recurrent models are also used to generate online product reviews of certain topic, sentiment, style and length \citep{ficler2017controlling}, affective dialogue responses \citep{ghosh2017affect}, or for modeling participant roles and topics in conversational systems \citep{mei2017coherent}. 

Alternative non-autoregressive architectures based on continuous diffusion models are adapted for text generation with semantic and syntactic constraints \citep{yang2022diffusion, austin2021structured, gong2022diffuseq}. Diffusion-LM \citep{li2022diffusion} gradually denoises  a sequence of Gaussian noise vectors into word vectors, resulting in a hierarchy of continuous latent representations which enables gradient-based methods to steer the text generation process. Nevertheless, training of diffusion models is slower to converge and decoding from these models takes longer time. To speed up the inference process, adaptive sampling strategies are applied for different generation stages in the context of story generation \citep{tang2023can}.

\paragraph{Format and Utility constraints} Text simplification models parameterized on constraints such as length, amount of paraphrasing, degree of lexical and syntactic complexity are used for generating texts easier to read and understand with simpler grammar and structure \citep{martin2019controllable}. Towards a similar goal of controlling the degree of lexical complexity, the \textit{\textbf{training loss function is changed to assign weights to words based on their complexity level}} \citep{nishihara2019controllable}. In text summarization, constraints on the output sequence length  for neural encoder-decoder
models are specified as \textit{\textbf{length embeddings}} and are passed as additional input to the decoder \citep{kikuchi2016controlling}.

Faithfulness in abstractive text summarization is enforced in a seq2seq model by conditioning on both the source text and extracted factual descriptions \citep{cao2018faithful}; this helps avoid generating false facts in  the output summary. Hybrid text summarization approaches combine an unsupervised sentence extractor which selects salient sentences from the input document with a sentence abstractor that paraphrases each extracted sentence to overcome limitations of parallel aligned datasets \citep{nikolov2020abstractive}.

Reinforcement learning is used for constrined NLG to directly optimize non-differentiable reward functions and evaluation metrics. While any user-defined reward function can be employed for training, most frequently optimized metrics with RL are BLEU for machine translation \citep{ranzato2015sequence}, %
ROUGE for text summarization \citep{ranzato2015sequence, paulus2018deep, wu2018learning, gao2019reward}, or human-defined conversation metrics focused on coherence, informativeness, sentiment, politeness, toxicity, question, repetition or semantic similarity \citep{li2016deep, saleh2019hierarchical, wu2018learning}. %
However, manually defined reward functions based on heuristics  cannot cover all crucial aspects of a natural realistic conversation \citep{bosselut2018discourse, gao2019reward}. In addition, rewards are commonly modeled at the word level accounting for the probability of generating each word in a sentence \citep{ranzato2015sequence, jaques2019way}; such low-level control makes credit assignment challenging since the number of actions available to the RL agent is equivalent to the number of words in the vocabulary. Defining a global score that measures complex aspects of text quality beyond local n-gram patterns and which can reliably approximate human judgments of text quality remains an open challenge \citep{bosselut2018discourse}.

In the RL framework the generative model is seen as an agent with parameters that define a policy and which interacts with an external environment by taking actions, receives a reward once it reaches the end of a sequence and  updates its internal state consequently. 
To this end, policy gradient methods %
are used to train  text generative models and alleviate issues such as exposure bias and loss functions which do not operate at the sequence level. However, policy gradient algorithms present large variance and generally struggle in settings with large action spaces such as natural language generation. In addition, they take very long time to converge \citep{choshen2019weaknesses} and the improvement in the optimized metrics is not always reflected in human evaluations of text quality. %
Training RL models to optimize n-gram evaluation measures based on local patterns provides only a limited and myopic perspective of overall text quality and does not necessarily lead to better text quality, overall coherence or discourse structure \citep{bosselut2018discourse}. Moreover, fine-tuning on such measures may yield deteriorated outputs despite increased automatic scores, while difficulty in constrained optimization with RL often leads to sparse, non-informative and delayed reward signals.

Learning RL rewards from human preferences aims to incorporate human feedback in text generation and teach models to follow human instructions \citep{ouyang2022training, chatgpt2022, openai2023gpt4}. Neural reward learning schemes train neural teachers that learn to score an ordered sequence of sentences and formulate rewards that guide coherent long text generation \citep{bosselut2018discourse}; the approach is used for generating cooking recipes given the dish title and the set of ingredients as constraints. Learning-to-rank algorithms are used to approximate ground-truth oracle rewards in extractive multi-document summarization to indicate the quality of a summary or preferences over summary pairs \citep{gao2019reward}. Machine
learnability of human rewards in neural machine translation models is approached by first training reward estimators on rewards collected from offline logs, then integrating these reward estimators in an off-policy RL setting \citep{kreutzer2018reliability}. Similarly, implicit human reactions such as sentiment or length of a conversation are used to learn rewards for fine-tuning off-policy RL models for dialog \citep{jaques2019way}. Nevertheless, human feedback is noisy, not well-defined, complex and inconsistent. Using RL to improve system
outputs with respect to human-centered metrics of conversation quality is highly dependent on  developing robust metrics for the particular application domain, for example increasing politeness or reducing toxicity of generated responses in dialogue generation.

Hard-constrained text generation in a non-monotonic order relies on a  tree-based text generation scheme, where a word is generated at an arbitrary position in the sentence, then binary trees of words to its left and right are recursively generated \citep{welleck2019non}. Learning proceeds in an incremental fashion in an imitation learning framework, where the policy gradually moves from imitating the oracle to reinforcing its own preferences and generating texts without a pre-specified word order. Nevertheless, the time complexity of the approach is $\mathcal{O}(n)$, same as for autoregressive models and the constructed tree does not reflect a high-level to low-level hierarchy of concepts. Constraint satisfaction problems with solutions that can be automatically verified are used to evaluate
how well LLMs adhere to logical constraints\citep{lin2025zebralogic}. Due to token mislalignment, enforcing strict constraints during generation can nevertheless lead to significant decrease in
reasoning performance and downstream accuracy \citep{beurer2024guiding}.

\subsection{Prompting Large Language Models}

Prompt-based learning, which became popular with the release of OpenAI GPT-3 \citep{brown2020language}, demonstrates it is possible to elicit factual and commonsense knowledge from large language models and steer them towards desired behaviours via a textual prompt. Instead of adapting models to downstream tasks via objective engineering as it is common during fine-tuning, prompt-based learning reformulates downstream tasks to resemble those encountered during the language model pre-training phase where a fill-in-the-blanks objective is used \citep{liu2023pre}. While prompting allows for manipulating the model behaviour to predict desired output, sometimes even without additional task-specific training, model performance on a given task is highly dependent on the quality of the prompt used to steer the model and how much conditioning text can fit into the model's input. In general, identifying the most appropriate prompt for a task is a challenge. While prompting provides a natural interface for humans to communicate with machines, human users have little knowledge of which instructions are compatible with a given model and need to experiment with a wide range of discrete prompts to find suitable ones that elicit desired behaviours \citep{zhou2022large}. Given that plain language prompts do not always produce the intended results, automated methods for prompt design are proposed in the literature, including searching over the discrete space of words guided by training data \citep{shin2020autoprompt}, prefix tuning which optimizes a task-specific continuous vector \citep{li2021prefix, hambardzumyan2021warp}, prompt tuning which learns soft prompts via backpropagation \citep{lester2021power, qin2021learning}, natural language prompt engineering where large language models themselves generate meta-prompts for solving a wide range of tasks \citep{reynolds2021prompt, zhou2022large} or inverse prompting which uses the generated text to inversely predict the prompt \citep{zou2021controllable}. Directional Stimulus Prompting \citep{li2023guiding} guides black-box language models such as ChatGPT \citep{chatgpt2022} towards desired outputs by optimizing a policy model trained to maximize rewards that measure the alignment between the generated text and desired topics and keywords on tasks such as text summarization and dialogue response generation. Steering LLMs to generate texts that reflect multiple perspectives and diverse opinions is achieved by first modeling data-driven personas when embedding individuals and their viewpoints into a continuous vector space, then using soft prompting techniques to map persona embeddings to specific tokens \citep{li2024steerability, hwang2023aligning, santurkar2023whose}.

While there is not much theoretical understanding behind the reasons why and how prompting works, it is assumed that prompting provides a way to steer large language models in particular directions by helping locate a specific task in the pre-trained model's existing space of learned tasks, phenomenon evidenced by the superior performance of some prompts over others \citep{reynolds2021prompt}. Nevertheless, prompting large language models is far from sufficient for robust and reliable constrained text generation. Prompting  approaches must be employed with caution, as models can deviate from the original prompt, fail to maintain the coherence and produce
texts on unrelated topics  \citep{zou2021controllable, hernandez2023measuring}, and may even degenerate into toxic text from seemingly innocuous prompts \citep{gehman2020realtoxicityprompts}. In improving prompting reliability, it is important to account for generalization outside of distribution, reducing social biases, ensuring fairness to different demographic groups, calibrating output probabilities and updating the model's factual knowledge and reasoning
chains \citep{si2022prompting, bach2022promptsource}. Adapting to new constraints without the need for model retraining can be done by verbalizing the constraints into natural language instructions, then appending these constraint verbalizations to natural language sentences \citep{zhou2023controlled}.

\textbf{Prompting Considerations} LLMs leverage vast amounts of information they learn from web-scale pre-training datasets which they store in their parameters, resulting in improved performance on many knowledge-intensive tasks \citep{brown2020language, chatgpt2022, openai2023gpt4}. Nevertheless, it is important to understand what kind of knowledge LLMs actually capture. In factuality assessments of LLMs, it is found that current systems tend to hallucinate and make up facts \citep{maynez2020faithfulness, tam2022evaluating, zhou2021detecting, lin2022truthfulqa}, and this behaviour becomes more predominant as the rarity of entities increases \citep{min2023factscore}. There is a strong correlation between the correctness of answering factoid questions and the number of pre-training documents relevant to that question \citep{kandpal2023large}; models are more accurate on instances whose terms are more prevalent in the training data, and struggle on questions containing long-tail terms with low document count. Similarly, mathematical reasoning capabilities are correlated with training data frequency, and the selection of the training corpus does impact the few-shot performance of LLMs \citep{razeghi2022impact, shin2022effect}. These findings suggest that low-order co-occurrence statistics in the pre-training dataset have a significant impact on model performance, leaving the open question of how much current models generalize beyond their training data. Ideally, a general purpose language model can generalize not only to unseen instances of known tasks, but also to new tasks. LLMs tend to rely on narrow, non-transferable procedures for task solving specialized to tasks seen during pre-training \citep{wu2023reasoning}; in counterfactual settings their performance degrades considerably, indicating overfitting to training tasks.

\section{Constrained NLG Evaluation}
\label{survey_nlg_evaluation}

Evaluation of constrained text generation is performed using the same evaluation approaches and methodologies available in the natural language generation literature. In general, evaluation  of the generated text is largely an unsolved and notoriously difficult problem \citep{borji2019pros}. Currently, there is no well-established consensus on how NLG systems should be evaluated, \citep{van2019best, gkatzia2015snapshot}, and the lack of meaningful quantitative evaluation metrics to accurately assess the quality of trained models is detrimental to the progress of the field. In the absence of well established evaluation measures, natural language evaluations are carried in a rather ad-hoc manner with a lot of variability across the proposed models and tasks on inconsistent benchmarks, resulting in misleading performance measures. Subjective evaluations based on visual inspection of the generated samples often lack scientific rigour, making it difficult to quantify and judge precisely the quality of a generative model \citep{hashimoto2019unifying}. In what follows we review the main methods for constrained text generation evaluation.

\paragraph{Lexical constraints} Measuring how many of the given lexical constraints are included in the generated outputs is done using \textit{\textbf{concept coverage}} \citep{lin2019commongen, lu2021neurologic}; the metric is computed as the the average percentage of input concepts that are present in the lemmatized outputs.

\paragraph{Semantic and syntactic constraints} Surface similarity based on \textit{\textbf{n-gram overlap metrics}}, such as BLEU \citep{papineni2002bleu}, ROUGE \citep{lin2004rouge}, METEOR \citep{banerjee2005meteor} measure to what extent the generative model can preserve content by retaining words commonly shared between the generated output and ground-truth references. Such metrics are commonly used to measure response relevance in dialogue systems %
\citep{galley2015deltableu, li2016persona},  
translation quality in neural machine translation  \citep{sennrich2016controlling}, summary quality in text summarization \citep{see2017get}. In general, the correlation between word overlap metrics and true text quality is a widely debated topic \citep{li2016deep}. Evaluation metrics based
on local n-gram patterns only provide a limited, myopic perspective of overall text quality and are notoriously poor at evaluating dialogue systems \citep{liu2016not, see2019makes, bosselut2018discourse}.

\textit{\textbf{Perplexity}} \citep{jelinek1977perplexity} based evaluation metrics are used to evaluate and compare language models, and measure the fluency and diversity of the generated samples \citep{madotto2020plug, bosselut2018discourse, li2016persona}. Reverse Perplexity \citep{zhao2018adversarially} and Forward Perplexity  \citep{kim2017adversarially} scores are calculated by training language models on synthetic samples, respectively real samples, and then using these trained models to measure perplexity real samples, respectively generated samples. Nevertheless, perplexity is a model dependent metric, and ``how likely a sentence is generated by a given model'' is
not directly comparable across different models \textbf{unless properly normalized}; finding the right normalization is a challenge that could potentially improve the evaluation of constrained text generation. Moreover, numerous studies find perplexity to be an inadequate measure of text quality \citep{theis2016note, fedus2018maskgan}, since models with high likelihood can generate low-quality samples, while samples of good quality can present low likelihood. In addition, infinite perplexity can still be obtained from a perfect model even when its ability to generate test sentences is removed  \citep{hashimoto2019unifying}. 

\textit{\textbf{P, R, F1}}  measure the distance of the generated samples to the real data manifold \citep{lucic2018gans}. When precision is high, the generated samples are close to the data manifold; when recall is high, the generator outputs samples that cover the manifold well. Metrics that aggregate precision and recall such as $F_\beta$, a generalization of the $F_1$ score, quantify the relative importance of precision and recall \citep{sajjadi2018assessing}. However, the non-synthetic data manifold is unknown and therefore impossible to compute in practice.

\textit{\textbf{Content diversity}} measures how different the generated sentences are from each other, by either considering word choice, topic and meaning \citep{vijayakumar2016diverse, gimpel2013systematic, ippolito2018comparison}, or by looking at the level of sentence interestingness or unlikeliness \citep{hashimoto2019unifying}. Perplexity on a reference set, $n$-gram diversity \citep{li2016diversity} and Self-BLEU \citep{zhu2018texygen} are commonly used measures of the diversity of the generated samples. In addition, Backward-BLEU \citep{shi2018toward} evaluates test data using the generated samples as reference; the higher the score the more diverse the generator output. Lexical diversity \citep{bache2013text} calculates the ratio of unique tokens to the total number of generated tokens. Similarly, Distinct-$k$ or Dist-$k$ \citep{li2016diversity} measures the total number of unique $k$-grams normalized by the total number of generated $k$-gram tokens to avoid favoring long
sentences. Nevertheless, the Dist-$k$ metric ignores the fact that  infrequent $k$-grams contribute more to
diversity than frequent ones and assign same weight to all $k$-grams that appear at least once. Distinct-1 and Distinct-2 are used to measure the diversity of constrained conversational responses \citep{baheti2018generating, zhang2018generating} and rhetoric constrained generated poems \citep{liu2019rhetorically}. Entropy based metrics such as Ent-$k$ \citep{zhang2018generating} reflect the frequency difference of $k$-grams and to analyze the information content
of the generated responses in dialogue systems \citep{serban2017hierarchical, mou2016sequence}.

Unlike traditional evaluation metrics based on heuristics, learnable metrics train machine learning models on human annotated datasets to learn a scoring function that reproduces human judgements. \textit{\textbf{Fully-learnt metrics}} leverage existing datasets of human ratings to learn automated evaluation metrics that fit the human data distribution, and can be tuned to measure specific properties of the generated texts, such as fluency, style, grammaticality, fidelity, etc. 
Linear regression based on human judgements is used to learn a model for scoring system summaries \citep{peyrard2017learning}. RUSE \citep{shimanaka2018ruse} combines  sentence embeddings in a multi-layer perceptron regressor model. ESIM \citep{chen2017enhanced, mathur2019putting} feeds the encoded representations of the candidate and the reference sentence into a feedforward regressor. BLEURT \citep{sellam2020bleurt} fine-tunes BERT \citep{devlin2018bert} on human ratings datasets for similarity score prediction. MAUDE \citep{sinha2020learning} is proposed for the evaluation of online dialogue conversations and leverages sentence representations from pre-trained BERT to train text encoders which can distinguish between valid dialogue responses and fake examples. BARTScore \citep{yuan2021bartscore} formulates the evaluation of generated text as a text generation task from pre-trained language models and measures the weighted probability of the generated text given another text as input or output. GPT Judge \citep{lin2022truthfulqa} fine-tunes GPT3 \citep{brown2020language} model on human annotated data to clasify answers of QA systems as true or false, evaluating factuality and truthfulness. The same evaluation metric, this time based on GPT-4 \citep{openai2023gpt4}, is used to establish via prompting whether texts generated by GPT-4 are more similar to human-written reference answers or GPT-3 machine-generated texts. Nevertheless, the GPT-4 evaluator is known to be biased in its preferences towards answers with longer length \citep{singhal2023long, wang2023far, li2023alpacaeval}. GPTScore \citep{fu2023gptscore} computes the conditional probability of generating the target text given specific context. FactScore \citep{min2023factscore} breaks generation into atomic pieces of information and evaluates the factual precision of long-form text by measuring the percentage of atomic facts supported by a reliable knowledge source. Other evaluation metrics based on probabilities inferred from pre-trained masked language models include InfoLM \citep{colombo2022infolm}, CTRLEval \citep{ke2022ctrleval}, MaskEval \citep{liu2022maskeval}. \textit{\textbf{Hybrid metrics}} combine learnt elements with human-defined logical rules, for example, contextual embeddings with token alignment rules. BERTscore \citep{zhang2019bertscore} evaluates generated text against gold standard references using soft-string similarity matches (i.e. cosine similarity) computed on pre-trained contextualized BERT \citep{devlin2018bert} token embeddings. MoverScore \citep{zhao2019moverscore} combines contextualized representations of system and reference texts with semantic measures of distance  computed using Word Mover’s Distance \citep{kusner2015word}; the metric is extended to evaluate multi-sentence texts \citep{clark2019sentence}. Human and statistical evaluation are combined in HUSE \citep{hashimoto2019unifying}, an evaluation framework which estimates the optimal error rate of predicting whether a piece of text is human-written or machine-generated. A limitation of learned evaluation metrics is that they often fail to generalize across different systems \citep{chaganty2018price}.

\paragraph{Utility constraints} A commonly used approach in the literature to assess whether generated texts have desirable attributes is to rely on an attribute classifier and measure the \textit{\textbf{classification score}}, i.e. the fraction of outputs generated by the model having the desired attribute \citep{hu2017toward, shen2017style, li2018delete}. \textbf{\textit{Adversarial evaluation}} \citep{bowman2015generating, kannan2017adversarial} employs an evaluator trained to distinguish machine-generated  from human-written texts, analogous to the discriminator in GANs \citep{goodfellow2014generative}. On this note, \textit{\textbf{pre-trained attribute classifiers}} and \textit{\textbf{class-specific discriminators}} measure how well the generated samples match the conditioning labels on attributes such as sentiment, tense, voice, mood and negation \citep{logeswaran2018content, li2017adversarial, bruni2017adversarial}, %
guarantee the accuracy of stylistic text transfer \citep{zhang2018style, shen2017style}, or are used to evaluate biases against certain demographics and quantify model fairness in downstream settings \citep{cao2022intrinsic, mathew2021hatexplain, kurita2019measuring}. %
\textit{\textbf{GLEU}} \citep{napoles2015ground} was originally proposed for grammatical error correction, and later adopted for the evaluation of text style transfer since both tasks require localized edits to the input sentence; GLEU is found to present a reasonable balance between target style match and content retention \citep{sudhakar2019transforming}. 

\textit{\textbf{Readability metrics}} such as Flesch-Kincaid Grade Level %
\citep{kincaid1975derivation} and Flesch Reading Ease \citep{flesch1979write} are used to measure the reading difficulty/simplicity of a piece of text. Both metrics are computed as linear combinations of the number of words per sentence and number of syllables per word with different weighting factors. Although these metrics are fast and easy to compute, they should not be used on their own but in combination with metrics that capture the grammaticality and meaning preservation of the generated output \citep{wubben2012sentence}. In addition, they were not designed for measuring text readability in scientific or specialized domains, and are only available for the English language.

\textbf{All constraints} While automated evaluation helps assess generated texts quickly and cheaply, the use of automated evaluation metrics is dependent upon their correlation with human judgements of text quality \citep{fomicheva2019taking}. \textit{\textbf{Human evaluations}} remain the gold-standard in natural language generation; automated evaluation metrics can be used as a proxy for human judgements only when there is reasonable correlation with human decisions. Ideally, automated evaluations are carried simultaneously with human annotation studies, and not as a replacement of human evaluations. In text style transfer, human evaluations are conducted to determine how accurately constrained text generation methods identify stylistic textual attributes in the source input and replace these with desired target attributes in generated sentences  \citep{sudhakar2019transforming}. In conversational systems, responses generated by open-domain chatbots are evaluated across two dimensions: \textit{i)} humanness, as a proxy for the fluency and coherence of the generated responses, and \textit{ii)} attribute consistency, to determine whether the style and topic enforced by the generation model are well captured \citep{madotto2020plug}. Human evaluations are also carried to determine the plausability of the generated response, its content richness and how much new information it adds to the conversation  \citep{baheti2018generating}. Outputs generated by neural conversational systems are also assessed for quality, style and topic to determine whether the acquisition of styles of famous personalities, characters, or professionals is achievable, and whether the conversational topic can be steered in particular directions \citep{wang2017steering}.

\paragraph{Limitations of current evaluation metrics} Given the wide diversity of evaluation paradigms, it becomes challenging to objectively compare models and research progress when different evaluation metrics are employed in each work. By far, human-quality texts are considered the ground-truth for evaluating the output of NLG systems, serving as an upper bound  measure of their performance. However, collecting human-quality texts and/or soliciting human judgements of text quality is a costly and time-consuming process which requires careful design choices. Often times automated metrics that present reasonable correlation with human evaluations of text quality are used as a proxy for human judgements, however these metrics come with their own limitations. A common complaint is the lack of good ways to encode what constitutes human-quality output in an automated metric \citep{clark2021all}. In addition, shortcomings of current evaluation metrics include poor correlations with human judgements, lack of interpretability of their scores, the presence of complex biases in their evaluations, poor adaptability across tasks and inability to capture nuances\citep{dubois2024length, khapra2021tutorial}. In what follows we discuss limitations of existing evaluation metrics, hoping to inform on the development of more robust evaluations for NLG systems.

\textbf{Word Overlap} metrics measure the lexical overlap between the model generated text and a set of human-written references. Metrics such as BLEU \citep{papineni2002bleu}, ROUGE \citep{lin2004rouge} and METEOR \citep{banerjee2005meteor} allow for fast and inexpensive development cycles and have been widely adopted for evaluating the output of natural language generation systems based on their correlation with human judgements at the time they were introduced, nevertheless their use is not without problems. On the one hand, the choice and quality of references is critical for improving the correlation between human and automated evaluation \citep{freitag2020bleu}. Current evaluation metrics are biased towards assigning higher scores to outputs that share a similar style with the reference, therefore collecting only a single style of references fails to reward systems that produce alternative but equally accurate outputs \citep{popovic2019reducing}; besides, collecting human-written references for new tasks is costly. On the other hand, these metrics assume that valid machine-generated responses present a significant degree of overlap with ground-truth references; this is problematic for open-ended text generation tasks that require diversity and creativity (for eg., dialogue generation), and in such cases their correlation with human judgements is relatively low \citep{liu2023gpteval, liu2016not, graham2019translationese, sellam2020bleurt}. For the evaluation of text simplification, BLEU presents weak or no correlation with grammaticality and meaning preservation for sentence splitting operations, therefore penalizing simpler sentences \citep{sulem2018bleu}.  In addition, improvements in BLEU do not necessarily reflect an improvement in machine translation quality  and there is a huge amount of variation for identically scored hypotheses \citep{callison2006re} (i.e. a wide variety of candidate outputs receive the same score when they present the same degree of overlap with the reference although they greatly vary). Word overlap metrics are also insufficient for measuring factual correctness of text summarization and fail to correlate with human judgements of factuality \citep{falke2019ranking, kryscinski2019neural, pagnoni2021understanding}. Even more concerning is that the great majority of automated metrics, and in particular conventional reference-based metrics such as BLEU \citep{papineni2002bleu} and CIDER \citep{vedantam2015cider}, are found to overrate machine-generated text over human-written text even though the machine text falls short of humans \citep{kasai2022transparent}. In addition, BLEU and ROUGE fail to accurately measure content quality, capture syntactic errors and do not reflect the reliability of NLG systems \citep{reiter2009investigation, stent2005evaluating}.  Using such evaluation metrics to compare systems may lead to drawing inaccurate conclusions, gives the false impression of progress and actively discourages the development of stronger generative models. 

Since BLEU is based on n-gram precision, lexical differences between the hypothesis and references are aggressively penalized even when they are similar or synonymous to the reference. Given that no partial credit is given if an n-gram does not exactly match a sub-sequence of the reference, BLEU is also hard to optimize due to the fact that learning objective is flat and cannot hill-climb through intermediate hypotheses that have  high semantic similarity or synonymy, but low n-gram overlap \citep{wieting2019beyond}. Alternative metrics based on word embeddings are easier to optimize as they output continuous values and capture fine-grained distinctions between similar outputs \citep{wieting2019beyond}. When used for measuring the quality of back-translations for data augmentation, BLEU only shows significant improvements for test examples if the source itself is a translation \citep{edunov2020evaluation}; whenever references are translations and the source itself is natural text, BLEU fails to capture human preference for source original sentences. While the use of multiple references substantially improves reference-based metrics, evaluations are often conducted using a single human-written reference per instance; in such cases strong referenceless metrics frequently achieve higher correlation with human judgements \citep{rei2020comet}. Developing evaluation metrics that correlate well with human judgements on an instance level could serve to augment and validate human annotations.

To overcome the limitations of reference-based evaluation metrics, reference-free natural language evaluators are proposed \citep{fu2023gptscore, wang2023chatgpt}. Simultaneously, evaluating the quality of generated texts based on a form-filling paradigm leverages large language models with chain-of-thoughts \citep{liu2023gpteval}: given a prompt that defines the evaluation task and desired evaluation criteria, the language model generates a chain-of-thought with detailed evaluation instructions based on which it will then score the generated text according to the defined criteria. While LLM-based metrics seem to outperform reference-based and reference-free evaluation metrics in terms of correlation with human judgements for open-ended and creative NLG tasks, they are very sensitive to the instructions and prompts given. Moreover, they tend to prefer LLM-generated texts over high quality human-written texts, which leads to biased predictions especially when used as reward signal for improving themselves. Using language models for ``self-evaluation'' indicates their predictions are well calibrated for token probabilites in-distribution, but they struggle with calibration in settings outside of the data distribution \citep{kadavath2022language}. 

\textbf{Model-Based Evaluation} metrics are becoming increasingly popular for NLG evaluation due to powerful representations learnt by pre-trained language models and high correlations with human judgements of text quality. However, current language models have well-known flaws and limitations, for example they assign high likelihood to degenerate texts, i.e. output that is bland, incoherent, or repetitive \citep{holtzman2019curious}, can be insensitive to perturbations such as word order randomization \citep{pham2021out}, negation \citep{ettinger2020bert} or named entity replacements \citep{balasubramanian2020s}, exploit superficial cues through the use of the self-attention mechanism \citep{pham2021out} and exhibit naive understanding of the meaning of sentences without complex reasoning \citep{ribeiro2020beyond}. To investigate the extent to which model-based evaluation metrics suffer from the same limitations as black-box pre-trained language models, stress tests are used to complement human correlation tests and detect the blind spots of evaluation metrics \citep{he2022blind}. The authors construct a noised hypothesis set by applying different synthetic errors to ground-truth human-written references; if this noised hypothesis set is not scored worse than the original unperturbed set, it means the evaluation metric fails the corresponding stress test. Stress tests reveal that model-based evaluation metrics can be insensitive to errors at the start and middle of the generations when based on pre-trained models that do not encode long-range context, their judgement can be misled by simply injecting valueless text spans into the hypotheses, are biased towards frequent n-grams, present a self-evaluation bias by unfairly ranking generations from their underlying base pre-trained language model higher than better quality generations from larger models, fail fluency tests (lemmatizing verbs, removing articles, prepositions or tokens at the end of the hypothesis) and consistency tests (sentence switching, replacement or negation). Complex biases, including a strong preference for longer outputs, are a common issue when relying on LLMs models to estimate response quality \citep{li2023alpacaeval, dubois2024length}. While some model-based metrics perform better than others, it is important to recognize their limitations and use each metric with awareness of its blind spots. To  mitigate the risks of drawing inaccurate conclusions based on a single metric,  using combinations of evaluation metrics that cover each other's blind spots is recommended. For example, evaluation metrics based on pre-trained language models that encode long range context could be more robust to errors in the beginning or middle of the generations, valueless text span injections can be identified by word-overlap based metrics such as ROUGE \citep{lin2004rouge}, biases towards frequent n-grams can be detected by using diversity metrics, and truncation errors can be recognized via precision, recall and F1 scores. To mitigate unfair biases, it is desirable to avoid using the same pre-trained language model for  generation as well as base for the evaluation metric, or comparing different pre-trained models using an evaluation metric that relies on one of these models. Finally, adding explainability on top of black-box evaluation metrics can help identify system quality issues and increase trust in the evaluation of NLG systems \citep{leiter2022towards}.

\textbf{Human Evaluation} is considered the gold standard for the evaluation of NLG systems, however there is no consensus on how these human studies should be conducted \citep{gkatzia2015snapshot, van2019best}. The large variability in the design of human evaluations leads to difficulty in comparing results across different studies and also impacts the reliability of the inferred conclusions. The lack of consistency in human evaluation can be attributed to different factors such as the level of expertise of human annotators, their cognitive biases, ambiguity of the annotation task itself, or the actual wording of questions and instructions presented to participants (``how something is asked as opposed to what is asked'') \citep{schoch2020problem}. Untrained human evaluators may provide inconsistent results and contradictory reasons behind their judgments: ``all that's human is not gold'' \citep{clark2021all}. Unsurprisingly, selecting a different subset of annotators can lead to different conclusions due to variations in individual annotators' understanding of the annotation scheme \citep{amidei2020identifying}. In general, it is hard to decompose, interpret and validate crowdworker evaluations \citep{kasai2022transparent}. Depending on the evaluation setup, it may be sensible to use qualified evaluators who have gone through extended training and can provide more reliable annotations. Moreover, improving the robustness and transparency of human evaluation guidelines is essential for increasing the reliability of human annotations. As the fluency of generated texts is improving, it is important to not only focus on surface-level aspects of text quality in human evaluations, but also to assess the informativeness and usefulness of generated texts in downstream settings \citep{clark2021all}.

\textbf{Future Outlook} While so far we have reviewed limitations of existing evaluation metrics, we would also like to note the metrics that are missing or are under-represented in the literature, particularly metrics for measuring the trustworthiness, factuality, fairness, bias, toxicity, efficiency, diversity, uncertainty quantification, calibration and robustness of text information systems. In addition, it is important for the community to focus on the interpretability aspect of evaluation metrics, particularly for model-based evaluations that currently function as a black-box \citep{leiter2022towards}. In the era of large language models, aspects such as knowledge, reasoning, memorization/copyright and disinformation are becoming increasingly important to quantify and analyze for NLG systems \citep{liang2022holistic}. Special attention also needs to be paid to existing datasets used to evaluate the generalization abilities of state-of-the-art methods to ensure there is no overlap between the train set and the test set; in such cases, evaluations inadvertently measure memorization instead of the model's ability to generalize, giving the false impression of improvements in performance  \citep{elangovan2021memorization}. Large language models in particular are known to memorize parts of their training data, phenomenon which becomes more predominant with increasing the model capacity and the repetition of training examples   \citep{carlini2022quantifying, razeghi2022impact}. On top of this, given that LLMs are trained on web-scale datasets, evaluations are subject to potential data contamination issues \citep{wu2023reasoning, dodge2021documenting, magar2022data}. Therefore, interpreting evaluation results must be done with caution accounting for the source pre-training data in determining to what extent current models generalize vs simply memorize training examples \citep{razeghi2022impact}; this also highlights the need to reconsider and redefine evaluation schemes for LLMs, and focus on debiasing current evaluation metrics \citep{li2023alpacaeval, dubois2024length}. Finally, it is important to consider how advances in generative models can benefit and inform the development of more suitable evaluation techniques, and vice versa. Bidimensional leaderboards \citep{kasai2022transparent} that simultaneously track progress in language generation models and evaluation metrics can bridge the gap between generation modeling and evaluation research. As generation models continue to improve, it is important to keep reassessing and updating evaluation metrics so that they accurately reflect the target objectives and correlate with human language use in the real world \citep{zellers2021turingadvice}.

\section{Constrained NLG Benchmarks and Datasets}

Datasets that capture a wide diversity of constraints and are representative of many real world situations are critical for advancing safe and robust constrained text generation. Existing benchmarks focused on politeness \citep{madaan2020politeness}, formality \citep{rao2018dear}, sentiment \citep{shen2017style}, writing style \citep{jhamtani2017shakespearizing} are rather limited in nature and do not offer fine-grained control over stylistic attributes. StylePTB \citep{lyu2021styleptb} aims to allow compositional transfer over a wider range of fine-grained stylistic constructs, including lexical, semantic, stylistic and thematic transfers.

CommonGen \citep{lin2020commongen} benchmark proposes the task of constrained text generation with generative commonsense reasoning, where given a set of concepts the task is to generate a coherent sentence describing an everyday scenario using the given concepts. To do this successfully, the generative model must reason over commonsense relations between the given concepts (relational reasoning), and infer novel combinations of familiar concepts (compositional generalization). Preliminary analysis shows that current state-of-the-art pre-trained models struggle at the task and generate implausible sentences by a large margin. Other benchmarks proposed in the literature focus on avoiding model hallucinations and assessing the veracity and factuality of current models \citep{hendrycks2020measuring, bhakthavatsalam2021think, talmor2019commonsenseqa}. TruthfulQA \citep{lin2022truthfulqa} benchmark is proposed for measuring the factual accuracy and truthfulness of QA systems. Surprisingly, in their preliminary experiments the authors find that larger language models are less truthful than smaller language models from the same family, neverthleless they are more informative. RealToxicityPrompts \citep{gehman2020realtoxicityprompts} aims to measure the extent to which toxic degeneration of large language models can be avoided, and the effectiveness of steering text generation algorithms away from producing racist, sexist and toxic content. Ideally, we want to have NLG models that are controllable, truthful, informative and perform well in the real world, however current pre-trained large language models can degenerate into toxic texts even from seemingly innocuous prompts. Instruction-Following Eval (IFEval) \citep{zhou2023instruction} for large language models aims to measure to what extent LLM models can generate texts that satisfy specific lexical, format and utility constraints, such as for example ``write in more than 400 words'' and ``mention the keyword of AI at least 3 times''. However, performance of current models on existing benchmarks is not necessarily representative of their real-world performance. This issue is amplified by the use of biased automated evaluators, for example towards models that generate longer outputs \citep{li2023alpacaeval, dubois2024length}. The research community not only needs better evaluation metrics (as outlined in Section \ref{survey_nlg_evaluation}), but also better benchmarks. Given the fragility of current NLG benchmarking practices, fallacious interpretations can be derived \citep{dehghani2021benchmark}. To minimize the discrepancy between model performance on a given benchmark and its actual usefulness when deployed in real-life situations, benchmarks used for assessing the capabilities of current NLG systems should accurately reflect the end task of interest, as well as the wide diversity of scenarios and constraints encountered in practice. Motivated by the observation that new advances in metrics and models should more directly inform and benefit each other, bidimensional leaderboards \citep{kasai2022transparent} are proposed to track progress in both generative models and evaluation metrics for constrained text generation tasks such as machine translation, text summarization and image captioning. FollowBench \citep{jiang2023followbench} evaluates the ability of LLM models to follow instructions with fine-grained constraints. Single constraints are added incrementally to the initial instruction, allowing to estimate the difficulty level at which models fail to satisfy multiple consecutive constraints. Overall, state-of-the-art LLM models are limited to following instructions with at most several constraints, which illustrates the difficulty of the multiple-constraint satisfaction problem and suggests there is significant potential for further improvement. In addition, only few instructions can be fully verified objectively and automatically, since edge cases make it hard to determine
if an instruction is followed \citep{zhou2023instruction}. Using a diverse pool of atomic, verifiable instructions with constraints that are relevant to real-world applications can help enhance the clarity and objectivity of the constrained NLG evaluation process on the proposed benchmarks. Factuality benchmarks \citep{jacovi2025facts, chen2023felm, muhlgay2024generating, iqbal-etal-2024-openfactcheck} aim to evaluate the factual accuracy of the generated responses in information-seeking scenarios.  Enhancing LLM factuality requires finding a delicate balance as it can compromise other desirable attributes, for example creativity and
novelty. Factuality is expected to remain a research challenge for the foreseeable future, particularly in long-form text generation tasks. Cognac \citep{chen2022controllable} measures whether LLMs conform to lexical level constraints by guiding models on what topics to generate, while also imposing knowledge-intensive constraints on what aspects the model should not to generate. Prompt-based approaches show a lot of promise in steering instruction-tuned LLM models away from generic outputs for stylistic tasks, but tend to perform less well for lexical and format constraints  \citep{ashok2024controllable, sun2023evaluating}.  

A larger issue in terms of natural language evaluation is the gap between how humans use language in the real world, and what current benchmarks can measure \citep{zellers2021turingadvice}. In addition, many datasets are not an effective indicator of model generalization and real world performance, particularly in the presence of overlap between the train and test sets, leading to inflated evaluation results \citep{elangovan2021memorization, dodge2021documenting}. Besides, since massive web-based datasets used to train large language models are often ``contaminated'' with downstream test sets, it is important to conduct in-depth analyses to disentangle genuine progress in natural language understanding/generalization from rote memorization \citep{magar2022data}; overlooking the impact of pre-training data can result in misleading interpretations of model performance \citep{razeghi2022impact}. Finally, we would like to draw attention on the lack of resources (datasets and evaluation metrics) for many languages other than English. 

In summary,  we outline below the reasons behind the mismatch between constrained NLG benchmarks and real-world use case scenarios:

\begin{itemize}
\item LLM models are trained on web-scale datasets with minimal curation to ensure data quality. Many pre-training datasets also contain various evaluation benchmarks that are used for assessing and comparing trained models. Due to these factors, interpreting evaluation results must be done with caution accounting for the source pre-training data in determining to what extent current models generalize vs simply memorize training examples. Auditing LLM models for test set contamination via statistical significance tests reveals verbatim contamination, i.e. LLM models are trained directly on the test sets they are evaluated on \citep{oren2023proving}. Because of this, LLM performance is likely overestimated on existing benchmarks and their behavior in practice is likely much inferior to reported results. In the context of constrained text generation, even if part of the benchmarks is revealed in training, it will significantly simplify the task as the constraints are no longer only testable on the model outputs (in other words, the model can find a shortcut by mimicking part of the training data that already satisfies these constraints). In addition, the inflated evaluation results on benchmarks are not an effective indicator of model generalization abilities and their real world performance. Indeed, recent work shows that emergent abilities of LLMs on many benchmarks are a mirage \citep{schaeffer2024emergent}, appearing due the researcher’s choice of evaluation metric rather than due to fundamental changes in model behavior with scale.

\item Current benchmarks are not representative of the diversity of real-world use cases and constraints, and often only have limited data coverage they are evaluating models on. Certain tasks such as question-answering are overly used in bechmarking LLMs. Due to lack of holistic evaluations, it may appear that LLMs perform well in controlled environments, however they fail in critical real-world applications and constraints, posing safety risks such as perpetuating bias, making unsafe decisions, or being vulnerable to manipulation and adversarial attacks \citep{williams2024targeted, dong2024attacks, shayegani2023survey}. In addition, recent analysis of state-of-the-art LLM benchmarks finds that they suffer from significant limitations, including biases, difficulties in measuring genuine reasoning, adaptability, implementation inconsistencies, prompt engineering complexity, lack of evaluator diversity, and the overlooking of cultural and ideological norms \citep{mcintosh2024inadequacies}.

\item Typically, benchmarks evaluate model performance across one single dimension and summarize results in the form of a single scalar value; this not only offers an incomplete picture of the model performance, but is also misleading when used to compare across models and rank model submissions. Besides, single-value benchmarks can often lead to “reward-hacking” and exploiting spurious features, such as annotators’ preference for more verbose responses \citep{sorensenposition}.

Going beyond monistic benchmarks that measure model performance on a single target objective, it is important to evaluate model performance on multiple constraints. Pluralistic benchmarks with more than one target objective to maximize aim to capture the entire spectrum of model performance across different attributes, making it feasible to compare and rank different models across multiple dimensions \citep{sorensenposition} . Pluralistic benchmarks can be categorized into: \textit{i) multi-objective benchmarks}, reporting evaluations across all objectives for all solutions; \textit{ii) trade-off steerable benchmarks}, designed to measure steerability of models and encourage models to trade off between different objectives at inference time, and \textit{iii) jury-pluralistic benchmarks} which model diverse human ratings, and allow to explicitly reason over which users or groups models are being aligned to for more fair outcomes.
In our view, multi-objective benchmarks and trade-off steerable benchmarks are particularly important for further advancing multi-objective constrained NLG.
\end{itemize}

In general, there is a lack of research consensus on how to properly benchmark models and measure scientific progress. %
In-depth analysis of inadequacies of LLM benchmarks reveals significant limitations, including biases, difficulties in measuring genuine reasoning, lack of adaptability, implementation inconsistencies, prompt engineering complexity, limited evaluator diversity, overlooking cultural and ideological norms \citep{mcintosh2024inadequacies}. Moving away from evaluations on static benchmarks to dynamic behavioral profiling, adopting standardized methodologies, regulatory certainties and ethical guidelines should be prioritized.}
We encourage more research in these directions to bridge the gap between current constrained NLG evaluations and the model performance in real-world settings.

\section{Discussion}

In what follows we summarize the main challenges for constrained NLG and outline open problems, then we present the most promising research directions in the authors' opinion for advancing the state-of-the-art for safe and reliable constrained NLG.

\subsection{Open Challenges}

In our view, constrained text generation is a more difficult problem compared to other instances of text generation. The difficulty arises from a multitude of factors, including lack of model expressiveness which makes it difficult for current models to incorporate constraints into the objective function, lack of suitable evaluation metrics to assess the extent to which constraints are satisfied (which becomes even more challenging in the presence of multiple constraints), difficulty in the constrained optimization of non-differentiable reward functions, and finally lack of constrained text generation datasets that are illustrative of a wide diversity of constraints. Due to these pressing issues, constrained text generation remains an open challenge in the research community. Advancing the state-of-the-art requires considerable collective and focused effort.

\textbf{Multiple constraint satisfaction} Most approaches proposed for constrained text satisfaction focus on generating sentences that meet one single desired constraint, nevertheless generating sequences that simultaneously satisfy multiple lexical constraints is an important open research problem in text generative models \citep{liu2019bfgan, latif2020backward, hsieh2021enconter}. While incorporating one constraint is already hard enough due to lack of model expressiveness, incorporating multiple constraints poses significant challenges in terms of defining the loss function accounting for all the desired constraints, difficulty in optimizing it and evaluating whether each constraint is satisfied.
Approaches that convert the multiple constraint satisfaction problem into allowing the inclusion of pre-specified lexical constraints at decoding time are not optimal either: on the one hand, decoding complexity increases exponentially or linearly in the number of constraints, and on the other hand forcing constraints at every step of the generation process impacts the quality and naturalness of  generated texts \citep{post2018fast}. Moreover, many model architectures are designed for sequential sentence generation only (vs. non-monotonic text generation) and it is non-trivial to impose decoding time constraints while maintaining optimal text generation quality \citep{miao2019cgmh}. 

Prompting methods are used to evaluate to what extent multiple constraints are satisfied by instruction-tuned LLM models \citep{jiang2023followbench}. Single constraints are added to instructions in an incremental fashion,  allowing to  estimate the upper limit of instruction following capabilities in LLMs and assess the difficulty
level at which models fail to follow instructions with multiple constraints. Overall, the more constraints are added to an instruction, the more rapid the decrease in performance of state-of-the-art LLM models; on average, at most three constraints are satisfied. Constraints such as role-playing, reasoning in complex situations, numerical planning, suggestion generation, recognizing and following patterns are identified as the most difficult constraints to satisfy. Prompting with multiple, verifiable, fine-grained constraints is proposed for assessing discrepancies/misunderstandings in following instructions that lead to unintended outputs \citep{zhou2023instruction, sun2023evaluating}. Making constrained NLG evaluations relevant to real-world applications is crucial for improving the reliability of conclusions drawn from current benchmark evaluations.

\textbf{Dynamically defined constraints} Current approaches to constrained text generation assume there is prior knowledge of the constrained textual attributes and the finite set of values these attributes can take. Nevertheless, there are situations when it may be desirable to impose constraints dynamically, for eg. in conversational systems depending on the system user's statements, reactions and emotions. When dynamically defining constraints, the main challenges are the lack of model expressiveness and robust ways to evaluate whether these constraints are satisfied. In the literature, controling the realization of a sentence based on another's sentence syntax and semantics is a less explored setting for constrained text generation with dynamic constraints which does not require prior knowledge of all the values the control variable might take on \citep{chen2019controllable}. Disentangled latent space representations of syntax and semantics are essential for the manipulation sentence attributes in tasks such as unsupervised paraphrase generation and syntax-transfer generation \citep{bao2019generating}. In the context of LLM models, handling dynamic and complex application constraints remains challenging and relatively under-explored area of research. Commonly used solutions for incorporating dynamic constraints leverage the reasoning and planing capabilities of LLM models via strategies such as model fine-tuning and reflection-based reasoning, for example Chain--of-Thought \citep{wei2022chain}, Tree-of-Thoughts \citep{yao2024tree} or Self-Play fine-tuning \citep{chen2024self}. Nevertheless, these approaches typically address constraints on a case-by-case basis, limiting their generalizability. Open questions are how to represent constraints for
LLMs effectively, guide them to reason within those constraints,
and accurately assess the correctness or fallacies in their reasoning process \citep{wei2024optimizing}.

\textbf{Generative reasoning} Current large-scale text generation models display impressive ability to generate fluent texts, nevertheless composing realistically plausible sentences in the presence of constraints remains a significant open challenge. This is illustrative of all challenges associated with constrained text generation, including lack of model expressiveness, lack of suitable evaluation metrics, difficulty in constrained optimization and lack of constrained text generation datasets. Endowing generative models with commonsense reasoning abilities is an important milestone towards advancing machine understanding and intelligence. In general, the great majority of models proposed in the literature only exploit superficial cues via self-attention to solve NLP tasks, without relying on syntactic information or complex reasoning \citep{pham2021out}. 

LLM models trained with reinforcement learning can perform complex reasoning tasks by thinking through a problem, producing a series of intermediate reasoning steps that allow to recognize and correct mistakes before attempting to give the final answer. Prompt-based reasoning with LLMs has lead to rapid advancements on many constrained NLG tasks, including mathematical, logical and commonsense reasoning problems, code generation, question answering, text summarization, machine translation, etc.   \citep{plaat2024reasoning, liu2025logical}. Despite the strong performance of
LLMs on certain reasoning tasks, the extent to which LLMs are actually capable of  reasoning and handling complex deductive problems (vs. exploiting superficial cues and shallow patterns in the data) remains uncertain \citep{huang2023towards, hosseini2024not}. Compared to human performance, LLMs multi-step logical and commonsense reasoning performance is lagging behind humans in various real-world domains such as murder mysteries, object placement or team assignment \citep{spraguemusr}.  Significant decline in LLM reasoning performance is reported as the complexity of constraint satisfaction problems increases, a phenomenon referred to as ``the curse of complexity for reasoning'' \citep{lin2025zebralogic}. Scaling up the number of reasoning tokens generated during inference, and ensuring the correctness and verifiability of the reasoning chain may alleviate the issue to some extent.

\textbf{Attribute specific datasets} The lack of annotated datasets for attribute specific text generation constitutes a bottleneck in the development and adaptation of models for tasks that require fine-grained control over style and topics. For example, in dialogue systems the absence of attribute annotated conversational datasets that can be used for fine-tuning large scale pre-trained models limits control over the generated responses for a desired attribute \citep{madotto2020plug}. Moreover, such attribute annotated datasets can help with the personalization of dialogue systems, make dialogues safe, supportive and engaging \citep{serban2015survey,zhang2018personalizing, ge2024scaling}. Personalized dialogue agents that display consistent personalities and viewpoints overcome the unsatisfying experience of a persona-free chit-chat
model and empower practical applications such
as personalized conversations. Nevertheless, imposing conversational goals on a dialogue agent for learning target-guided strategies requires keyword-augmented conversation datasets for learning how to steer the conversation towards a designated target subject \citep{tang2019target}.

\textbf{Rule constraints} While most research that is currently trying to address constrained text generation is focusing on the incorporation of pre-defined utility or lexical constraints to various degrees of success on simple tasks with narrow scope \citep{ashok2024controllable, sun2023evaluating, zhang2023survey}, the satisfaction of rule based constraints is equally relevant, particularly when used to define format and syntactic conditions on the output. However, the lack of model expressiveness makes it challenging to incorporate rule based constraints into the loss function at training time. We encourage more effort in this direction likely to open a plethora of new possibilities in how constraints are specified, incorporated and satisfied in models particularly designed for constrained neural text generation. 

Prompt-based control for lexically constrained text generation remains challenging for the following reasons: \textit{i)} current LLM models tend to display \textit{position bias} \citep{liu2023lost}, and only satisfy lexical constraints that appear within specific positions in the input, \textit{ii)} \textit{decoding parameters lack of sensitivity} to incorporate lexical constraints, and \textit{iii)} \textit{complexity of compound word constraints}, which are often misinterpreted or altered in meaning by LLM models. Overall, LLMs
struggle to adapt to increasingly complex lexical
constraints with prompt-based control \citep{li2024control}; the more lexical constraints are added to the prompt, the more significant the decrease in performance of LLM models for constrained text generation \citep{jiang2023followbench}. In addition, there is an inherent trade-off between generating
text of high quality and satisfying hard constraints  \citep{iso2024autotemplate}. LLMs often struggle to meet fine-grained hard constraints \citep{sun2023evaluating}, and evaluations on out-of-distribution constraints aiming to differentiate constraint-following
abilities from over-fitting (IFEval \citep{zhou2023instruction} vs. IFEval-OOD \citep{lambert2024t}) report that a lot of the claimed success of prompts targeted for constrained instruction following may be simply attributed to overfitting.

\textbf{Evaluation of constrained text generation} In general, evaluation of text generative models is an open challenge. The field is missing robust automated evaluation metrics that correlate with human judgements across multiple dimensions of text quality. Evaluation of models for constrained text generation is currently done using the same flawed existing metrics commonly used in unconditional and conditional text generation evaluation, or in an informal way often times in the absence of a rigorous evaluation procedure. Human evaluation remains the gold standard way to assess text quality, however designing evaluation metrics tailored specifically at assessing whether generated texts meet desired constraints altogether with new benchmark datasets for the evaluation of constrained text generation are important next steps \citep{latif2020backward, ruan2024better, chen2023comprehensive, zhou2022deconstructing, hu2024llm}. 

\textbf{Adversarial Attacks} Adversarial examples exploit vulnerabilities in text generation models and represent an active research area. Adversarial triggers in the form of input-agnostic sequences of tokens concatenated to any input dataset can trigger a pre-trained language models to produce biased, racist and discriminatory outputs even when these models are carefully fine-tuned and optimized against adversarial triggers  \citep{wallace2019universal}.  Gradient-based adversarial trigger phrase search techniques are used to generate input prompts to a pre-language model that induce biases in the generated output and allows to study strategies for bias mitigation \citep{sheng2020towards}. Constrained text generation models that are robust to adversarial attacks are needed for the beneficial use of machine learning and artificial intelligence technology in real world applications, as well as to mitigate any potential societal harms and biases associated with the deployment of large language models \citep{chowdhury2024breaking, gallegos2024bias}. 

Other important open challenges include the use of constrained text generation for personalized  agents in a wide variety of contexts \citep{zhang2024personalization, liu2025surveypersonalizedlargelanguage}, such as in dialogue settings \citep{zhang2018personalizing}, and new benchmark datasets that are reflective of real-world constraints for both training/fine-tuning and evaluating constrained text generation models \citep{ziyu2023through, mcintosh2024inadequacies, xu2024benchmarking}.

\subsection{Promising Research Directions for Advancing Constrained NLG}

Despite the many open challenges, we believe there are promising approaches in the literature that merit special attention for advancing constrained NLG. We present these below.

\textbf{Reinforcement Learning from Human Feedback (RLHF)} is the predominant paradigm for aligning LLMs to human preferences given helpfulness, harmless and safety constraints \citep{ouyang2022training}. RLHF performance is strongly dependent upon the quality of the reward model, and defining rewards for real-world tasks, especially with the presence of constraints, is non-trivial. Reward functions are fragile and notoriously difficult to specify, particularly for tasks with complex goals \citep{mckinney2023fragility}. An outstanding RLHF challenge is the issue of reward hacking, where LLM policies learn to exploit failures of the reward model and achieve seemingly high rewards without meeting the underlying objectives \citep{rame2024warm}.

Despite these challenges, RLHF can play a key role in advancing constrained NLG, conditioned on the design of more robust and reliable reward functions. To mitigate reward hacking in particular, it is necessary to have reward models that can reliably score generations despite distribution shifts, are robust to label noise and inconsistencies in human preferences. Preliminary approaches that explore prediction ensembling \citep{christiano2017deep} or weight averaging \citep{rame2024warm} of multiple reward models are designed to act as regularization preserving only those mechanisms that are invariant across runs, helping reduce reliance on spurious features and memorization of corrupt/noisy training examples. While these approaches may help delay reward hacking to some extent, they do not fully solve the problems of reliability under distribution shifts and robustness to noisy labels. We believe there is a lot of space to explore more efficient solutions for training robust reward models for real-world constraints that accurately reflect human preferences, entirely prevent reward hacking (instead of just delaying it), and generalize to out-of-distribution settings. 

RL policies that perform well for diverse reward functions (not just one reward model) can accommodate diverse user preferences and advance NLG with multiple constraints. Multi-objective reinforcement learning (MORL) algorithms can be used to learn Pareto-optimal policies and control the learnt policies accounting for multiple objectives \citep{liang2024robust}. For example, multi-objective reward modeling allows to dynamically control the trade-off between diverse user preferences via arithmetic operations in the vector reward space \citep{wang2024arithmetic}. In many real-world settings it may be difficult to accurately specify constraints mathematically; in such situations, it may be possible to learn constraints directly from user provided demonstrations, even when the reward function is unknown \citep{lindner2024learning, malik2021inverse}.

\textbf{Mechanistic Interpretability} Recent works aiming to understand the inner workings of LLMs find that these models have internal representations that encode concepts in a disentangled manner. If one can identify which part of the representation subspace corresponds to a given concept, then it its possible manipulate the concepts expressed by the model through algebraic manipulation of the representation \citep{wang2024concept}. The linear representation hypothesis \citep{parklinear} posits that high-level concepts are represented linearly as directions in the representation space. Assuming it is possible to identify these linear concept representations, linear algebraic operations can be performed on the representation space for fine-grained control of LLM outputs. Interventions using steering vectors that change the value a concept takes without changing other concepts show that in carefully designed test cases it is possible to change, for eg. the output from English to French by adding a suitable English/French steering vector. If the linear representation hypothesis were to hold true, this could potentially open up new methods that advance constrained NLG. Open problems are the identifiability of learned representations, to what extent they capture real-world structure, and what assumptions need to be made about the geometry of the representation space.

\textbf{Causal Interventions / Causal Probing / Inference Time Interventions} 
Recent work shows that it is possible to perform direct manipulation of computational mechanisms inside LLMs. In particular, causal tracing approaches first identify neuron activations corresponding to a particular concept, then edit the corresponding weights to change model outputs in a desirable way \citep{meng2022locating, meng2022mass}. However, the connection between causality-based localization and model editing is still unclear, as localization performed by causal tracing is not indicative of which layer to select for model editing \citep{hase2024does}. Developing more reliable causal mechanisms for localizing where knowledge is stored inside a neural network, as well as robust ways to edit the internal knowledge of LLM models to ensure generated outputs satisfy given constraints, can pave the way for advancing constrained NLG.

Causal probing can be used to analyze how intervening on latent properties of the model’s representation have an impact on the model outputs \citep{canby2024measuring}. For example, linear probing has been used to predict whether the answer will be truthful or not before it is actually generated \citep{joshi2023personas}. While measuring the effectiveness of probing interventions in LLM models is an open research area, causal probing is a promising direction to explore whether constraints are present in the generated output before generation even begins. Inference-time intervention (ITI) \citep{li2024inference} locates directions that correspond to specific concepts (for eg., truthfulness) and shifts model activations along these directions at inference time. Potentially integrating ITI with causal probing and intervention mechanisms, and a better understanding of the geometry of the representation space, could help enforce complex attributes/constraints in the output.

\textbf{Reasoning-based approaches / Constrained Decoding approaches / Rejection Sampling}
Graph-constrained reasoning \citep{luo2024graph} aims to connect the unstructured reasoning in LLMs with the structured knowledge found in knowledge graphs (KG); the model constrains the LLM decoding process to reasoning paths that encode KG information. Generating KG-grounded paths helps alleviate reasoning issues due to lack of knowledge, mitigates hallucinations and enhances the faithfulness of generated responses. Constrained chains of reasoning \citep{lin2024constrained} leverage domain knowledge and the causal relations between concepts to construct reasoning chains that improve consistency of the generated responses.

Constrained decoding approaches enforce adherence to constraints during generation, while (ideally) minimally intervening on other non-target aspects to avoid misalignment \citep{beurer2024guiding}. Guided decoding \citep{lu2021neurologic} employs an auxiliary evaluation function that captures to what extent partial outputs satisfy given goals; the method can be combined with search algorithms to generate outputs that satisfy specific constraints. Combining inference-time search algorithms such as Monte-Carlo Tree Search (MCTS) with RLHF fine-tuned models \citep{liu2024don} demonstrates that value-guided decoding with MCTS is a crucial component for achieving model steerability and constraint satisfaction (for eg., sentiment steering, toxicity reduction, helpful and harmless chatbots). Value models trained as byproducts when aligning LLMs to human preferences have only recently been employed as evaluation functions for scoring partial/incomplete sequences and steering LLM models; it would be interesting to use MCTS as a policy optimization operator to search for contrained sequences with high rewards. 
Overall, we believe that combining reasoning-based approaches with guided decoding and MCTS has a lot of potential to improve the state-of-the-art for constrained NLG, including more faithful, consistent generations and less hallucinations.

Statistical rejection sampling can be used to discard partial samples that do not meet given constraints; this technique has been widely employed in LLAMA-3 family of models \citep{dubey2024llama}, and is found to improve the alignment with human preferences in constrained optimization settings \citep{liu2023statistical}. Best-of-$N$ rejection sampling \citep{stiennon2020learning} draws $n$ samples from the LLM, ranks them on the target attribute of interest, and returns the best sample. Despite its simplicity, this strategy has been found to be surprisingly effective in practice, however Best-of-$N$ sampling comes at considerable inference cost. Approaches trained to mimic this distribution achieve high win-rates while minimally affecting other off-target aspects of the generation \citep{gui2024bonbon}, allowing for better control of LLM models.

\textbf{Improving model architectures} 
The mismatch between how LLM models are trained with a next-word prediction objective and how they are used in practice (for eg., for long-term open-ended dialogue with users) leads to inconsistency in their behaviour over long horizons. The Transformer attention mechanism decays over long exchanges, causing chatbots to stray away from prompted behaviour and resulting in instruction drift that degrades the quality of the outputs over lengthy dialogues \citep{li2024measuring}. Improving instruction stability in LLMs, particularly in long-form conversations, can lead to more stable and robust prompting, improve the performance of current models and their abilities to generate texts accounting for given constraints. %
There is a need for better understanding how LLMs use input context and potentially design novel attention mechanisms that more robustly capture information within long input contexts.

Autoregressive LLMs fail to generalize in surprising ways. The reversal curse \citep{berglundreversal} is one instance where a model trained on a sentence of the form “A is B” will not automatically generalize to the reverse direction “B is A”, therefore failing to deduce the reverse relationship – this directly impacts the LLM models’ ability to generate constrained texts. Further analysis of LLMs generalization using influence functions \citep{grosse2023studying} %
finds that training examples that match the order (“A precedes B”) are far more influential than examples with reverse order (“B precedes A”). Fine-tuning and data augmentation approaches are used to alleviate the issue, however it does point to a basic failure of logical deduction in LLMs training. Incorporating logical deduction and causal-based reasoning during LLM training could help. %

\textbf{Accounting for constraints / Better guidance during training} Classifier guidance \citep{dhariwal2021diffusion} introduces an extra trained classifier to guide diffusion model generations in particular desirable directions. Methods aiming to improve this approach propose classifier-free guidance \citep{ho2022classifier}, showing it is possible to steer LLMs using a pure generative model; a conditional and an unconditional diffusion model are jointly trained, and their score estimates are combined to achieve fine-grained control over the generated outputs. There is a lot of potential to extend the use of such approaches beyond the vision domain to the text domain for more robust constrained NLG.

\textbf{Multi-objective / pluralistic benchmarks}
Aligning LLM models to pluralistic human values requires the capability to accurately steer models in directions representing a diverse set of human values and perspectives \citep{sorensenposition}. Steerable pluralism, i.e. faithfully steering LLMs to represent particular attributes or perspectives, plays a key role in personalizing and customizing models to various users or target populations. Nevertheless, current benchmarks for constrained NLG are monistic, focusing on a single objective. Pluralistic benchmarks have more than one objective to maximize (each objective is measured separately) and allow for more explicit trade-offs between constraints at inference time.

\textbf{Hybrid Human-AI Collaborative Approaches} Interactive systems that allow LLM models and their end users to write collaboratively have the potential to enhance constrained text generation by empowering users to choose which (partial) model generations are in line with their needs and expectations. %
CoAuthor \citep{lee2022coauthor} proposes a collaborative human-AI approach for text generation where a writer and a model take turns interactively in writing a story and editing it. Such methods empower users with fine-grained control over the model outputs and open up new opportunities in assistive writing in a steerable fashion.
Causal inference can play a key role in modeling the human-AI collaboration byanswering counterfactual ``what-if'' questions on how the outcome of the collaboration would change if humans employed a different text editing/refinement strategy. For example, causal estimands (Incremental Stylistic Effect) \citep{zhang2024causal} can be used to measure the average impact of infinitesimally shifting a text towards a specific style, such as increasing the degree of politeness or formality.

\section{Conclusion}

In this work, we have presented the reasons why constrained natural language generation is an important, yet highly challenging and largely unsolved research problem. Our first contribution consists in clarifying the difference between the ambiguous use of unconditional, conditional and constrained terms in the natural language generation literature, and draw clear boundaries between these concepts by exemplifying instances of  natural language generation tasks with their associated conditions and constraints. Among different paradigms of text generation, we consider constrained text generation to be particularly challenging (if not the most challenging), yet also extremely useful. We identify general reasons why constrained natural language generation deserves significant more attention in the research community, including the lack of model expressiveness in incorporating constraints into the objective function at training time, difficulty in constrained optimization algorithms, the lack of suitable evaluation metrics for robustly assessing, comparing model outputs and claiming success in constrained natural language generation, as well as the lack of constrained text generation datasets/benchmarks that are representative of a wide range of real-world constraints for training, fine-tuning and evaluating these models. We then survey a representative body of recent literature on constrained text generation using neural networks, presenting the main approaches and methods used, as well as their limitations. Our work serves as an informative guide for both researchers and practitioners to become familiar with the current methodology and main challenges, in the hope of advancing state-of-the-art constrained NLG. We invite future work in solving the outlined challenges for better, useful, safer and more robust constrained natural language generation and evaluation.

\bibliography{main}

\begin{thebibliography}{384}
\providecommand{\natexlab}[1]{#1}
\providecommand{\url}[1]{\texttt{#1}}
\expandafter\ifx\csname urlstyle\endcsname\relax
  \providecommand{\doi}[1]{doi: #1}\else
  \providecommand{\doi}{doi: \begingroup \urlstyle{rm}\Url}\fi

\bibitem[Achiam et~al.(2023)Achiam, Adler, Agarwal, Ahmad, Akkaya, Aleman, Almeida, Altenschmidt, Altman, Anadkat, et~al.]{openai2023gpt4}
Josh Achiam, Steven Adler, Sandhini Agarwal, Lama Ahmad, Ilge Akkaya, Florencia~Leoni Aleman, Diogo Almeida, Janko Altenschmidt, Sam Altman, Shyamal Anadkat, et~al.
\newblock Gpt-4 technical report.
\newblock \emph{arXiv preprint arXiv:2303.08774}, 2023.

\bibitem[Adolphs et~al.(2023)Adolphs, Gao, Xu, Shuster, Sukhbaatar, and Weston]{adolphs2023cringe}
Leonard Adolphs, Tianyu Gao, Jing Xu, Kurt Shuster, Sainbayar Sukhbaatar, and Jason Weston.
\newblock The cringe loss: Learning what language not to model.
\newblock In \emph{Proceedings of the 61st Annual Meeting of the Association for Computational Linguistics (Volume 1: Long Papers)}, pp.\  8854--8874, 2023.

\bibitem[Amidei et~al.(2020)Amidei, Piwek, and Willis]{amidei2020identifying}
Jacopo Amidei, Paul Piwek, and Alistair Willis.
\newblock Identifying annotator bias: A new irt-based method for bias identification.
\newblock In \emph{Proceedings of the 28th International Conference on Computational Linguistics}, pp.\  4787--4797, 2020.

\bibitem[Amini et~al.(2024)Amini, Vieira, and Cotterell]{amini2024variational}
Afra Amini, Tim Vieira, and Ryan Cotterell.
\newblock Variational best-of-n alignment.
\newblock \emph{arXiv preprint arXiv:2407.06057}, 2024.

\bibitem[Anderson et~al.(2017)Anderson, Fernando, Johnson, and Gould]{anderson2017guided}
Peter Anderson, Basura Fernando, Mark Johnson, and Stephen Gould.
\newblock Guided open vocabulary image captioning with constrained beam search.
\newblock In \emph{Proceedings of the 2017 Conference on Empirical Methods in Natural Language Processing}, pp.\  936--945, 2017.

\bibitem[Ashok \& Poczos(2024)Ashok and Poczos]{ashok2024controllable}
Dhananjay Ashok and Barnabas Poczos.
\newblock Controllable text generation in the instruction-tuning era.
\newblock \emph{arXiv preprint arXiv:2405.01490}, 2024.

\bibitem[Austin et~al.(2021)Austin, Johnson, Ho, Tarlow, and van~den Berg]{austin2021structured}
Jacob Austin, Daniel~D Johnson, Jonathan Ho, Daniel Tarlow, and Rianne van~den Berg.
\newblock Structured denoising diffusion models in discrete state-spaces.
\newblock In \emph{Advances in Neural Information Processing Systems}, 2021.

\bibitem[Bach et~al.(2022)Bach, Sanh, Yong, Webson, Raffel, Nayak, Sharma, Kim, Bari, F{\'e}vry, et~al.]{bach2022promptsource}
Stephen Bach, Victor Sanh, Zheng~Xin Yong, Albert Webson, Colin Raffel, Nihal~V Nayak, Abheesht Sharma, Taewoon Kim, M~Saiful Bari, Thibault F{\'e}vry, et~al.
\newblock Promptsource: An integrated development environment and repository for natural language prompts.
\newblock In \emph{Proceedings of the 60th Annual Meeting of the Association for Computational Linguistics: System Demonstrations}, pp.\  93--104, 2022.

\bibitem[Bache et~al.(2013)Bache, Newman, and Smyth]{bache2013text}
Kevin Bache, David Newman, and Padhraic Smyth.
\newblock Text-based measures of document diversity.
\newblock In \emph{Proceedings of the 19th ACM SIGKDD international conference on Knowledge discovery and data mining}, pp.\  23--31. ACM, 2013.

\bibitem[Baheti et~al.(2018)Baheti, Ritter, Li, and Dolan]{baheti2018generating}
Ashutosh Baheti, Alan Ritter, Jiwei Li, and William~B Dolan.
\newblock Generating more interesting responses in neural conversation models with distributional constraints.
\newblock In \emph{Proceedings of the 2018 Conference on Empirical Methods in Natural Language Processing}, pp.\  3970--3980, 2018.

\bibitem[Bakker et~al.(2022)Bakker, Chadwick, Sheahan, Tessler, Campbell-Gillingham, Balaguer, McAleese, Glaese, Aslanides, Botvinick, et~al.]{bakker2022fine}
Michiel Bakker, Martin Chadwick, Hannah Sheahan, Michael Tessler, Lucy Campbell-Gillingham, Jan Balaguer, Nat McAleese, Amelia Glaese, John Aslanides, Matt Botvinick, et~al.
\newblock Fine-tuning language models to find agreement among humans with diverse preferences.
\newblock \emph{Advances in Neural Information Processing Systems}, 35:\penalty0 38176--38189, 2022.

\bibitem[Balakrishnan et~al.(2019)Balakrishnan, Rao, Upasani, White, and Subba]{balakrishnan2019constrained}
Anusha Balakrishnan, Jinfeng Rao, Kartikeya Upasani, Michael White, and Rajen Subba.
\newblock Constrained decoding for neural nlg from compositional representations in task-oriented dialogue.
\newblock In \emph{Proceedings of the 57th Annual Meeting of the Association for Computational Linguistics}, pp.\  831--844, 2019.

\bibitem[Balasubramanian et~al.(2020)Balasubramanian, Jain, Jindal, Awasthi, and Sarawagi]{balasubramanian2020s}
Sriram Balasubramanian, Naman Jain, Gaurav Jindal, Abhijeet Awasthi, and Sunita Sarawagi.
\newblock What’s in a name? are bert named entity representations just as good for any other name?
\newblock In \emph{Proceedings of the 5th Workshop on Representation Learning for NLP}, pp.\  205--214, 2020.

\bibitem[Banerjee \& Lavie(2005)Banerjee and Lavie]{banerjee2005meteor}
Satanjeev Banerjee and Alon Lavie.
\newblock Meteor: An automatic metric for mt evaluation with improved correlation with human judgments.
\newblock In \emph{Proceedings of the acl workshop on intrinsic and extrinsic evaluation measures for machine translation and/or summarization}, pp.\  65--72, 2005.

\bibitem[Bao et~al.(2019)Bao, Zhou, Huang, Li, Mou, Vechtomova, Dai, and Chen]{bao2019generating}
Yu~Bao, Hao Zhou, Shujian Huang, Lei Li, Lili Mou, Olga Vechtomova, Xinyu Dai, and Jiajun Chen.
\newblock Generating sentences from disentangled syntactic and semantic spaces.
\newblock In \emph{Proceedings of the 57th Annual Meeting of the Association for Computational Linguistics}, pp.\  6008--6019, 2019.

\bibitem[Bengio et~al.(2003)Bengio, Ducharme, Vincent, and Jauvin]{bengio2003neural}
Yoshua Bengio, R{\'e}jean Ducharme, Pascal Vincent, and Christian Jauvin.
\newblock A neural probabilistic language model.
\newblock \emph{Journal of machine learning research}, 3\penalty0 (Feb):\penalty0 1137--1155, 2003.

\bibitem[Berglund et~al.(2024)Berglund, Tong, Kaufmann, Balesni, Stickland, Korbak, and Evans]{berglundreversal}
Lukas Berglund, Meg Tong, Maximilian Kaufmann, Mikita Balesni, Asa~Cooper Stickland, Tomasz Korbak, and Owain Evans.
\newblock The reversal curse: Llms trained on “a is b” fail to learn “b is a”.
\newblock In \emph{The Twelfth International Conference on Learning Representations}, 2024.

\bibitem[Beurer-Kellner et~al.(2024)Beurer-Kellner, Fischer, and Vechev]{beurer2024guiding}
Luca Beurer-Kellner, Marc Fischer, and Martin Vechev.
\newblock Guiding llms the right way: fast, non-invasive constrained generation.
\newblock In \emph{Proceedings of the 41st International Conference on Machine Learning}, pp.\  3658--3673, 2024.

\bibitem[Bhakthavatsalam et~al.(2021)Bhakthavatsalam, Khashabi, Khot, Mishra, Richardson, Sabharwal, Schoenick, Tafjord, and Clark]{bhakthavatsalam2021think}
Sumithra Bhakthavatsalam, Daniel Khashabi, Tushar Khot, Bhavana~Dalvi Mishra, Kyle Richardson, Ashish Sabharwal, Carissa Schoenick, Oyvind Tafjord, and Peter Clark.
\newblock Think you have solved direct-answer question answering? try arc-da, the direct-answer ai2 reasoning challenge.
\newblock \emph{arXiv preprint arXiv:2102.03315}, 2021.

\bibitem[Borji(2019)]{borji2019pros}
Ali Borji.
\newblock Pros and cons of gan evaluation measures.
\newblock \emph{Computer Vision and Image Understanding}, 179:\penalty0 41--65, 2019.

\bibitem[Bosselut et~al.(2018)Bosselut, {\c{C}}elikyilmaz, He, Gao, Huang, and Choi]{bosselut2018discourse}
Antoine Bosselut, Asli {\c{C}}elikyilmaz, Xiaodong He, Jianfeng Gao, Po-Sen Huang, and Yejin Choi.
\newblock Discourse-aware neural rewards for coherent text generation.
\newblock In \emph{NAACL-HLT}, 2018.

\bibitem[Bowman et~al.(2016)Bowman, Vilnis, Vinyals, Dai, Jozefowicz, and Bengio]{bowman2015generating}
Samuel~R Bowman, Luke Vilnis, Oriol Vinyals, Andrew Dai, Rafal Jozefowicz, and Samy Bengio.
\newblock Generating sentences from a continuous space.
\newblock In \emph{Proceedings of The 20th SIGNLL Conference on Computational Natural Language Learning}, pp.\  10--21, 2016.

\bibitem[Brown et~al.(2020)Brown, Mann, Ryder, Subbiah, Kaplan, Dhariwal, Neelakantan, Shyam, Sastry, Askell, et~al.]{brown2020language}
Tom Brown, Benjamin Mann, Nick Ryder, Melanie Subbiah, Jared~D Kaplan, Prafulla Dhariwal, Arvind Neelakantan, Pranav Shyam, Girish Sastry, Amanda Askell, et~al.
\newblock Language models are few-shot learners.
\newblock \emph{Advances in neural information processing systems}, 33:\penalty0 1877--1901, 2020.

\bibitem[Bruni \& Fern{\'a}ndez(2017)Bruni and Fern{\'a}ndez]{bruni2017adversarial}
Elia Bruni and Raquel Fern{\'a}ndez.
\newblock Adversarial evaluation for open-domain dialogue generation.
\newblock In \emph{Proceedings of the 18th Annual SIGdial Meeting on Discourse and Dialogue}, pp.\  284--288, 2017.

\bibitem[Callison-Burch et~al.(2006)Callison-Burch, Osborne, and Koehn]{callison2006re}
Chris Callison-Burch, Miles Osborne, and Philipp Koehn.
\newblock Re-evaluation the role of bleu in machine translation research.
\newblock In \emph{11th Conference of the European Chapter of the Association for Computational Linguistics}, 2006.

\bibitem[Canby et~al.(2024)Canby, Davies, Rastogi, and Hockenmaier]{canby2024measuring}
Marc Canby, Adam Davies, Chirag Rastogi, and Julia Hockenmaier.
\newblock Measuring the reliability of causal probing methods: Tradeoffs, limitations, and the plight of nullifying interventions.
\newblock \emph{arXiv preprint arXiv:2408.15510}, 2024.

\bibitem[Cao et~al.(2022)Cao, Pruksachatkun, Chang, Gupta, Kumar, Dhamala, and Galstyan]{cao2022intrinsic}
Yang~Trista Cao, Yada Pruksachatkun, Kai-Wei Chang, Rahul Gupta, Varun Kumar, Jwala Dhamala, and Aram Galstyan.
\newblock On the intrinsic and extrinsic fairness evaluation metrics for contextualized language representations.
\newblock In \emph{Proceedings of the 60th Annual Meeting of the Association for Computational Linguistics (Volume 2: Short Papers)}, pp.\  561--570, 2022.

\bibitem[Cao et~al.(2018)Cao, Wei, Li, and Li]{cao2018faithful}
Ziqiang Cao, Furu Wei, Wenjie Li, and Sujian Li.
\newblock Faithful to the original: Fact aware neural abstractive summarization.
\newblock In \emph{Thirty-Second AAAI Conference on Artificial Intelligence}, 2018.

\bibitem[Carlini et~al.(2022)Carlini, Ippolito, Jagielski, Lee, Tramer, and Zhang]{carlini2022quantifying}
Nicholas Carlini, Daphne Ippolito, Matthew Jagielski, Katherine Lee, Florian Tramer, and Chiyuan Zhang.
\newblock Quantifying memorization across neural language models.
\newblock In \emph{The Eleventh International Conference on Learning Representations}, 2022.

\bibitem[Chaganty et~al.(2018)Chaganty, Mussmann, and Liang]{chaganty2018price}
Arun Chaganty, Stephen Mussmann, and Percy Liang.
\newblock The price of debiasing automatic metrics in natural language evalaution.
\newblock In \emph{Proceedings of the 56th Annual Meeting of the Association for Computational Linguistics (Volume 1: Long Papers)}, pp.\  643--653, 2018.

\bibitem[Chan et~al.(2019)Chan, Kitaev, Guu, Stern, and Uszkoreit]{chan2019kermit}
William Chan, Nikita Kitaev, Kelvin Guu, Mitchell Stern, and Jakob Uszkoreit.
\newblock Kermit: Generative insertion-based modeling for sequences.
\newblock \emph{arXiv preprint arXiv:1906.01604}, 2019.

\bibitem[Chen et~al.(2022)Chen, Li, Chen, and Narasimhan]{chen2022controllable}
Howard Chen, Huihan Li, Danqi Chen, and Karthik Narasimhan.
\newblock Controllable text generation with language constraints.
\newblock \emph{arXiv preprint arXiv:2212.10466}, 2022.

\bibitem[Chen et~al.(2024{\natexlab{a}})Chen, Zhu, Chen, Soselia, Zhou, Goldstein, Huang, Shoeybi, and Catanzaro]{chen2024odin}
Lichang Chen, Chen Zhu, Jiuhai Chen, Davit Soselia, Tianyi Zhou, Tom Goldstein, Heng Huang, Mohammad Shoeybi, and Bryan Catanzaro.
\newblock Odin: Disentangled reward mitigates hacking in rlhf.
\newblock In \emph{International Conference on Machine Learning}, pp.\  7935--7952. PMLR, 2024{\natexlab{a}}.

\bibitem[Chen et~al.(2019)Chen, Tang, Wiseman, and Gimpel]{chen2019controllable}
Mingda Chen, Qingming Tang, Sam Wiseman, and Kevin Gimpel.
\newblock Controllable paraphrase generation with a syntactic exemplar.
\newblock In \emph{Proceedings of the 57th Annual Meeting of the Association for Computational Linguistics}, pp.\  5972--5984, 2019.

\bibitem[Chen et~al.(2017)Chen, Zhu, Ling, Wei, Jiang, and Inkpen]{chen2017enhanced}
Qian Chen, Xiaodan Zhu, Zhen-Hua Ling, Si~Wei, Hui Jiang, and Diana Inkpen.
\newblock Enhanced lstm for natural language inference.
\newblock In \emph{Proceedings of the 55th Annual Meeting of the Association for Computational Linguistics (Volume 1: Long Papers)}, pp.\  1657--1668, 2017.

\bibitem[Chen et~al.(2023)Chen, Zhao, Zhang, Chern, Gao, Liu, and He]{chen2023felm}
Shiqi Chen, Yiran Zhao, Jinghan Zhang, I-Chun Chern, Siyang Gao, Pengfei Liu, and Junxian He.
\newblock Felm: benchmarking factuality evaluation of large language models.
\newblock In \emph{Proceedings of the 37th International Conference on Neural Information Processing Systems}, pp.\  44502--44523, 2023.

\bibitem[Chen \& Wan(2023)Chen and Wan]{chen2023comprehensive}
Xiang Chen and Xiaojun Wan.
\newblock A comprehensive evaluation of constrained text generation for large language models.
\newblock \emph{arXiv preprint arXiv:2310.16343}, 2023.

\bibitem[Chen et~al.(2024{\natexlab{b}})Chen, Deng, Yuan, Ji, and Gu]{chen2024self}
Zixiang Chen, Yihe Deng, Huizhuo Yuan, Kaixuan Ji, and Quanquan Gu.
\newblock Self-play fine-tuning converts weak language models to strong language models.
\newblock In \emph{International Conference on Machine Learning}, pp.\  6621--6642. PMLR, 2024{\natexlab{b}}.

\bibitem[Choshen et~al.(2020)Choshen, Fox, Aizenbud, and Abend]{choshen2019weaknesses}
Leshem Choshen, Lior Fox, Zohar Aizenbud, and Omri Abend.
\newblock On the weaknesses of reinforcement learning for neural machine translation.
\newblock In \emph{International Conference on Learning Representations}, 2020.

\bibitem[Chowdhury et~al.(2024)Chowdhury, Islam, Kumar, Shezan, Jain, and Chadha]{chowdhury2024breaking}
Arijit~Ghosh Chowdhury, Md~Mofijul Islam, Vaibhav Kumar, Faysal~Hossain Shezan, Vinija Jain, and Aman Chadha.
\newblock Breaking down the defenses: A comparative survey of attacks on large language models.
\newblock \emph{arXiv preprint arXiv:2403.04786}, 2024.

\bibitem[Christiano et~al.(2017)Christiano, Leike, Brown, Martic, Legg, and Amodei]{christiano2017deep}
Paul~F Christiano, Jan Leike, Tom Brown, Miljan Martic, Shane Legg, and Dario Amodei.
\newblock Deep reinforcement learning from human preferences.
\newblock In \emph{Advances in Neural Information Processing Systems}, pp.\  4299--4307, 2017.

\bibitem[Clark et~al.(2018)Clark, Ji, and Smith]{clark2018neural}
Elizabeth Clark, Yangfeng Ji, and Noah~A Smith.
\newblock Neural text generation in stories using entity representations as context.
\newblock In \emph{Proceedings of the 2018 Conference of the North American Chapter of the Association for Computational Linguistics: Human Language Technologies, Volume 1 (Long Papers)}, pp.\  2250--2260, 2018.

\bibitem[Clark et~al.(2019)Clark, Celikyilmaz, and Smith]{clark2019sentence}
Elizabeth Clark, Asli Celikyilmaz, and Noah~A Smith.
\newblock Sentence mover’s similarity: Automatic evaluation for multi-sentence texts.
\newblock In \emph{Proceedings of the 57th Annual Meeting of the Association for Computational Linguistics}, pp.\  2748--2760, 2019.

\bibitem[Clark et~al.(2021)Clark, August, Serrano, Haduong, Gururangan, and Smith]{clark2021all}
Elizabeth Clark, Tal August, Sofia Serrano, Nikita Haduong, Suchin Gururangan, and Noah~A Smith.
\newblock All that’s ‘human’is not gold: Evaluating human evaluation of generated text.
\newblock In \emph{Proceedings of the 59th Annual Meeting of the Association for Computational Linguistics and the 11th International Joint Conference on Natural Language Processing (Volume 1: Long Papers)}, pp.\  7282--7296, 2021.

\bibitem[Cohen et~al.(2024)Cohen, Biran, Yoran, Globerson, and Geva]{cohen2024evaluating}
Roi Cohen, Eden Biran, Ori Yoran, Amir Globerson, and Mor Geva.
\newblock Evaluating the ripple effects of knowledge editing in language models.
\newblock \emph{Transactions of the Association for Computational Linguistics}, 12:\penalty0 283--298, 2024.

\bibitem[Colombo et~al.(2022)Colombo, Clavel, and Piantanida]{colombo2022infolm}
Pierre Jean~A Colombo, Chlo{\'e} Clavel, and Pablo Piantanida.
\newblock Infolm: A new metric to evaluate summarization \& data2text generation.
\newblock In \emph{Proceedings of the AAAI Conference on Artificial Intelligence}, volume~36, pp.\  10554--10562, 2022.

\bibitem[Crego et~al.(2016)Crego, Kim, Klein, Rebollo, Yang, Senellart, Akhanov, Brunelle, Coquard, Deng, et~al.]{crego2016systran}
Josep Crego, Jungi Kim, Guillaume Klein, Anabel Rebollo, Kathy Yang, Jean Senellart, Egor Akhanov, Patrice Brunelle, Aurelien Coquard, Yongchao Deng, et~al.
\newblock Systran's pure neural machine translation systems.
\newblock \emph{arXiv preprint arXiv:1610.05540}, 2016.

\bibitem[Dai et~al.(2022)Dai, Dong, Hao, Sui, Chang, and Wei]{dai2022knowledge}
Damai Dai, Li~Dong, Yaru Hao, Zhifang Sui, Baobao Chang, and Furu Wei.
\newblock Knowledge neurons in pretrained transformers.
\newblock In \emph{Proceedings of the 60th Annual Meeting of the Association for Computational Linguistics (Volume 1: Long Papers)}, pp.\  8493--8502, 2022.

\bibitem[Dathathri et~al.(2020)Dathathri, Madotto, Lan, Hung, Frank, Molino, Yosinski, and Liu]{dathathri2019plug}
Sumanth Dathathri, Andrea Madotto, Janice Lan, Jane Hung, Eric Frank, Piero Molino, Jason Yosinski, and Rosanne Liu.
\newblock Plug and play language models: A simple approach to controlled text generation.
\newblock In \emph{International Conference on Learning Representations}, 2020.

\bibitem[Dehghani et~al.(2021)Dehghani, Tay, Gritsenko, Zhao, Houlsby, Diaz, Metzler, and Vinyals]{dehghani2021benchmark}
Mostafa Dehghani, Yi~Tay, Alexey~A Gritsenko, Zhe Zhao, Neil Houlsby, Fernando Diaz, Donald Metzler, and Oriol Vinyals.
\newblock The benchmark lottery.
\newblock \emph{arXiv preprint arXiv:2107.07002}, 2021.

\bibitem[Devlin et~al.(2019)Devlin, Chang, Lee, and Toutanova]{devlin2018bert}
Jacob Devlin, Ming-Wei Chang, Kenton Lee, and Kristina Toutanova.
\newblock Bert: Pre-training of deep bidirectional transformers for language understanding.
\newblock In \emph{Proceedings of the 2019 conference of the North American chapter of the association for computational linguistics: human language technologies, volume 1 (long and short papers)}, pp.\  4171--4186, 2019.

\bibitem[Dhariwal \& Nichol(2021)Dhariwal and Nichol]{dhariwal2021diffusion}
Prafulla Dhariwal and Alexander Nichol.
\newblock Diffusion models beat gans on image synthesis.
\newblock \emph{Advances in neural information processing systems}, 34:\penalty0 8780--8794, 2021.

\bibitem[Dodge et~al.(2021)Dodge, Sap, Marasovi{\'c}, Agnew, Ilharco, Groeneveld, Mitchell, and Gardner]{dodge2021documenting}
Jesse Dodge, Maarten Sap, Ana Marasovi{\'c}, William Agnew, Gabriel Ilharco, Dirk Groeneveld, Margaret Mitchell, and Matt Gardner.
\newblock Documenting large webtext corpora: A case study on the colossal clean crawled corpus.
\newblock In \emph{Proceedings of the 2021 Conference on Empirical Methods in Natural Language Processing}, pp.\  1286--1305, 2021.

\bibitem[Donahue et~al.(2020)Donahue, Lee, and Liang]{donahue2020enabling}
Chris Donahue, Mina Lee, and Percy Liang.
\newblock Enabling language models to fill in the blanks.
\newblock In \emph{Proceedings of the 58th Annual Meeting of the Association for Computational Linguistics}, pp.\  2492--2501. Association for Computational Linguistics, 2020.

\bibitem[Dong et~al.(2023)Dong, Wang, Sreedhar, Wu, and Kuchaiev]{dong2023steerlm}
Yi~Dong, Zhilin Wang, Makesh Sreedhar, Xianchao Wu, and Oleksii Kuchaiev.
\newblock Steerlm: Attribute conditioned sft as an (user-steerable) alternative to rlhf.
\newblock In \emph{Findings of the Association for Computational Linguistics: EMNLP 2023}, pp.\  11275--11288, 2023.

\bibitem[Dong et~al.(2024)Dong, Zhou, Yang, Shao, and Qiao]{dong2024attacks}
Zhichen Dong, Zhanhui Zhou, Chao Yang, Jing Shao, and Yu~Qiao.
\newblock Attacks, defenses and evaluations for llm conversation safety: A survey.
\newblock In \emph{Proceedings of the 2024 Conference of the North American Chapter of the Association for Computational Linguistics: Human Language Technologies (Volume 1: Long Papers)}, pp.\  6734--6747, 2024.

\bibitem[Dubey et~al.(2024)Dubey, Jauhri, Pandey, Kadian, Al-Dahle, Letman, Mathur, Schelten, Yang, Fan, et~al.]{dubey2024llama}
Abhimanyu Dubey, Abhinav Jauhri, Abhinav Pandey, Abhishek Kadian, Ahmad Al-Dahle, Aiesha Letman, Akhil Mathur, Alan Schelten, Amy Yang, Angela Fan, et~al.
\newblock The llama 3 herd of models.
\newblock \emph{arXiv preprint arXiv:2407.21783}, 2024.

\bibitem[Dubois et~al.(2024{\natexlab{a}})Dubois, Galambosi, Liang, and Hashimoto]{dubois2024length}
Yann Dubois, Bal{\'a}zs Galambosi, Percy Liang, and Tatsunori~B Hashimoto.
\newblock Length-controlled alpacaeval: A simple way to debias automatic evaluators.
\newblock \emph{arXiv preprint arXiv:2404.04475}, 2024{\natexlab{a}}.

\bibitem[Dubois et~al.(2024{\natexlab{b}})Dubois, Li, Taori, Zhang, Gulrajani, Ba, Guestrin, Liang, and Hashimoto]{dubois2024alpacafarm}
Yann Dubois, Chen~Xuechen Li, Rohan Taori, Tianyi Zhang, Ishaan Gulrajani, Jimmy Ba, Carlos Guestrin, Percy~S Liang, and Tatsunori~B Hashimoto.
\newblock Alpacafarm: A simulation framework for methods that learn from human feedback.
\newblock \emph{Advances in Neural Information Processing Systems}, 36, 2024{\natexlab{b}}.

\bibitem[Edunov et~al.(2020)Edunov, Ott, Ranzato, and Auli]{edunov2020evaluation}
Sergey Edunov, Myle Ott, Marc’Aurelio Ranzato, and Michael Auli.
\newblock On the evaluation of machine translation systems trained with back-translation.
\newblock In \emph{Proceedings of the 58th Annual Meeting of the Association for Computational Linguistics}, pp.\  2836--2846, 2020.

\bibitem[Elangovan et~al.(2021)Elangovan, He, and Verspoor]{elangovan2021memorization}
Aparna Elangovan, Jiayuan He, and Karin Verspoor.
\newblock Memorization vs. generalization: Quantifying data leakage in nlp performance evaluation.
\newblock In \emph{Proceedings of the 16th Conference of the European Chapter of the Association for Computational Linguistics: Main Volume}, pp.\  1325--1335, 2021.

\bibitem[Ettinger(2020)]{ettinger2020bert}
Allyson Ettinger.
\newblock What bert is not: Lessons from a new suite of psycholinguistic diagnostics for language models.
\newblock \emph{Transactions of the Association for Computational Linguistics}, 8:\penalty0 34--48, 2020.

\bibitem[Falke et~al.(2019)Falke, Ribeiro, Utama, Dagan, and Gurevych]{falke2019ranking}
Tobias Falke, Leonardo~FR Ribeiro, Prasetya~Ajie Utama, Ido Dagan, and Iryna Gurevych.
\newblock Ranking generated summaries by correctness: An interesting but challenging application for natural language inference.
\newblock In \emph{Proceedings of the 57th annual meeting of the association for computational linguistics}, pp.\  2214--2220, 2019.

\bibitem[Fan et~al.(2018{\natexlab{a}})Fan, Grangier, and Auli]{fan2018controllable}
Angela Fan, David Grangier, and Michael Auli.
\newblock Controllable abstractive summarization.
\newblock In \emph{Proceedings of the 2nd Workshop on Neural Machine Translation and Generation}, pp.\  45--54, 2018{\natexlab{a}}.

\bibitem[Fan et~al.(2018{\natexlab{b}})Fan, Lewis, and Dauphin]{fan2018hierarchical}
Angela Fan, Mike Lewis, and Yann Dauphin.
\newblock Hierarchical neural story generation.
\newblock In \emph{Proceedings of the 56th Annual Meeting of the Association for Computational Linguistics (Volume 1: Long Papers)}, pp.\  889--898, 2018{\natexlab{b}}.

\bibitem[Fan et~al.(2019)Fan, Lewis, and Dauphin]{fan2019strategies}
Angela Fan, Mike Lewis, and Yann Dauphin.
\newblock Strategies for structuring story generation.
\newblock In \emph{Proceedings of the 57th Annual Meeting of the Association for Computational Linguistics}, pp.\  2650--2660, 2019.

\bibitem[Fedus et~al.(2018)Fedus, Goodfellow, and Dai]{fedus2018maskgan}
William Fedus, Ian Goodfellow, and Andrew~M Dai.
\newblock Maskgan: Better text generation via filling in the \_.
\newblock In \emph{International Conference on Learning Representations}, 2018.

\bibitem[Ficler \& Goldberg(2017)Ficler and Goldberg]{ficler2017controlling}
Jessica Ficler and Yoav Goldberg.
\newblock Controlling linguistic style aspects in neural language generation.
\newblock \emph{EMNLP 2017}, pp.\ ~94, 2017.

\bibitem[Flesch(1979)]{flesch1979write}
Rudolf~Franz Flesch.
\newblock \emph{How to write plain English: A book for lawyers and consumers}.
\newblock Harpercollins, 1979.

\bibitem[Fomicheva \& Specia(2019)Fomicheva and Specia]{fomicheva2019taking}
Marina Fomicheva and Lucia Specia.
\newblock Taking mt evaluation metrics to extremes: Beyond correlation with human judgments.
\newblock \emph{Computational Linguistics}, 45\penalty0 (3):\penalty0 515--558, 2019.

\bibitem[Freitag et~al.(2020)Freitag, Grangier, and Caswell]{freitag2020bleu}
Markus Freitag, David Grangier, and Isaac Caswell.
\newblock Bleu might be guilty but references are not innocent.
\newblock In \emph{Proceedings of the 2020 Conference on Empirical Methods in Natural Language Processing (EMNLP)}, pp.\  61--71, 2020.

\bibitem[Fu et~al.(2024)Fu, Ng, Jiang, and Liu]{fu2023gptscore}
Jinlan Fu, See~Kiong Ng, Zhengbao Jiang, and Pengfei Liu.
\newblock Gptscore: Evaluate as you desire.
\newblock In \emph{Proceedings of the 2024 Conference of the North American Chapter of the Association for Computational Linguistics: Human Language Technologies (Volume 1: Long Papers)}, pp.\  6556--6576, 2024.

\bibitem[Fu \& Feng(2018)Fu and Feng]{fu2018natural}
Yao Fu and Yansong Feng.
\newblock Natural answer generation with heterogeneous memory.
\newblock In \emph{Proceedings of the 2018 Conference of the North American Chapter of the Association for Computational Linguistics: Human Language Technologies, Volume 1 (Long Papers)}, pp.\  185--195, 2018.

\bibitem[Fu et~al.(2018)Fu, Tan, Peng, Zhao, and Yan]{fu2018style}
Zhenxin Fu, Xiaoye Tan, Nanyun Peng, Dongyan Zhao, and Rui Yan.
\newblock Style transfer in text: Exploration and evaluation.
\newblock In \emph{Thirty-Second AAAI Conference on Artificial Intelligence}, 2018.

\bibitem[Gallegos et~al.(2024)Gallegos, Rossi, Barrow, Tanjim, Kim, Dernoncourt, Yu, Zhang, and Ahmed]{gallegos2024bias}
Isabel~O Gallegos, Ryan~A Rossi, Joe Barrow, Md~Mehrab Tanjim, Sungchul Kim, Franck Dernoncourt, Tong Yu, Ruiyi Zhang, and Nesreen~K Ahmed.
\newblock Bias and fairness in large language models: A survey.
\newblock \emph{Computational Linguistics}, pp.\  1--79, 2024.

\bibitem[Galley et~al.(2015)Galley, Brockett, Sordoni, Ji, Auli, Quirk, Mitchell, Gao, and Dolan]{galley2015deltableu}
Michel Galley, Chris Brockett, Alessandro Sordoni, Yangfeng Ji, Michael Auli, Chris Quirk, Margaret Mitchell, Jianfeng Gao, and Bill Dolan.
\newblock deltableu: A discriminative metric for generation tasks with intrinsically diverse targets.
\newblock In \emph{Proceedings of the 53rd Annual Meeting of the Association for Computational Linguistics and the 7th International Joint Conference on Natural Language Processing (Volume 2: Short Papers)}, volume~2, pp.\  445--450, 2015.

\bibitem[Gao et~al.(2019{\natexlab{a}})Gao, Galley, and Li]{gao2019neural}
Jianfeng Gao, Michel Galley, and Lihong Li.
\newblock \emph{Neural approaches to conversational AI: Question answering, task-oriented dialogues and social chatbots}.
\newblock Now Foundations and Trends, 2019{\natexlab{a}}.

\bibitem[Gao et~al.(2019{\natexlab{b}})Gao, Meyer, Mesgar, and Gurevych]{gao2019reward}
Yang Gao, Christian Meyer, Mohsen Mesgar, and Iryna Gurevych.
\newblock Reward learning for efficient reinforcement learning in extractive document summarisation.
\newblock In \emph{Proceedings of the 28th International Joint Conference on Artificial Intelligence}, pp.\  2350--2356, 2019{\natexlab{b}}.

\bibitem[Gatt \& Krahmer(2018)Gatt and Krahmer]{gatt2018survey}
Albert Gatt and Emiel Krahmer.
\newblock Survey of the state of the art in natural language generation: Core tasks, applications and evaluation.
\newblock \emph{Journal of Artificial Intelligence Research}, 61:\penalty0 65--170, 2018.

\bibitem[Ge et~al.(2024)Ge, Chan, Wang, Yu, Mi, and Yu]{ge2024scaling}
Tao Ge, Xin Chan, Xiaoyang Wang, Dian Yu, Haitao Mi, and Dong Yu.
\newblock Scaling synthetic data creation with 1,000,000,000 personas.
\newblock \emph{arXiv preprint arXiv:2406.20094}, 2024.

\bibitem[Gehman et~al.(2020)Gehman, Gururangan, Sap, Choi, and Smith]{gehman2020realtoxicityprompts}
Samuel Gehman, Suchin Gururangan, Maarten Sap, Yejin Choi, and Noah~A Smith.
\newblock Realtoxicityprompts: Evaluating neural toxic degeneration in language models.
\newblock In \emph{Findings of the Association for Computational Linguistics: EMNLP 2020}, pp.\  3356--3369, 2020.

\bibitem[Gerv{\'a}s(2009)]{gervas2009computational}
Pablo Gerv{\'a}s.
\newblock Computational approaches to storytelling and creativity.
\newblock \emph{AI Magazine}, 30\penalty0 (3):\penalty0 49--49, 2009.

\bibitem[Ghazvininejad et~al.(2016)Ghazvininejad, Shi, Choi, and Knight]{ghazvininejad2016generating}
Marjan Ghazvininejad, Xing Shi, Yejin Choi, and Kevin Knight.
\newblock Generating topical poetry.
\newblock In \emph{Proceedings of the 2016 Conference on Empirical Methods in Natural Language Processing}, pp.\  1183--1191, 2016.

\bibitem[Ghazvininejad et~al.(2017)Ghazvininejad, Shi, Priyadarshi, and Knight]{ghazvininejad2017hafez}
Marjan Ghazvininejad, Xing Shi, Jay Priyadarshi, and Kevin Knight.
\newblock Hafez: an interactive poetry generation system.
\newblock In \emph{Proceedings of ACL 2017, System Demonstrations}, pp.\  43--48, 2017.

\bibitem[Ghosh et~al.(2017)Ghosh, Chollet, Laksana, Morency, and Scherer]{ghosh2017affect}
Sayan Ghosh, Mathieu Chollet, Eugene Laksana, Louis-Philippe Morency, and Stefan Scherer.
\newblock Affect-lm: A neural language model for customizable affective text generation.
\newblock In \emph{Proceedings of the 55th Annual Meeting of the Association for Computational Linguistics (Volume 1: Long Papers)}, pp.\  634--642, 2017.

\bibitem[Gimpel et~al.(2013)Gimpel, Batra, Dyer, and Shakhnarovich]{gimpel2013systematic}
Kevin Gimpel, Dhruv Batra, Chris Dyer, and Gregory Shakhnarovich.
\newblock A systematic exploration of diversity in machine translation.
\newblock In \emph{Proceedings of the 2013 Conference on Empirical Methods in Natural Language Processing}, pp.\  1100--1111, 2013.

\bibitem[Gkatzia \& Mahamood(2015)Gkatzia and Mahamood]{gkatzia2015snapshot}
Dimitra Gkatzia and Saad Mahamood.
\newblock A snapshot of nlg evaluation practices 2005-2014.
\newblock In \emph{Proceedings of the 15th European Workshop on Natural Language Generation (ENLG)}, pp.\  57--60, 2015.

\bibitem[Goldfarb-Tarrant et~al.(2019)Goldfarb-Tarrant, Feng, and Peng]{goldfarb2019plan}
Seraphina Goldfarb-Tarrant, Haining Feng, and Nanyun Peng.
\newblock Plan, write, and revise: an interactive system for open-domain story generation.
\newblock In \emph{Proceedings of the 2019 Conference of the North American Chapter of the Association for Computational Linguistics (Demonstrations)}, pp.\  89--97, 2019.

\bibitem[Gong et~al.(2022)Gong, Li, Feng, Wu, and Kong]{gong2022diffuseq}
Shansan Gong, Mukai Li, Jiangtao Feng, Zhiyong Wu, and Lingpeng Kong.
\newblock Diffuseq: Sequence to sequence text generation with diffusion models.
\newblock In \emph{The Eleventh International Conference on Learning Representations}, 2022.

\bibitem[Goodfellow et~al.(2014)Goodfellow, Pouget-Abadie, Mirza, Xu, Warde-Farley, Ozair, Courville, and Bengio]{goodfellow2014generative}
Ian Goodfellow, Jean Pouget-Abadie, Mehdi Mirza, Bing Xu, David Warde-Farley, Sherjil Ozair, Aaron Courville, and Yoshua Bengio.
\newblock Generative adversarial nets.
\newblock In \emph{Advances in neural information processing systems}, pp.\  2672--2680, 2014.

\bibitem[Graham et~al.(2019)Graham, Haddow, and Koehn]{graham2019translationese}
Yvette Graham, Barry Haddow, and Philipp Koehn.
\newblock Translationese in machine translation evaluation.
\newblock \emph{arXiv preprint arXiv:1906.09833}, 2019.

\bibitem[Grosse et~al.(2023)Grosse, Bae, Anil, Elhage, Tamkin, Tajdini, Steiner, Li, Durmus, Perez, et~al.]{grosse2023studying}
Roger Grosse, Juhan Bae, Cem Anil, Nelson Elhage, Alex Tamkin, Amirhossein Tajdini, Benoit Steiner, Dustin Li, Esin Durmus, Ethan Perez, et~al.
\newblock Studying large language model generalization with influence functions.
\newblock \emph{arXiv preprint arXiv:2308.03296}, 2023.

\bibitem[Gu et~al.(2019{\natexlab{a}})Gu, Liu, and Cho]{gu2019insertion}
Jiatao Gu, Qi~Liu, and Kyunghyun Cho.
\newblock Insertion-based decoding with automatically inferred generation order.
\newblock \emph{Transactions of the Association for Computational Linguistics}, 7:\penalty0 661--676, 2019{\natexlab{a}}.

\bibitem[Gu et~al.(2019{\natexlab{b}})Gu, Wang, and Zhao]{gu2019levenshtein}
Jiatao Gu, Changhan Wang, and Junbo Zhao.
\newblock Levenshtein transformer.
\newblock \emph{Advances in neural information processing systems}, 32, 2019{\natexlab{b}}.

\bibitem[Gui et~al.(2024)Gui, G\^arbacea, and Veitch]{gui2024bonbon}
Lin Gui, Cristina G\^arbacea, and Victor Veitch.
\newblock Bonbon alignment for large language models and the sweetness of best-of-n sampling.
\newblock In \emph{The Thirty-eighth Annual Conference on Neural Information Processing Systems}, 2024.

\bibitem[Gupta et~al.(2024)Gupta, Rao, and Anumanchipalli]{gupta2024model}
Akshat Gupta, Anurag Rao, and Gopala Anumanchipalli.
\newblock Model editing at scale leads to gradual and catastrophic forgetting.
\newblock In \emph{Findings of the Association for Computational Linguistics ACL 2024}, pp.\  15202--15232, 2024.

\bibitem[Guu et~al.(2018)Guu, Hashimoto, Oren, and Liang]{guu2018generating}
Kelvin Guu, Tatsunori~B Hashimoto, Yonatan Oren, and Percy Liang.
\newblock Generating sentences by editing prototypes.
\newblock \emph{Transactions of the Association of Computational Linguistics}, 6:\penalty0 437--450, 2018.

\bibitem[H.~Lee et~al.(2020)H.~Lee, Shu, Achananuparp, Prasetyo, Liu, Lim, and Varshney]{h2020recipegpt}
Helena H.~Lee, Ke~Shu, Palakorn Achananuparp, Philips~Kokoh Prasetyo, Yue Liu, Ee-Peng Lim, and Lav~R Varshney.
\newblock Recipegpt: Generative pre-training based cooking recipe generation and evaluation system.
\newblock In \emph{Companion Proceedings of the Web Conference 2020}, pp.\  181--184, 2020.

\bibitem[Hambardzumyan et~al.(2021)Hambardzumyan, Khachatrian, and May]{hambardzumyan2021warp}
Karen Hambardzumyan, Hrant Khachatrian, and Jonathan May.
\newblock Warp: Word-level adversarial reprogramming.
\newblock In \emph{Proceedings of the 59th Annual Meeting of the Association for Computational Linguistics and the 11th International Joint Conference on Natural Language Processing (Volume 1: Long Papers)}, pp.\  4921--4933, 2021.

\bibitem[Han et~al.(2023)Han, Xu, Li, Fung, Sun, Jiang, Abdelzaher, and Ji]{han2023lm}
Chi Han, Jialiang Xu, Manling Li, Yi~Fung, Chenkai Sun, Nan Jiang, Tarek Abdelzaher, and Heng Ji.
\newblock Lm-switch: Lightweight language model conditioning in word embedding space.
\newblock \emph{arXiv preprint arXiv:2305.12798}, 2023.

\bibitem[Hase et~al.(2024)Hase, Bansal, Kim, and Ghandeharioun]{hase2024does}
Peter Hase, Mohit Bansal, Been Kim, and Asma Ghandeharioun.
\newblock Does localization inform editing? surprising differences in causality-based localization vs. knowledge editing in language models.
\newblock \emph{Advances in Neural Information Processing Systems}, 36, 2024.

\bibitem[Hashimoto et~al.(2019)Hashimoto, Zhang, and Liang]{hashimoto2019unifying}
Tatsunori~B Hashimoto, Hugh Zhang, and Percy Liang.
\newblock Unifying human and statistical evaluation for natural language generation.
\newblock In \emph{Proceedings of the 2019 Conference of the North American Chapter of the Association for Computational Linguistics: Human Language Technologies, Volume 1 (Long and Short Papers)}, pp.\  1689--1701, 2019.

\bibitem[He et~al.(2023)He, Zhang, Wang, Kumar, Cho, Glass, and Tsvetkov]{he2022blind}
Tianxing He, Jingyu Zhang, Tianle Wang, Sachin Kumar, Kyunghyun Cho, James Glass, and Yulia Tsvetkov.
\newblock On the blind spots of model-based evaluation metrics for text generation.
\newblock In \emph{ACL: Annual Meeting of the Association for Computational Linguistics}, 2023.

\bibitem[Hendrycks et~al.(2021)Hendrycks, Burns, Basart, Zou, Mazeika, Song, and Steinhardt]{hendrycks2020measuring}
Dan Hendrycks, Collin Burns, Steven Basart, Andy Zou, Mantas Mazeika, Dawn Song, and Jacob Steinhardt.
\newblock Measuring massive multitask language understanding.
\newblock In \emph{International Conference on Learning Representations}, 2021.

\bibitem[Hernandez et~al.(2024)Hernandez, Li, and Andreas]{hernandez2023measuring}
Evan Hernandez, Belinda~Z. Li, and Jacob Andreas.
\newblock Inspecting and editing knowledge representations in language models.
\newblock In \emph{First Conference on Language Modeling}, 2024.

\bibitem[Ho \& Salimans(2022)Ho and Salimans]{ho2022classifier}
Jonathan Ho and Tim Salimans.
\newblock Classifier-free diffusion guidance.
\newblock \emph{arXiv preprint arXiv:2207.12598}, 2022.

\bibitem[Hokamp \& Liu(2017)Hokamp and Liu]{hokamp2017lexically}
Chris Hokamp and Qun Liu.
\newblock Lexically constrained decoding for sequence generation using grid beam search.
\newblock In \emph{Proceedings of the 55th Annual Meeting of the Association for Computational Linguistics (Volume 1: Long Papers)}, pp.\  1535--1546, 2017.

\bibitem[Holtzman et~al.(2018)Holtzman, Buys, Forbes, Bosselut, Golub, and Choi]{holtzman2018learning}
Ari Holtzman, Jan Buys, Maxwell Forbes, Antoine Bosselut, David Golub, and Yejin Choi.
\newblock Learning to write with cooperative discriminators.
\newblock In \emph{Proceedings of the 56th Annual Meeting of the Association for Computational Linguistics (Volume 1: Long Papers)}, pp.\  1638--1649, 2018.

\bibitem[Holtzman et~al.(2020)Holtzman, Buys, Du, Forbes, and Choi]{holtzman2019curious}
Ari Holtzman, Jan Buys, Li~Du, Maxwell Forbes, and Yejin Choi.
\newblock The curious case of neural text degeneration.
\newblock In \emph{International Conference on Learning Representations}, 2020.

\bibitem[Hosseini et~al.(2024)Hosseini, Sordoni, Toyama, Courville, and Agarwal]{hosseini2024not}
Arian Hosseini, Alessandro Sordoni, Daniel Toyama, Aaron Courville, and Rishabh Agarwal.
\newblock Not all llm reasoners are created equal.
\newblock \emph{arXiv preprint arXiv:2410.01748}, 2024.

\bibitem[Hsieh et~al.(2021)Hsieh, Lee, and Lim]{hsieh2021enconter}
Lee~Hsun Hsieh, Yang-Yin Lee, and Ee-Peng Lim.
\newblock Enconter: Entity constrained progressive sequence generation via insertion-based transformer.
\newblock In \emph{Proceedings of the 16th Conference of the European Chapter of the Association for Computational Linguistics: Main Volume}, pp.\  3590--3599, 2021.

\bibitem[Hu et~al.(2019)Hu, Khayrallah, Culkin, Xia, Chen, Post, and Van~Durme]{hu2019improved}
J~Edward Hu, Huda Khayrallah, Ryan Culkin, Patrick Xia, Tongfei Chen, Matt Post, and Benjamin Van~Durme.
\newblock Improved lexically constrained decoding for translation and monolingual rewriting.
\newblock In \emph{Proceedings of the 2019 Conference of the North American Chapter of the Association for Computational Linguistics: Human Language Technologies, Volume 1 (Long and Short Papers)}, pp.\  839--850, 2019.

\bibitem[Hu et~al.(2024)Hu, Gao, Hu, Zhang, Chen, Xu, and Wan]{hu2024llm}
Xinyu Hu, Mingqi Gao, Sen Hu, Yang Zhang, Yicheng Chen, Teng Xu, and Xiaojun Wan.
\newblock Are llm-based evaluators confusing nlg quality criteria?
\newblock In \emph{Proceedings of the 62nd Annual Meeting of the Association for Computational Linguistics (Volume 1: Long Papers)}, pp.\  9530--9570, 2024.

\bibitem[Hu et~al.(2017)Hu, Yang, Liang, Salakhutdinov, and Xing]{hu2017toward}
Zhiting Hu, Zichao Yang, Xiaodan Liang, Ruslan Salakhutdinov, and Eric~P Xing.
\newblock Toward controlled generation of text.
\newblock In \emph{Proceedings of the 34th International Conference on Machine Learning-Volume 70}, pp.\  1587--1596. JMLR. org, 2017.

\bibitem[Huang \& Chang(2023)Huang and Chang]{huang2023towards}
Jie Huang and Kevin Chen-Chuan Chang.
\newblock Towards reasoning in large language models: A survey.
\newblock In \emph{61st Annual Meeting of the Association for Computational Linguistics, ACL 2023}, pp.\  1049--1065. Association for Computational Linguistics (ACL), 2023.

\bibitem[Hwang et~al.(2023)Hwang, Majumder, and Tandon]{hwang2023aligning}
EunJeong Hwang, Bodhisattwa Majumder, and Niket Tandon.
\newblock Aligning language models to user opinions.
\newblock In \emph{Findings of the Association for Computational Linguistics: EMNLP 2023}, pp.\  5906--5919, 2023.

\bibitem[Ilharco et~al.(2023)Ilharco, Ribeiro, Wortsman, Schmidt, Hajishirzi, and Farhadi]{ilharco2022editing}
Gabriel Ilharco, Marco~Tulio Ribeiro, Mitchell Wortsman, Ludwig Schmidt, Hannaneh Hajishirzi, and Ali Farhadi.
\newblock Editing models with task arithmetic.
\newblock In \emph{The Eleventh International Conference on Learning Representations}, 2023.

\bibitem[Ippolito et~al.(2018)Ippolito, Kriz, Sedoc, Kustikova, Callison-Burch, Kriz, Miltsakaki, Apidianaki, Callison-Burch, Hewitt, et~al.]{ippolito2018comparison}
Daphne Ippolito, Reno Kriz, Joao Sedoc, Maria Kustikova, Chris Callison-Burch, Reno Kriz, Eleni Miltsakaki, Marianna Apidianaki, Chris Callison-Burch, John Hewitt, et~al.
\newblock Comparison of diverse decoding methods from conditional language models.
\newblock In \emph{Proceedings of the 57th Conference of the Association for Computational Linguistics}. Association for Computational Linguistics, 2018.

\bibitem[Iqbal et~al.(2024)Iqbal, Wang, Wang, Georgiev, Geng, Gurevych, and Nakov]{iqbal-etal-2024-openfactcheck}
Hasan Iqbal, Yuxia Wang, Minghan Wang, Georgi~Nenkov Georgiev, Jiahui Geng, Iryna Gurevych, and Preslav Nakov.
\newblock {O}pen{F}act{C}heck: A unified framework for factuality evaluation of {LLM}s.
\newblock In Delia~Irazu Hernandez~Farias, Tom Hope, and Manling Li (eds.), \emph{Proceedings of the 2024 Conference on Empirical Methods in Natural Language Processing: System Demonstrations}, pp.\  219--229, Miami, Florida, USA, November 2024. Association for Computational Linguistics.
\newblock \doi{10.18653/v1/2024.emnlp-demo.23}.
\newblock URL \url{https://aclanthology.org/2024.emnlp-demo.23/}.

\bibitem[Iso(2024)]{iso2024autotemplate}
Hayate Iso.
\newblock Autotemplate: A simple recipe for lexically constrained text generation.
\newblock In \emph{INLG}, 2024.
\newblock URL \url{https://arxiv.org/abs/2211.08387}.

\bibitem[Iyyer et~al.(2014)Iyyer, Boyd-Graber, Claudino, Socher, and Daum{\'e}~III]{iyyer2014neural}
Mohit Iyyer, Jordan Boyd-Graber, Leonardo Claudino, Richard Socher, and Hal Daum{\'e}~III.
\newblock A neural network for factoid question answering over paragraphs.
\newblock In \emph{Proceedings of the 2014 conference on empirical methods in natural language processing (EMNLP)}, pp.\  633--644, 2014.

\bibitem[Iyyer et~al.(2018)Iyyer, Wieting, Gimpel, and Zettlemoyer]{iyyer2018adversarial}
Mohit Iyyer, John Wieting, Kevin Gimpel, and Luke Zettlemoyer.
\newblock Adversarial example generation with syntactically controlled paraphrase networks.
\newblock In \emph{Proceedings of the 2018 Conference of the North American Chapter of the Association for Computational Linguistics: Human Language Technologies, Volume 1 (Long Papers)}, pp.\  1875--1885, 2018.

\bibitem[Jacovi et~al.(2025)Jacovi, Wang, Alberti, Tao, Lipovetz, Olszewska, Haas, Liu, Keating, Bloniarz, et~al.]{jacovi2025facts}
Alon Jacovi, Andrew Wang, Chris Alberti, Connie Tao, Jon Lipovetz, Kate Olszewska, Lukas Haas, Michelle Liu, Nate Keating, Adam Bloniarz, et~al.
\newblock The facts grounding leaderboard: Benchmarking llms' ability to ground responses to long-form input.
\newblock \emph{arXiv preprint arXiv:2501.03200}, 2025.

\bibitem[Jaques et~al.(2019)Jaques, Ghandeharioun, Shen, Ferguson, Lapedriza, Jones, Gu, and Picard]{jaques2019way}
Natasha Jaques, Asma Ghandeharioun, Judy~Hanwen Shen, Craig Ferguson, Agata Lapedriza, Noah Jones, Shixiang Gu, and Rosalind Picard.
\newblock Way off-policy batch deep reinforcement learning of implicit human preferences in dialog.
\newblock \emph{arXiv preprint arXiv:1907.00456}, 2019.

\bibitem[Jelinek et~al.(1977)Jelinek, Mercer, Bahl, and Baker]{jelinek1977perplexity}
Fred Jelinek, Robert~L Mercer, Lalit~R Bahl, and James~K Baker.
\newblock Perplexity?a measure of the difficulty of speech recognition tasks.
\newblock \emph{The Journal of the Acoustical Society of America}, 62\penalty0 (S1):\penalty0 S63--S63, 1977.

\bibitem[Jhamtani et~al.(2017)Jhamtani, Gangal, Hovy, and Nyberg]{jhamtani2017shakespearizing}
Harsh Jhamtani, Varun Gangal, Eduard Hovy, and Eric Nyberg.
\newblock Shakespearizing modern language using copy-enriched sequence-to-sequence models.
\newblock \emph{EMNLP 2017}, 6:\penalty0 10, 2017.

\bibitem[Jiang et~al.(2024)Jiang, Wang, Zeng, Zhong, Li, Mi, Shang, Jiang, Liu, and Wang]{jiang2023followbench}
Yuxin Jiang, Yufei Wang, Xingshan Zeng, Wanjun Zhong, Liangyou Li, Fei Mi, Lifeng Shang, Xin Jiang, Qun Liu, and Wei Wang.
\newblock Followbench: A multi-level fine-grained constraints following benchmark for large language models.
\newblock In \emph{Proceedings of the 62nd Annual Meeting of the Association for Computational Linguistics (Volume 1: Long Papers)}, pp.\  4667--4688, 2024.

\bibitem[John et~al.(2019)John, Mou, Bahuleyan, and Vechtomova]{john2019disentangled}
Vineet John, Lili Mou, Hareesh Bahuleyan, and Olga Vechtomova.
\newblock Disentangled representation learning for non-parallel text style transfer.
\newblock In \emph{Proceedings of the 57th Annual Meeting of the Association for Computational Linguistics}, pp.\  424--434, 2019.

\bibitem[Joshi et~al.(2024)Joshi, Rando, Saparov, Kim, and He]{joshi2023personas}
Nitish Joshi, Javier Rando, Abulhair Saparov, Najoung Kim, and He~He.
\newblock Personas as a way to model truthfulness in language models.
\newblock In \emph{Proceedings of the 2024 Conference on Empirical Methods in Natural Language Processing}, pp.\  6346--6359, 2024.

\bibitem[Kabir et~al.(2023)Kabir, Udo-Imeh, Kou, and Zhang]{kabir2023answers}
Samia Kabir, David~N Udo-Imeh, Bonan Kou, and Tianyi Zhang.
\newblock Who answers it better? an in-depth analysis of chatgpt and stack overflow answers to software engineering questions.
\newblock \emph{arXiv preprint arXiv:2308.02312}, 2023.

\bibitem[Kadavath et~al.(2022)Kadavath, Conerly, Askell, Henighan, Drain, Perez, Schiefer, Dodds, DasSarma, Tran-Johnson, et~al.]{kadavath2022language}
Saurav Kadavath, Tom Conerly, Amanda Askell, Tom Henighan, Dawn Drain, Ethan Perez, Nicholas Schiefer, Zac~Hatfield Dodds, Nova DasSarma, Eli Tran-Johnson, et~al.
\newblock Language models (mostly) know what they know.
\newblock \emph{arXiv preprint arXiv:2207.05221}, 2022.

\bibitem[Kajiwara(2019)]{kajiwara2019negative}
Tomoyuki Kajiwara.
\newblock Negative lexically constrained decoding for paraphrase generation.
\newblock In \emph{Proceedings of the 57th Annual Meeting of the Association for Computational Linguistics}, pp.\  6047--6052, 2019.

\bibitem[Kandpal et~al.(2023)Kandpal, Deng, Roberts, Wallace, and Raffel]{kandpal2023large}
Nikhil Kandpal, Haikang Deng, Adam Roberts, Eric Wallace, and Colin Raffel.
\newblock Large language models struggle to learn long-tail knowledge.
\newblock In \emph{International Conference on Machine Learning}, pp.\  15696--15707. PMLR, 2023.

\bibitem[Kannan \& Vinyals(2017)Kannan and Vinyals]{kannan2017adversarial}
Anjuli Kannan and Oriol Vinyals.
\newblock Adversarial evaluation of dialogue models.
\newblock \emph{arXiv preprint arXiv:1701.08198}, 2017.

\bibitem[Kasai et~al.(2022)Kasai, Sakaguchi, Dunagan, Morrison, Le~Bras, Choi, and Smith]{kasai2022transparent}
Jungo Kasai, Keisuke Sakaguchi, Lavinia Dunagan, Jacob Morrison, Ronan Le~Bras, Yejin Choi, and Noah~A Smith.
\newblock Transparent human evaluation for image captioning.
\newblock In \emph{Proceedings of the 2022 Conference of the North American Chapter of the Association for Computational Linguistics: Human Language Technologies}, pp.\  3464--3478, 2022.

\bibitem[Ke et~al.(2022)Ke, Zhou, Lin, Li, Zhou, Zhu, and Huang]{ke2022ctrleval}
Pei Ke, Hao Zhou, Yankai Lin, Peng Li, Jie Zhou, Xiaoyan Zhu, and Minlie Huang.
\newblock Ctrleval: An unsupervised reference-free metric for evaluating controlled text generation.
\newblock In \emph{Proceedings of the 60th Annual Meeting of the Association for Computational Linguistics (Volume 1: Long Papers)}, pp.\  2306--2319, 2022.

\bibitem[Keskar et~al.(2019)Keskar, McCann, Varshney, Xiong, and Socher]{keskar2019ctrl}
Nitish~Shirish Keskar, Bryan McCann, Lav~R Varshney, Caiming Xiong, and Richard Socher.
\newblock Ctrl: A conditional transformer language model for controllable generation.
\newblock \emph{arXiv preprint arXiv:1909.05858}, 2019.

\bibitem[Khalifa et~al.(2021)Khalifa, Elsahar, and Dymetman]{khalifa2020distributional}
Muhammad Khalifa, Hady Elsahar, and Marc Dymetman.
\newblock A distributional approach to controlled text generation.
\newblock In \emph{International Conference on Learning Representations}, 2021.

\bibitem[Khapra \& Sai(2021)Khapra and Sai]{khapra2021tutorial}
Mitesh~M Khapra and Ananya~B Sai.
\newblock A tutorial on evaluation metrics used in natural language generation.
\newblock In \emph{Proceedings of the 2021 Conference of the North American Chapter of the Association for Computational Linguistics: Human Language Technologies: Tutorials}, pp.\  15--19, 2021.

\bibitem[Kikuchi et~al.(2016)Kikuchi, Neubig, Sasano, Takamura, and Okumura]{kikuchi2016controlling}
Yuta Kikuchi, Graham Neubig, Ryohei Sasano, Hiroya Takamura, and Manabu Okumura.
\newblock Controlling output length in neural encoder-decoders.
\newblock In \emph{EMNLP}, 2016.

\bibitem[Kim et~al.(2017)Kim, Zhang, Rush, LeCun, et~al.]{kim2017adversarially}
Yoon Kim, Kelly Zhang, Alexander~M Rush, Yann LeCun, et~al.
\newblock Adversarially regularized autoencoders for generating discrete structures.
\newblock \emph{arXiv preprint arXiv:1706.04223}, 2:\penalty0 12, 2017.

\bibitem[Kincaid et~al.(1975)Kincaid, Fishburne~Jr, Rogers, and Chissom]{kincaid1975derivation}
J~Peter Kincaid, Robert~P Fishburne~Jr, Richard~L Rogers, and Brad~S Chissom.
\newblock Derivation of new readability formulas (automated readability index, fog count and flesch reading ease formula) for navy enlisted personnel.
\newblock 1975.

\bibitem[Knowles \& Koehn(2016)Knowles and Koehn]{knowles2016neural}
Rebecca Knowles and Philipp Koehn.
\newblock Neural interactive translation prediction.
\newblock In \emph{Twelfth Conference of The Association for Machine Translation in the Americas}, pp.\  107--120. Association for Machine Translation in the Americas, AMTA, 2016.

\bibitem[Kong et~al.(2024)Kong, Wang, Mu, Du, Zhuang, Zhou, Song, Zhang, Wang, and Zhang]{kong2024aligning}
Lingkai Kong, Haorui Wang, Wenhao Mu, Yuanqi Du, Yuchen Zhuang, Yifei Zhou, Yue Song, Rongzhi Zhang, Kai Wang, and Chao Zhang.
\newblock Aligning large language models with representation editing: A control perspective.
\newblock \emph{Advances in Neural Information Processing Systems}, 37:\penalty0 37356--37384, 2024.

\bibitem[Kong et~al.(2019)Kong, Li, Neubig, Hovy, and Yang]{kong2019adversarial}
Xiang Kong, Bohan Li, Graham Neubig, Eduard Hovy, and Yiming Yang.
\newblock An adversarial approach to high-quality, sentiment-controlled neural dialogue generation.
\newblock \emph{arXiv preprint arXiv:1901.07129}, 2019.

\bibitem[Kratzwald et~al.(2019)Kratzwald, Eigenmann, and Feuerriegel]{kratzwald2019rankqa}
Bernhard Kratzwald, Anna Eigenmann, and Stefan Feuerriegel.
\newblock Rankqa: Neural question answering with answer re-ranking.
\newblock In \emph{Proceedings of the 57th Annual Meeting of the Association for Computational Linguistics}, pp.\  6076--6085, 2019.

\bibitem[Krause et~al.(2021)Krause, Gotmare, McCann, Keskar, Joty, Socher, and Rajani]{krause2020gedi}
Ben Krause, Akhilesh~Deepak Gotmare, Bryan McCann, Nitish~Shirish Keskar, Shafiq Joty, Richard Socher, and Nazneen~Fatema Rajani.
\newblock Gedi: Generative discriminator guided sequence generation.
\newblock In \emph{Findings of the Association for Computational Linguistics: EMNLP 2021}, pp.\  4929--4952, 2021.

\bibitem[Kreutzer et~al.(2018)Kreutzer, Uyheng, and Riezler]{kreutzer2018reliability}
Julia Kreutzer, Joshua Uyheng, and Stefan Riezler.
\newblock Reliability and learnability of human bandit feedback for sequence-to-sequence reinforcement learning.
\newblock In \emph{Proceedings of the 56th Annual Meeting of the Association for Computational Linguistics (Volume 1: Long Papers)}, pp.\  1777--1788, 2018.

\bibitem[Kryscinski et~al.(2019)Kryscinski, Keskar, McCann, Xiong, and Socher]{kryscinski2019neural}
Wojciech Kryscinski, Nitish~Shirish Keskar, Bryan McCann, Caiming Xiong, and Richard Socher.
\newblock Neural text summarization: A critical evaluation.
\newblock In \emph{Proceedings of the 2019 Conference on Empirical Methods in Natural Language Processing and the 9th International Joint Conference on Natural Language Processing (EMNLP-IJCNLP)}, pp.\  540--551, 2019.

\bibitem[Kurita et~al.(2019)Kurita, Vyas, Pareek, Black, and Tsvetkov]{kurita2019measuring}
Keita Kurita, Nidhi Vyas, Ayush Pareek, Alan~W Black, and Yulia Tsvetkov.
\newblock Measuring bias in contextualized word representations.
\newblock In \emph{Proceedings of the First Workshop on Gender Bias in Natural Language Processing}, pp.\  166--172, 2019.

\bibitem[Kusner et~al.(2015)Kusner, Sun, Kolkin, and Weinberger]{kusner2015word}
Matt Kusner, Yu~Sun, Nicholas Kolkin, and Kilian Weinberger.
\newblock From word embeddings to document distances.
\newblock In \emph{International conference on machine learning}, pp.\  957--966. PMLR, 2015.

\bibitem[Laban et~al.(2020)Laban, Hsi, Canny, and Hearst]{laban2020summary}
Philippe Laban, Andrew Hsi, John Canny, and Marti~A Hearst.
\newblock The summary loop: Learning to write abstractive summaries without examples.
\newblock In \emph{Proceedings of the 58th Annual Meeting of the Association for Computational Linguistics}, volume~1, 2020.

\bibitem[Lambert et~al.(2024)Lambert, Morrison, Pyatkin, Huang, Ivison, Brahman, Miranda, Liu, Dziri, Lyu, et~al.]{lambert2024t}
Nathan Lambert, Jacob Morrison, Valentina Pyatkin, Shengyi Huang, Hamish Ivison, Faeze Brahman, Lester James~V Miranda, Alisa Liu, Nouha Dziri, Shane Lyu, et~al.
\newblock Tulu 3: Pushing frontiers in open language model post-training.
\newblock \emph{arXiv preprint arXiv:2411.15124}, 2024.

\bibitem[Lample et~al.(2018)Lample, Subramanian, Smith, Denoyer, Ranzato, and Boureau]{lample2018multiple}
Guillaume Lample, Sandeep Subramanian, Eric Smith, Ludovic Denoyer, Marc'Aurelio Ranzato, and Y-Lan Boureau.
\newblock Multiple-attribute text rewriting.
\newblock In \emph{International Conference on Learning Representations}, 2018.

\bibitem[Latif et~al.(2020)Latif, Bashir, Agha, and Latif]{latif2020backward}
Seemab Latif, Sarmad Bashir, Mir Muntasar~Ali Agha, and Rabia Latif.
\newblock Backward-forward sequence generative network for multiple lexical constraints.
\newblock In \emph{IFIP International Conference on Artificial Intelligence Applications and Innovations}, pp.\  39--50. Springer, 2020.

\bibitem[Lee et~al.(2018)Lee, Mansimov, and Cho]{lee2018deterministic}
Jason Lee, Elman Mansimov, and Kyunghyun Cho.
\newblock Deterministic non-autoregressive neural sequence modeling by iterative refinement.
\newblock In \emph{Proceedings of the 2018 Conference on Empirical Methods in Natural Language Processing}, pp.\  1173--1182, 2018.

\bibitem[Lee et~al.(2022)Lee, Liang, and Yang]{lee2022coauthor}
Mina Lee, Percy Liang, and Qian Yang.
\newblock Coauthor: Designing a human-ai collaborative writing dataset for exploring language model capabilities.
\newblock In \emph{Proceedings of the 2022 CHI conference on human factors in computing systems}, pp.\  1--19, 2022.

\bibitem[Leiter et~al.(2022)Leiter, Lertvittayakumjorn, Fomicheva, Zhao, Gao, and Eger]{leiter2022towards}
Christoph Leiter, Piyawat Lertvittayakumjorn, Marina Fomicheva, Wei Zhao, Yang Gao, and Steffen Eger.
\newblock Towards explainable evaluation metrics for natural language generation.
\newblock \emph{arXiv preprint arXiv:2203.11131}, 2022.

\bibitem[Lester et~al.(2021)Lester, Al-Rfou, and Constant]{lester2021power}
Brian Lester, Rami Al-Rfou, and Noah Constant.
\newblock The power of scale for parameter-efficient prompt tuning.
\newblock In \emph{Proceedings of the 2021 Conference on Empirical Methods in Natural Language Processing}, pp.\  3045--3059, 2021.

\bibitem[Li et~al.(2024{\natexlab{a}})Li, Wang, Meng, Chang, and Peng]{li2024control}
Bingxuan Li, Yiwei Wang, Tao Meng, Kai-Wei Chang, and Nanyun Peng.
\newblock Control large language models via divide and conquer.
\newblock In \emph{Proceedings of the 2024 Conference on Empirical Methods in Natural Language Processing}, pp.\  15240--15256, 2024{\natexlab{a}}.

\bibitem[Li et~al.(2016{\natexlab{a}})Li, Galley, Brockett, Gao, and Dolan]{li2016diversity}
Jiwei Li, Michel Galley, Chris Brockett, Jianfeng Gao, and Bill Dolan.
\newblock A diversity-promoting objective function for neural conversation models.
\newblock In \emph{Proceedings of NAACL-HLT}, pp.\  110--119, 2016{\natexlab{a}}.

\bibitem[Li et~al.(2016{\natexlab{b}})Li, Galley, Brockett, Spithourakis, Gao, and Dolan]{li2016persona}
Jiwei Li, Michel Galley, Chris Brockett, Georgios Spithourakis, Jianfeng Gao, and William~B Dolan.
\newblock A persona-based neural conversation model.
\newblock In \emph{Proceedings of the 54th Annual Meeting of the Association for Computational Linguistics (Volume 1: Long Papers)}, pp.\  994--1003, 2016{\natexlab{b}}.

\bibitem[Li et~al.(2016{\natexlab{c}})Li, Monroe, Ritter, Jurafsky, Galley, and Gao]{li2016deep}
Jiwei Li, Will Monroe, Alan Ritter, Dan Jurafsky, Michel Galley, and Jianfeng Gao.
\newblock Deep reinforcement learning for dialogue generation.
\newblock In \emph{Proceedings of the 2016 Conference on Empirical Methods in Natural Language Processing}, pp.\  1192--1202, 2016{\natexlab{c}}.

\bibitem[Li et~al.(2017)Li, Monroe, Shi, Jean, Ritter, and Jurafsky]{li2017adversarial}
Jiwei Li, Will Monroe, Tianlin Shi, S{\'e}bastien Jean, Alan Ritter, and Dan Jurafsky.
\newblock Adversarial learning for neural dialogue generation.
\newblock In \emph{Proceedings of the 2017 Conference on Empirical Methods in Natural Language Processing}, pp.\  2157--2169, 2017.

\bibitem[Li et~al.(2018)Li, Jia, He, and Liang]{li2018delete}
Juncen Li, Robin Jia, He~He, and Percy Liang.
\newblock Delete, retrieve, generate: a simple approach to sentiment and style transfer.
\newblock In \emph{Proceedings of the 2018 Conference of the North American Chapter of the Association for Computational Linguistics: Human Language Technologies, Volume 1 (Long Papers)}, pp.\  1865--1874, 2018.

\bibitem[Li et~al.(2024{\natexlab{b}})Li, Peris, Mehrabi, Goyal, Chang, Galstyan, Zemel, and Gupta]{li2024steerability}
Junyi Li, Charith Peris, Ninareh Mehrabi, Palash Goyal, Kai-Wei Chang, Aram Galstyan, Richard Zemel, and Rahul Gupta.
\newblock The steerability of large language models toward data-driven personas.
\newblock In \emph{Proceedings of the 2024 Conference of the North American Chapter of the Association for Computational Linguistics: Human Language Technologies (Volume 1: Long Papers)}, pp.\  7283--7298, 2024{\natexlab{b}}.

\bibitem[Li et~al.()Li, Liu, Bashkansky, Bau, Vi{\'e}gas, Pfister, and Wattenberg]{li2024measuring}
Kenneth Li, Tianle Liu, Naomi Bashkansky, David Bau, Fernanda Vi{\'e}gas, Hanspeter Pfister, and Martin Wattenberg.
\newblock Measuring and controlling instruction (in) stability in language model dialogs.
\newblock In \emph{First Conference on Language Modeling}.

\bibitem[Li et~al.(2023{\natexlab{a}})Li, Hopkins, Bau, Vi{\'e}gas, Pfister, and Wattenberg]{li2022emergent}
Kenneth Li, Aspen~K Hopkins, David Bau, Fernanda Vi{\'e}gas, Hanspeter Pfister, and Martin Wattenberg.
\newblock Emergent world representations: Exploring a sequence model trained on a synthetic task.
\newblock In \emph{The Eleventh International Conference on Learning Representations}, 2023{\natexlab{a}}.

\bibitem[Li et~al.(2024{\natexlab{c}})Li, Patel, Vi{\'e}gas, Pfister, and Wattenberg]{li2024inference}
Kenneth Li, Oam Patel, Fernanda Vi{\'e}gas, Hanspeter Pfister, and Martin Wattenberg.
\newblock Inference-time intervention: Eliciting truthful answers from a language model.
\newblock \emph{Advances in Neural Information Processing Systems}, 36, 2024{\natexlab{c}}.

\bibitem[Li et~al.(2022)Li, Thickstun, Gulrajani, Liang, and Hashimoto]{li2022diffusion}
Xiang Li, John Thickstun, Ishaan Gulrajani, Percy~S Liang, and Tatsunori~B Hashimoto.
\newblock Diffusion-lm improves controllable text generation.
\newblock \emph{Advances in Neural Information Processing Systems}, 35:\penalty0 4328--4343, 2022.

\bibitem[Li \& Liang(2021)Li and Liang]{li2021prefix}
Xiang~Lisa Li and Percy Liang.
\newblock Prefix-tuning: Optimizing continuous prompts for generation.
\newblock In \emph{Proceedings of the 59th Annual Meeting of the Association for Computational Linguistics and the 11th International Joint Conference on Natural Language Processing (Volume 1: Long Papers)}, pp.\  4582--4597, 2021.

\bibitem[Li et~al.(2023{\natexlab{b}})Li, Zhang, Dubois, Taori, Gulrajani, Guestrin, Liang, and Hashimoto]{li2023alpacaeval}
Xuechen Li, Tianyi Zhang, Yann Dubois, Rohan Taori, Ishaan Gulrajani, Carlos Guestrin, Percy Liang, and Tatsunori~B Hashimoto.
\newblock Alpacaeval: An automatic evaluator of instruction-following models, 2023{\natexlab{b}}.

\bibitem[Li et~al.(2023{\natexlab{c}})Li, Peng, He, Galley, Gao, and Yan]{li2023guiding}
Zekun Li, Baolin Peng, Pengcheng He, Michel Galley, Jianfeng Gao, and Xifeng Yan.
\newblock Guiding large language models via directional stimulus prompting.
\newblock \emph{Advances in Neural Information Processing Systems}, 36:\penalty0 62630--62656, 2023{\natexlab{c}}.

\bibitem[Li et~al.(2023{\natexlab{d}})Li, Arous, Reddy, and Cheung]{li2023evaluating}
Zichao Li, Ines Arous, Siva Reddy, and Jackie Chi~Kit Cheung.
\newblock Evaluating dependencies in fact editing for language models: Specificity and implication awareness.
\newblock In \emph{Findings of the Association for Computational Linguistics: EMNLP 2023}, pp.\  7623--7636, 2023{\natexlab{d}}.

\bibitem[Liang et~al.(2023)Liang, Bommasani, Lee, Tsipras, Soylu, Yasunaga, Zhang, Narayanan, Wu, Kumar, et~al.]{liang2022holistic}
Percy Liang, Rishi Bommasani, Tony Lee, Dimitris Tsipras, Dilara Soylu, Michihiro Yasunaga, Yian Zhang, Deepak Narayanan, Yuhuai Wu, Ananya Kumar, et~al.
\newblock Holistic evaluation of language models.
\newblock \emph{Transactions on Machine Learning Research}, 2023.

\bibitem[Liang et~al.(2024)Liang, Chen, Wang, Wu, Fu, Shi, Wu, and Ye]{liang2024robust}
Xize Liang, Chao Chen, Jie Wang, Yue Wu, Zhihang Fu, Zhihao Shi, Feng Wu, and Jieping Ye.
\newblock Robust preference optimization with provable noise tolerance for llms.
\newblock \emph{arXiv preprint arXiv:2404.04102}, 2024.

\bibitem[Lin et~al.(2019)Lin, Shen, Xing, Zhou, and Ren]{lin2019commongen}
Bill~Yuchen Lin, Ming Shen, Yu~Xing, Pei Zhou, and Xiang Ren.
\newblock Commongen: A constrained text generation dataset towards generative commonsense reasoning.
\newblock \emph{arXiv preprint arXiv:1911.03705}, 2019.

\bibitem[Lin et~al.(2020)Lin, Zhou, Shen, Zhou, Bhagavatula, Choi, and Ren]{lin2020commongen}
Bill~Yuchen Lin, Wangchunshu Zhou, Ming Shen, Pei Zhou, Chandra Bhagavatula, Yejin Choi, and Xiang Ren.
\newblock Commongen: A constrained text generation challenge for generative commonsense reasoning.
\newblock In \emph{Findings of the Association for Computational Linguistics: EMNLP 2020}, pp.\  1823--1840, 2020.

\bibitem[Lin et~al.(2024{\natexlab{a}})Lin, Ravichander, Lu, Dziri, Sclar, Chandu, Bhagavatula, and Choi]{lin2023unlocking}
Bill~Yuchen Lin, Abhilasha Ravichander, Ximing Lu, Nouha Dziri, Melanie Sclar, Khyathi Chandu, Chandra Bhagavatula, and Yejin Choi.
\newblock The unlocking spell on base llms: Rethinking alignment via in-context learning.
\newblock In \emph{The Twelfth International Conference on Learning Representations}, 2024{\natexlab{a}}.

\bibitem[Lin et~al.(2025)Lin, Bras, Richardson, Sabharwal, Poovendran, Clark, and Choi]{lin2025zebralogic}
Bill~Yuchen Lin, Ronan~Le Bras, Kyle Richardson, Ashish Sabharwal, Radha Poovendran, Peter Clark, and Yejin Choi.
\newblock Zebralogic: On the scaling limits of llms for logical reasoning.
\newblock \emph{arXiv preprint arXiv:2502.01100}, 2025.

\bibitem[Lin(2004)]{lin2004rouge}
Chin-Yew Lin.
\newblock Rouge: A package for automatic evaluation of summaries.
\newblock \emph{Text Summarization Branches Out}, 2004.

\bibitem[Lin et~al.(2022)Lin, Hilton, and Evans]{lin2022truthfulqa}
Stephanie Lin, Jacob Hilton, and Owain Evans.
\newblock Truthfulqa: Measuring how models mimic human falsehoods.
\newblock In \emph{Proceedings of the 60th Annual Meeting of the Association for Computational Linguistics (Volume 1: Long Papers)}, pp.\  3214--3252, 2022.

\bibitem[Lin et~al.(2024{\natexlab{b}})Lin, Chan, Song, and Liu]{lin2024constrained}
Zizheng Lin, Chunkit Chan, Yangqiu Song, and Xin Liu.
\newblock Constrained reasoning chains for enhancing theory-of-mind in large language models.
\newblock In \emph{Pacific Rim International Conference on Artificial Intelligence}, pp.\  354--360. Springer, 2024{\natexlab{b}}.

\bibitem[Lindner et~al.(2024)Lindner, Chen, Tschiatschek, Hofmann, and Krause]{lindner2024learning}
David Lindner, Xin Chen, Sebastian Tschiatschek, Katja Hofmann, and Andreas Krause.
\newblock Learning safety constraints from demonstrations with unknown rewards.
\newblock In \emph{International Conference on Artificial Intelligence and Statistics}, pp.\  2386--2394. PMLR, 2024.

\bibitem[Liu et~al.(2021{\natexlab{a}})Liu, Sap, Lu, Swayamdipta, Bhagavatula, Smith, and Choi]{liu2021dexperts}
Alisa Liu, Maarten Sap, Ximing Lu, Swabha Swayamdipta, Chandra Bhagavatula, Noah~A Smith, and Yejin Choi.
\newblock Dexperts: Decoding-time controlled text generation with experts and anti-experts.
\newblock In \emph{Proceedings of the 59th Annual Meeting of the Association for Computational Linguistics and the 11th International Joint Conference on Natural Language Processing (Volume 1: Long Papers)}, 2021{\natexlab{a}}.

\bibitem[Liu et~al.(2016)Liu, Lowe, Serban, Noseworthy, Charlin, and Pineau]{liu2016not}
Chia-Wei Liu, Ryan Lowe, Iulian Serban, Mike Noseworthy, Laurent Charlin, and Joelle Pineau.
\newblock How not to evaluate your dialogue system: An empirical study of unsupervised evaluation metrics for dialogue response generation.
\newblock In \emph{Proceedings of the 2016 Conference on Empirical Methods in Natural Language Processing}, pp.\  2122--2132, 2016.

\bibitem[Liu et~al.(2019{\natexlab{a}})Liu, Fu, Qu, and Lv]{liu2019bfgan}
Dayiheng Liu, Jie Fu, Qian Qu, and Jiancheng Lv.
\newblock Bfgan: Backward and forward generative adversarial networks for lexically constrained sentence generation.
\newblock \emph{IEEE/ACM Transactions on Audio, Speech, and Language Processing}, 27\penalty0 (12):\penalty0 2350--2361, 2019{\natexlab{a}}.

\bibitem[Liu et~al.(2025{\natexlab{a}})Liu, Fu, Ding, Ning, Zhang, Liu, and Zhang]{liu2025logical}
Hanmeng Liu, Zhizhang Fu, Mengru Ding, Ruoxi Ning, Chaoli Zhang, Xiaozhang Liu, and Yue Zhang.
\newblock Logical reasoning in large language models: A survey.
\newblock \emph{arXiv preprint arXiv:2502.09100}, 2025{\natexlab{a}}.

\bibitem[Liu et~al.(2024{\natexlab{a}})Liu, Cohen, Pasunuru, Choi, Hajishirzi, and Celikyilmaz]{liu2024don}
Jiacheng Liu, Andrew Cohen, Ramakanth Pasunuru, Yejin Choi, Hannaneh Hajishirzi, and Asli Celikyilmaz.
\newblock Don't throw away your value model! generating more preferable text with value-guided monte-carlo tree search decoding.
\newblock In \emph{First Conference on Language Modeling}, 2024{\natexlab{a}}.

\bibitem[Liu et~al.(2025{\natexlab{b}})Liu, Qiu, Li, Dai, Zhu, Hu, Yang, and King]{liu2025surveypersonalizedlargelanguage}
Jiahong Liu, Zexuan Qiu, Zhongyang Li, Quanyu Dai, Jieming Zhu, Minda Hu, Menglin Yang, and Irwin King.
\newblock A survey of personalized large language models: Progress and future directions, 2025{\natexlab{b}}.
\newblock URL \url{https://arxiv.org/abs/2502.11528}.

\bibitem[Liu et~al.(2024{\natexlab{b}})Liu, Yu, Zhang, Li, Zhang, and Ji]{liu2024evedit}
Jiateng Liu, Pengfei Yu, Yuji Zhang, Sha Li, Zixuan Zhang, and Heng Ji.
\newblock Evedit: Event-based knowledge editing with deductive editing boundaries.
\newblock \emph{arXiv preprint arXiv:2402.11324}, 2024{\natexlab{b}}.

\bibitem[Liu et~al.(2024{\natexlab{c}})Liu, Lin, Hewitt, Paranjape, Bevilacqua, Petroni, and Liang]{liu2023lost}
Nelson~F Liu, Kevin Lin, John Hewitt, Ashwin Paranjape, Michele Bevilacqua, Fabio Petroni, and Percy Liang.
\newblock Lost in the middle: How language models use long contexts.
\newblock \emph{Transactions of the Association for Computational Linguistics}, 12:\penalty0 157--173, 2024{\natexlab{c}}.

\bibitem[Liu et~al.(2023{\natexlab{a}})Liu, Yuan, Fu, Jiang, Hayashi, and Neubig]{liu2023pre}
Pengfei Liu, Weizhe Yuan, Jinlan Fu, Zhengbao Jiang, Hiroaki Hayashi, and Graham Neubig.
\newblock Pre-train, prompt, and predict: A systematic survey of prompting methods in natural language processing.
\newblock \emph{ACM Computing Surveys}, 55\penalty0 (9):\penalty0 1--35, 2023{\natexlab{a}}.

\bibitem[Liu et~al.(2018)Liu, Saleh, Pot, Goodrich, Sepassi, Kaiser, and Shazeer]{liu2018generating}
Peter~J Liu, Mohammad Saleh, Etienne Pot, Ben Goodrich, Ryan Sepassi, Lukasz Kaiser, and Noam Shazeer.
\newblock Generating wikipedia by summarizing long sequences.
\newblock In \emph{International Conference on Learning Representations}, 2018.

\bibitem[Liu et~al.(2020)Liu, Xu, Jia, Ma, Wang, and Vosoughi]{liu2020data}
Ruibo Liu, Guangxuan Xu, Chenyan Jia, Weicheng Ma, Lili Wang, and Soroush Vosoughi.
\newblock Data boost: Text data augmentation through reinforcement learning guided conditional generation.
\newblock In \emph{Proceedings of the 2020 Conference on Empirical Methods in Natural Language Processing (EMNLP)}, pp.\  9031--9041, 2020.

\bibitem[Liu et~al.(2024{\natexlab{d}})Liu, Zhao, Joshi, Khalman, Saleh, Liu, and Liu]{liu2023statistical}
Tianqi Liu, Yao Zhao, Rishabh Joshi, Misha Khalman, Mohammad Saleh, Peter~J Liu, and Jialu Liu.
\newblock Statistical rejection sampling improves preference optimization.
\newblock In \emph{The Twelfth International Conference on Learning Representations}, 2024{\natexlab{d}}.

\bibitem[Liu et~al.(2023{\natexlab{b}})Liu, Khalifa, and Wang]{liu2023bolt}
Xin Liu, Muhammad Khalifa, and Lu~Wang.
\newblock Bolt: Fast energy-based controlled text generation with tunable biases.
\newblock In \emph{The 61st Annual Meeting Of The Association For Computational Linguistics}, 2023{\natexlab{b}}.

\bibitem[Liu et~al.(2023{\natexlab{c}})Liu, Iter, Xu, Wang, Xu, and Zhu]{liu2023gpteval}
Yang Liu, Dan Iter, Yichong Xu, Shuohang Wang, Ruochen Xu, and Chenguang Zhu.
\newblock G-eval: Nlg evaluation using gpt-4 with better human alignment.
\newblock In \emph{Proceedings of the 2023 Conference on Empirical Methods in Natural Language Processing}, pp.\  2511--2522, 2023{\natexlab{c}}.

\bibitem[Liu et~al.(2021{\natexlab{b}})Liu, Zhang, Han, Zhang, and Tu]{liu2021constrained}
Yixian Liu, Liwen Zhang, Wenjuan Han, Yue Zhang, and Kewei Tu.
\newblock Constrained text generation with global guidance--case study on commongen.
\newblock \emph{arXiv preprint arXiv:2103.07170}, 2021{\natexlab{b}}.

\bibitem[Liu et~al.(2022)Liu, Bawden, Scaliom, Sagot, and Cheung]{liu2022maskeval}
Yu~Lu Liu, Rachel Bawden, Thomas Scaliom, Beno{\^\i}t Sagot, and Jackie Chi~Kit Cheung.
\newblock Maskeval: Weighted mlm-based evaluation for text summarization and simplification.
\newblock \emph{arXiv preprint arXiv:2205.12394}, 2022.

\bibitem[Liu et~al.(2019{\natexlab{b}})Liu, Fu, Cao, de~Melo, Tam, Niu, and Zhou]{liu2019rhetorically}
Zhiqiang Liu, Zuohui Fu, Jie Cao, Gerard de~Melo, Yik-Cheung Tam, Cheng Niu, and Jie Zhou.
\newblock Rhetorically controlled encoder-decoder for modern chinese poetry generation.
\newblock In \emph{Proceedings of the 57th Annual Meeting of the Association for Computational Linguistics}, pp.\  1992--2001, 2019{\natexlab{b}}.

\bibitem[Logeswaran et~al.(2018)Logeswaran, Lee, and Bengio]{logeswaran2018content}
Lajanugen Logeswaran, Honglak Lee, and Samy Bengio.
\newblock Content preserving text generation with attribute controls.
\newblock In \emph{Advances in Neural Information Processing Systems}, pp.\  5103--5113, 2018.

\bibitem[Loginova et~al.(2018)Loginova, Varanasi, and Neumann]{loginova2018towards}
Ekaterina Loginova, Stalin Varanasi, and G{\"u}nter Neumann.
\newblock Towards multilingual neural question answering.
\newblock In \emph{European Conference on Advances in Databases and Information Systems}, pp.\  274--285. Springer, 2018.

\bibitem[Lu et~al.(2021)Lu, West, Zellers, Le~Bras, Bhagavatula, and Choi]{lu2021neurologic}
Ximing Lu, Peter West, Rowan Zellers, Ronan Le~Bras, Chandra Bhagavatula, and Yejin Choi.
\newblock Neurologic decoding:(un) supervised neural text generation with predicate logic constraints.
\newblock In \emph{Proceedings of the 2021 Conference of the North American Chapter of the Association for Computational Linguistics: Human Language Technologies}, pp.\  4288--4299, 2021.

\bibitem[Lu et~al.(2022)Lu, Welleck, Hessel, Jiang, Qin, West, Ammanabrolu, and Choi]{lu2022quark}
Ximing Lu, Sean Welleck, Jack Hessel, Liwei Jiang, Lianhui Qin, Peter West, Prithviraj Ammanabrolu, and Yejin Choi.
\newblock Quark: Controllable text generation with reinforced unlearning.
\newblock \emph{Advances in neural information processing systems}, 35:\penalty0 27591--27609, 2022.

\bibitem[Lucic et~al.(2018)Lucic, Kurach, Michalski, Gelly, and Bousquet]{lucic2018gans}
Mario Lucic, Karol Kurach, Marcin Michalski, Sylvain Gelly, and Olivier Bousquet.
\newblock Are gans created equal? a large-scale study.
\newblock In \emph{Advances in neural information processing systems}, pp.\  700--709, 2018.

\bibitem[Luo et~al.(2024)Luo, Zhao, Gong, Haffari, and Pan]{luo2024graph}
Linhao Luo, Zicheng Zhao, Chen Gong, Gholamreza Haffari, and Shirui Pan.
\newblock Graph-constrained reasoning: Faithful reasoning on knowledge graphs with large language models.
\newblock \emph{arXiv preprint arXiv:2410.13080}, 2024.

\bibitem[Lyu et~al.(2021)Lyu, Liang, Pham, Hovy, P{\'o}czos, Salakhutdinov, and Morency]{lyu2021styleptb}
Yiwei Lyu, Paul~Pu Liang, Hai Pham, Eduard Hovy, Barnab{\'a}s P{\'o}czos, Ruslan Salakhutdinov, and Louis-Philippe Morency.
\newblock Styleptb: A compositional benchmark for fine-grained controllable text style transfer.
\newblock In \emph{Proceedings of the 2021 Conference of the North American Chapter of the Association for Computational Linguistics: Human Language Technologies}, pp.\  2116--2138, 2021.

\bibitem[Madaan et~al.(2020)Madaan, Setlur, Parekh, P{\'o}czos, Neubig, Yang, Salakhutdinov, Black, and Prabhumoye]{madaan2020politeness}
Aman Madaan, Amrith Setlur, Tanmay Parekh, Barnab{\'a}s P{\'o}czos, Graham Neubig, Yiming Yang, Ruslan Salakhutdinov, Alan~W Black, and Shrimai Prabhumoye.
\newblock Politeness transfer: A tag and generate approach.
\newblock In \emph{Proceedings of the 58th Annual Meeting of the Association for Computational Linguistics}, pp.\  1869--1881, 2020.

\bibitem[Madotto et~al.(2020)Madotto, Ishii, Lin, Dathathri, and Fung]{madotto2020plug}
Andrea Madotto, Etsuko Ishii, Zhaojiang Lin, Sumanth Dathathri, and Pascale Fung.
\newblock Plug-and-play conversational models.
\newblock In \emph{Proceedings of the 2020 Conference on Empirical Methods in Natural Language Processing: Findings}, pp.\  2422--2433, 2020.

\bibitem[Magar \& Schwartz(2022)Magar and Schwartz]{magar2022data}
Inbal Magar and Roy Schwartz.
\newblock Data contamination: From memorization to exploitation.
\newblock In \emph{Proceedings of the 60th Annual Meeting of the Association for Computational Linguistics (Volume 2: Short Papers)}, pp.\  157--165, 2022.

\bibitem[Malik et~al.(2021)Malik, Anwar, Aghasi, and Ahmed]{malik2021inverse}
Shehryar Malik, Usman Anwar, Alireza Aghasi, and Ali Ahmed.
\newblock Inverse constrained reinforcement learning.
\newblock In \emph{International conference on machine learning}, pp.\  7390--7399. PMLR, 2021.

\bibitem[Martin et~al.(2020)Martin, de~la Clergerie, Sagot, and Bordes]{martin2019controllable}
Louis Martin, {\'E}ric~Villemonte de~la Clergerie, Beno{\^\i}t Sagot, and Antoine Bordes.
\newblock Controllable sentence simplification.
\newblock In \emph{Proceedings of The 12th Language Resources and Evaluation Conference}, pp.\  4689--4698, 2020.

\bibitem[Mathew et~al.(2021)Mathew, Saha, Yimam, Biemann, Goyal, and Mukherjee]{mathew2021hatexplain}
Binny Mathew, Punyajoy Saha, Seid~Muhie Yimam, Chris Biemann, Pawan Goyal, and Animesh Mukherjee.
\newblock Hatexplain: A benchmark dataset for explainable hate speech detection.
\newblock In \emph{Proceedings of the AAAI conference on artificial intelligence}, volume~35, pp.\  14867--14875, 2021.

\bibitem[Mathur et~al.(2019)Mathur, Baldwin, and Cohn]{mathur2019putting}
Nitika Mathur, Timothy Baldwin, and Trevor Cohn.
\newblock Putting evaluation in context: Contextual embeddings improve machine translation evaluation.
\newblock In \emph{Proceedings of the 57th Annual Meeting of the Association for Computational Linguistics}, pp.\  2799--2808, 2019.

\bibitem[Maynez et~al.(2020)Maynez, Narayan, Bohnet, and McDonald]{maynez2020faithfulness}
Joshua Maynez, Shashi Narayan, Bernd Bohnet, and Ryan McDonald.
\newblock On faithfulness and factuality in abstractive summarization.
\newblock In \emph{Proceedings of the 58th Annual Meeting of the Association for Computational Linguistics}, pp.\  1906--1919, 2020.

\bibitem[McIntosh et~al.(2024)McIntosh, Susnjak, Liu, Watters, and Halgamuge]{mcintosh2024inadequacies}
Timothy~R McIntosh, Teo Susnjak, Tong Liu, Paul Watters, and Malka~N Halgamuge.
\newblock Inadequacies of large language model benchmarks in the era of generative artificial intelligence.
\newblock \emph{arXiv preprint arXiv:2402.09880}, 2024.

\bibitem[McKinney et~al.(2023)McKinney, Duan, Krueger, and Gleave]{mckinney2023fragility}
Lev McKinney, Yawen Duan, David Krueger, and Adam Gleave.
\newblock On the fragility of learned reward functions.
\newblock \emph{arXiv preprint arXiv:2301.03652}, 2023.

\bibitem[Mei et~al.(2017)Mei, Bansal, and Walter]{mei2017coherent}
Hongyuan Mei, Mohit Bansal, and Matthew~R Walter.
\newblock Coherent dialogue with attention-based language models.
\newblock In \emph{Thirty-First AAAI Conference on Artificial Intelligence}, 2017.

\bibitem[Meng et~al.()Meng, Sharma, Andonian, Belinkov, and Bau]{meng2022mass}
Kevin Meng, Arnab~Sen Sharma, Alex~J Andonian, Yonatan Belinkov, and David Bau.
\newblock Mass-editing memory in a transformer.
\newblock In \emph{The Eleventh International Conference on Learning Representations}.

\bibitem[Meng et~al.(2022)Meng, Bau, Andonian, and Belinkov]{meng2022locating}
Kevin Meng, David Bau, Alex Andonian, and Yonatan Belinkov.
\newblock Locating and editing factual associations in gpt.
\newblock \emph{Advances in Neural Information Processing Systems}, 35:\penalty0 17359--17372, 2022.

\bibitem[Miao et~al.(2019)Miao, Zhou, Mou, Yan, and Li]{miao2019cgmh}
Ning Miao, Hao Zhou, Lili Mou, Rui Yan, and Lei Li.
\newblock Cgmh: Constrained sentence generation by metropolis-hastings sampling.
\newblock In \emph{Proceedings of the AAAI Conference on Artificial Intelligence}, volume~33, pp.\  6834--6842, 2019.

\bibitem[Miao \& Blunsom(2016)Miao and Blunsom]{miao2016language}
Yishu Miao and Phil Blunsom.
\newblock Language as a latent variable: Discrete generative models for sentence compression.
\newblock In \emph{Proceedings of the 2016 Conference on Empirical Methods in Natural Language Processing}, pp.\  319--328, 2016.

\bibitem[Min et~al.(2023)Min, Krishna, Lyu, Lewis, Yih, Koh, Iyyer, Zettlemoyer, and Hajishirzi]{min2023factscore}
Sewon Min, Kalpesh Krishna, Xinxi Lyu, Mike Lewis, Wen-tau Yih, Pang Koh, Mohit Iyyer, Luke Zettlemoyer, and Hannaneh Hajishirzi.
\newblock Factscore: Fine-grained atomic evaluation of factual precision in long form text generation.
\newblock In \emph{Proceedings of the 2023 Conference on Empirical Methods in Natural Language Processing}, pp.\  12076--12100, 2023.

\bibitem[Mireshghallah et~al.(2022)Mireshghallah, Goyal, and Berg-Kirkpatrick]{mireshghallah2022mix}
Fatemehsadat Mireshghallah, Kartik Goyal, and Taylor Berg-Kirkpatrick.
\newblock Mix and match: Learning-free controllable text generationusing energy language models.
\newblock In \emph{Proceedings of the 60th Annual Meeting of the Association for Computational Linguistics (Volume 1: Long Papers)}, pp.\  401--415, 2022.

\bibitem[Mitchell et~al.(2022)Mitchell, Lin, Bosselut, Finn, and Manning]{mitchell2021fast}
Eric Mitchell, Charles Lin, Antoine Bosselut, Chelsea Finn, and Christopher~D Manning.
\newblock Fast model editing at scale.
\newblock In \emph{International Conference on Learning Representations}, 2022.

\bibitem[Mou et~al.(2015)Mou, Yan, Li, Zhang, and Jin]{mou2015backward}
Lili Mou, Rui Yan, Ge~Li, Lu~Zhang, and Zhi Jin.
\newblock Backward and forward language modeling for constrained sentence generation.
\newblock \emph{arXiv preprint arXiv:1512.06612}, 2015.

\bibitem[Mou et~al.(2016)Mou, Song, Yan, Li, Zhang, and Jin]{mou2016sequence}
Lili Mou, Yiping Song, Rui Yan, Ge~Li, Lu~Zhang, and Zhi Jin.
\newblock Sequence to backward and forward sequences: A content-introducing approach to generative short-text conversation.
\newblock In \emph{Proceedings of COLING 2016, the 26th International Conference on Computational Linguistics: Technical Papers}, pp.\  3349--3358, 2016.

\bibitem[Mueller et~al.(2017)Mueller, Gifford, and Jaakkola]{mueller2017sequence}
Jonas Mueller, David Gifford, and Tommi Jaakkola.
\newblock Sequence to better sequence: continuous revision of combinatorial structures.
\newblock In \emph{Proceedings of the 34th International Conference on Machine Learning-Volume 70}, pp.\  2536--2544. JMLR. org, 2017.

\bibitem[Muhlgay et~al.(2024)Muhlgay, Ram, Magar, Levine, Ratner, Belinkov, Abend, Leyton-Brown, Shashua, and Shoham]{muhlgay2024generating}
Dor Muhlgay, Ori Ram, Inbal Magar, Yoav Levine, Nir Ratner, Yonatan Belinkov, Omri Abend, Kevin Leyton-Brown, Amnon Shashua, and Yoav Shoham.
\newblock Generating benchmarks for factuality evaluation of language models.
\newblock In \emph{Proceedings of the 18th Conference of the European Chapter of the Association for Computational Linguistics (Volume 1: Long Papers)}, pp.\  49--66, 2024.

\bibitem[Nakano et~al.(2021)Nakano, Hilton, Balaji, Wu, Ouyang, Kim, Hesse, Jain, Kosaraju, Saunders, et~al.]{nakano2021webgpt}
Reiichiro Nakano, Jacob Hilton, Suchir Balaji, Jeff Wu, Long Ouyang, Christina Kim, Christopher Hesse, Shantanu Jain, Vineet Kosaraju, William Saunders, et~al.
\newblock Webgpt: Browser-assisted question-answering with human feedback.
\newblock \emph{arXiv preprint arXiv:2112.09332}, 2021.

\bibitem[Nallapati et~al.(2016)Nallapati, Zhou, Gulcehre, Xiang, et~al.]{nallapati2016abstractive}
Ramesh Nallapati, Bowen Zhou, Caglar Gulcehre, Bing Xiang, et~al.
\newblock Abstractive text summarization using sequence-to-sequence rnns and beyond.
\newblock In \emph{Proceedings of the 20th SIGNLL Conference on Computational Natural Language Learning}, pp.\  280--290, 2016.

\bibitem[Nallapati et~al.(2017)Nallapati, Zhai, and Zhou]{nallapati2017summarunner}
Ramesh Nallapati, Feifei Zhai, and Bowen Zhou.
\newblock Summarunner: A recurrent neural network based sequence model for extractive summarization of documents.
\newblock In \emph{Thirty-First AAAI Conference on Artificial Intelligence}, 2017.

\bibitem[Napoles et~al.(2015)Napoles, Sakaguchi, Post, and Tetreault]{napoles2015ground}
Courtney Napoles, Keisuke Sakaguchi, Matt Post, and Joel Tetreault.
\newblock Ground truth for grammatical error correction metrics.
\newblock In \emph{Proceedings of the 53rd Annual Meeting of the Association for Computational Linguistics and the 7th International Joint Conference on Natural Language Processing (Volume 2: Short Papers)}, pp.\  588--593, 2015.

\bibitem[Nikolov \& Hahnloser(2020)Nikolov and Hahnloser]{nikolov2020abstractive}
Nikola~I Nikolov and Richard Hahnloser.
\newblock Abstractive document summarization without parallel data.
\newblock In \emph{Proceedings of The 12th Language Resources and Evaluation Conference}, pp.\  6638--6644, 2020.

\bibitem[Nishihara et~al.(2019)Nishihara, Kajiwara, and Arase]{nishihara2019controllable}
Daiki Nishihara, Tomoyuki Kajiwara, and Yuki Arase.
\newblock Controllable text simplification with lexical constraint loss.
\newblock In \emph{Proceedings of the 57th Annual Meeting of the Association for Computational Linguistics: Student Research Workshop}, pp.\  260--266, 2019.

\bibitem[Niu \& Bansal(2018)Niu and Bansal]{niu2018polite}
Tong Niu and Mohit Bansal.
\newblock Polite dialogue generation without parallel data.
\newblock \emph{Transactions of the Association for Computational Linguistics}, 6:\penalty0 373--389, 2018.

\bibitem[Oliveira(2017)]{oliveira2017survey}
Hugo~Gon{\c{c}}alo Oliveira.
\newblock A survey on intelligent poetry generation: Languages, features, techniques, reutilisation and evaluation.
\newblock In \emph{Proceedings of the 10th International Conference on Natural Language Generation}, pp.\  11--20, 2017.

\bibitem[OpenAI(2022)]{chatgpt2022}
OpenAI.
\newblock Introducing chatgpt.
\newblock \emph{\url{https://openai.com/blog/chatgpt}}, 2022.

\bibitem[Oren et~al.(2024)Oren, Meister, Chatterji, Ladhak, and Hashimoto]{oren2023proving}
Yonatan Oren, Nicole Meister, Niladri~S Chatterji, Faisal Ladhak, and Tatsunori Hashimoto.
\newblock Proving test set contamination in black-box language models.
\newblock In \emph{The Twelfth International Conference on Learning Representations}, 2024.

\bibitem[Orgad et~al.(2023)Orgad, Kawar, and Belinkov]{orgad2023editing}
Hadas Orgad, Bahjat Kawar, and Yonatan Belinkov.
\newblock Editing implicit assumptions in text-to-image diffusion models.
\newblock In \emph{Proceedings of the IEEE/CVF International Conference on Computer Vision}, pp.\  7053--7061, 2023.

\bibitem[Ouyang et~al.(2022)Ouyang, Wu, Jiang, Almeida, Wainwright, Mishkin, Zhang, Agarwal, Slama, Ray, et~al.]{ouyang2022training}
Long Ouyang, Jeffrey Wu, Xu~Jiang, Diogo Almeida, Carroll Wainwright, Pamela Mishkin, Chong Zhang, Sandhini Agarwal, Katarina Slama, Alex Ray, et~al.
\newblock Training language models to follow instructions with human feedback.
\newblock \emph{Advances in Neural Information Processing Systems}, 35:\penalty0 27730--27744, 2022.

\bibitem[Pagnoni et~al.(2021)Pagnoni, Balachandran, and Tsvetkov]{pagnoni2021understanding}
Artidoro Pagnoni, Vidhisha Balachandran, and Yulia Tsvetkov.
\newblock Understanding factuality in abstractive summarization with frank: A benchmark for factuality metrics.
\newblock In \emph{Proceedings of the 2021 Conference of the North American Chapter of the Association for Computational Linguistics: Human Language Technologies}, pp.\  4812--4829, 2021.

\bibitem[Pan et~al.(2023)Pan, Chan, Zou, Li, Basart, Woodside, Zhang, Emmons, and Hendrycks]{pan2023rewards}
Alexander Pan, Jun~Shern Chan, Andy Zou, Nathaniel Li, Steven Basart, Thomas Woodside, Hanlin Zhang, Scott Emmons, and Dan Hendrycks.
\newblock Do the rewards justify the means? measuring trade-offs between rewards and ethical behavior in the machiavelli benchmark.
\newblock In \emph{International Conference on Machine Learning}, pp.\  26837--26867. PMLR, 2023.

\bibitem[Papineni et~al.(2002)Papineni, Roukos, Ward, and Zhu]{papineni2002bleu}
Kishore Papineni, Salim Roukos, Todd Ward, and Wei-Jing Zhu.
\newblock Bleu: a method for automatic evaluation of machine translation.
\newblock In \emph{Proceedings of the 40th annual meeting on association for computational linguistics}, pp.\  311--318. Association for Computational Linguistics, 2002.

\bibitem[Park et~al.(2024{\natexlab{a}})Park, Choe, and Veitch]{parklinear}
Kiho Park, Yo~Joong Choe, and Victor Veitch.
\newblock The linear representation hypothesis and the geometry of large language models.
\newblock In \emph{Forty-first International Conference on Machine Learning}, 2024{\natexlab{a}}.

\bibitem[Park et~al.(2024{\natexlab{b}})Park, Rafailov, Ermon, and Finn]{park2024disentangling}
Ryan Park, Rafael Rafailov, Stefano Ermon, and Chelsea Finn.
\newblock Disentangling length from quality in direct preference optimization.
\newblock In \emph{Findings of the Association for Computational Linguistics ACL 2024}, pp.\  4998--5017, 2024{\natexlab{b}}.

\bibitem[Pasunuru et~al.(2020)Pasunuru, Guo, and Bansal]{pasunuru2020dorb}
Ramakanth Pasunuru, Han Guo, and Mohit Bansal.
\newblock Dorb: Dynamically optimizing multiple rewards with bandits.
\newblock In \emph{Proceedings of the 2020 Conference on Empirical Methods in Natural Language Processing (EMNLP)}, pp.\  7766--7780, 2020.

\bibitem[Paulus et~al.(2018)Paulus, Xiong, and Socher]{paulus2018deep}
Romain Paulus, Caiming Xiong, and Richard Socher.
\newblock A deep reinforced model for abstractive summarization.
\newblock In \emph{International Conference on Learning Representations}, 2018.

\bibitem[Pei et~al.(2023)Pei, Yang, and Klein]{pei2023preadd}
Jonathan Pei, Kevin Yang, and Dan Klein.
\newblock Preadd: Prefix-adaptive decoding for controlled text generation.
\newblock In \emph{The 61st Annual Meeting Of The Association For Computational Linguistics}, 2023.

\bibitem[Peng et~al.(2018)Peng, Ghazvininejad, May, and Knight]{peng2018towards}
Nanyun Peng, Marjan Ghazvininejad, Jonathan May, and Kevin Knight.
\newblock Towards controllable story generation.
\newblock In \emph{Proceedings of the First Workshop on Storytelling}, pp.\  43--49, 2018.

\bibitem[Peyrard et~al.(2017)Peyrard, Botschen, and Gurevych]{peyrard2017learning}
Maxime Peyrard, Teresa Botschen, and Iryna Gurevych.
\newblock Learning to score system summaries for better content selection evaluation.
\newblock In \emph{Proceedings of the Workshop on New Frontiers in Summarization}, pp.\  74--84, 2017.

\bibitem[Pham et~al.(2021)Pham, Bui, Mai, and Nguyen]{pham2021out}
Thang Pham, Trung Bui, Long Mai, and Anh Nguyen.
\newblock Out of order: How important is the sequential order of words in a sentence in natural language understanding tasks?
\newblock In \emph{Findings of the Association for Computational Linguistics: ACL-IJCNLP 2021}, pp.\  1145--1160, 2021.

\bibitem[Plaat et~al.(2024)Plaat, Wong, Verberne, Broekens, van Stein, and Back]{plaat2024reasoning}
Aske Plaat, Annie Wong, Suzan Verberne, Joost Broekens, Niki van Stein, and Thomas Back.
\newblock Reasoning with large language models, a survey.
\newblock \emph{arXiv preprint arXiv:2407.11511}, 2024.

\bibitem[Popovi{\'c}(2019)]{popovic2019reducing}
Maja Popovi{\'c}.
\newblock On reducing translation shifts in translations intended for mt evaluation.
\newblock In \emph{Proceedings of Machine Translation Summit XVII: Translator, Project and User Tracks}, pp.\  80--87, 2019.

\bibitem[Post \& Vilar(2018)Post and Vilar]{post2018fast}
Matt Post and David Vilar.
\newblock Fast lexically constrained decoding with dynamic beam allocation for neural machine translation.
\newblock In \emph{Proceedings of the 2018 Conference of the North American Chapter of the Association for Computational Linguistics: Human Language Technologies, Volume 1 (Long Papers)}, pp.\  1314--1324, 2018.

\bibitem[Qin \& Eisner(2021)Qin and Eisner]{qin2021learning}
Guanghui Qin and Jason Eisner.
\newblock Learning how to ask: Querying lms with mixtures of soft prompts.
\newblock In \emph{Proceedings of the 2021 Conference of the North American Chapter of the Association for Computational Linguistics: Human Language Technologies (NAACL-HLT)}, 2021.

\bibitem[Qin et~al.(2019)Qin, Bosselut, Holtzman, Bhagavatula, Clark, and Choi]{qin2019counterfactual}
Lianhui Qin, Antoine Bosselut, Ari Holtzman, Chandra Bhagavatula, Elizabeth Clark, and Yejin Choi.
\newblock Counterfactual story reasoning and generation.
\newblock In \emph{Proceedings of the 2019 Conference on Empirical Methods in Natural Language Processing and the 9th International Joint Conference on Natural Language Processing (EMNLP-IJCNLP)}, pp.\  5046--5056, 2019.

\bibitem[Qin et~al.(2022)Qin, Welleck, Khashabi, and Choi]{qin2022cold}
Lianhui Qin, Sean Welleck, Daniel Khashabi, and Yejin Choi.
\newblock Cold decoding: Energy-based constrained text generation with langevin dynamics.
\newblock \emph{Advances in Neural Information Processing Systems}, 35:\penalty0 9538--9551, 2022.

\bibitem[Radford et~al.(2018)Radford, Narasimhan, Salimans, and Sutskever]{radford2018improving}
Alec Radford, Karthik Narasimhan, Tim Salimans, and Ilya Sutskever.
\newblock Improving language understanding by generative pre-training.
\newblock 2018.

\bibitem[Radford et~al.(2019)Radford, Wu, Child, Luan, Amodei, and Sutskever]{radford2019language}
Alec Radford, Jeffrey Wu, Rewon Child, David Luan, Dario Amodei, and Ilya Sutskever.
\newblock Language models are unsupervised multitask learners.
\newblock \emph{OpenAI Blog}, 1:\penalty0 8, 2019.

\bibitem[Rafailov et~al.(2024)Rafailov, Sharma, Mitchell, Manning, Ermon, and Finn]{rafailov2024direct}
Rafael Rafailov, Archit Sharma, Eric Mitchell, Christopher~D Manning, Stefano Ermon, and Chelsea Finn.
\newblock Direct preference optimization: Your language model is secretly a reward model.
\newblock \emph{Advances in Neural Information Processing Systems}, 36, 2024.

\bibitem[Rame et~al.(2024{\natexlab{a}})Rame, Couairon, Dancette, Gaya, Shukor, Soulier, and Cord]{rame2024rewarded}
Alexandre Rame, Guillaume Couairon, Corentin Dancette, Jean-Baptiste Gaya, Mustafa Shukor, Laure Soulier, and Matthieu Cord.
\newblock Rewarded soups: towards pareto-optimal alignment by interpolating weights fine-tuned on diverse rewards.
\newblock \emph{Advances in Neural Information Processing Systems}, 36, 2024{\natexlab{a}}.

\bibitem[Rame et~al.(2024{\natexlab{b}})Rame, Vieillard, Hussenot, Dadashi-Tazehozi, Cideron, Bachem, and Ferret]{rame2024warm}
Alexandre Rame, Nino Vieillard, Leonard Hussenot, Robert Dadashi-Tazehozi, Geoffrey Cideron, Olivier Bachem, and Johan Ferret.
\newblock Warm: On the benefits of weight averaged reward models.
\newblock In \emph{International Conference on Machine Learning}, pp.\  42048--42073. PMLR, 2024{\natexlab{b}}.

\bibitem[Ranzato et~al.(2016)Ranzato, Chopra, Auli, and Zaremba]{ranzato2015sequence}
Marc’Aurelio Ranzato, Sumit Chopra, Michael Auli, and Wojciech Zaremba.
\newblock Sequence level training with recurrent neural networks.
\newblock In \emph{4th International Conference on Learning Representations, ICLR}, 2016.

\bibitem[Rao \& Tetreault(2018)Rao and Tetreault]{rao2018dear}
Sudha Rao and Joel Tetreault.
\newblock Dear sir or madam, may i introduce the gyafc dataset: Corpus, benchmarks and metrics for formality style transfer.
\newblock In \emph{Proceedings of the 2018 Conference of the North American Chapter of the Association for Computational Linguistics: Human Language Technologies, Volume 1 (Long Papers)}, pp.\  129--140, 2018.

\bibitem[Rawat et~al.(2021)Rawat, Zhu, Li, Yu, Zaheer, Kumar, and Bhojanapalli]{rawat2021modifying}
Ankit~Singh Rawat, Chen Zhu, Daliang Li, Felix Yu, Manzil Zaheer, Sanjiv Kumar, and Srinadh Bhojanapalli.
\newblock Modifying memories in transformer models.
\newblock In \emph{International Conference on Machine Learning (ICML)}, volume 2020, 2021.

\bibitem[Razeghi et~al.(2022)Razeghi, Iv, Gardner, and Singh]{razeghi2022impact}
Yasaman Razeghi, Robert L~Logan Iv, Matt Gardner, and Sameer Singh.
\newblock Impact of pretraining term frequencies on few-shot numerical reasoning.
\newblock In \emph{Findings of the Association for Computational Linguistics: EMNLP 2022}, pp.\  840--854, 2022.

\bibitem[Rei et~al.(2020)Rei, Stewart, Farinha, and Lavie]{rei2020comet}
Ricardo Rei, Craig Stewart, Ana~C Farinha, and Alon Lavie.
\newblock Comet: A neural framework for mt evaluation.
\newblock In \emph{Proceedings of the 2020 Conference on Empirical Methods in Natural Language Processing (EMNLP)}, pp.\  2685--2702, 2020.

\bibitem[Reid \& Neubig(2022)Reid and Neubig]{reid2022learning}
Machel Reid and Graham Neubig.
\newblock Learning to model editing processes.
\newblock In \emph{Findings of the Association for Computational Linguistics: EMNLP 2022}, pp.\  3822--3832, 2022.

\bibitem[Reiter \& Belz(2009)Reiter and Belz]{reiter2009investigation}
Ehud Reiter and Anja Belz.
\newblock An investigation into the validity of some metrics for automatically evaluating natural language generation systems.
\newblock \emph{Computational Linguistics}, 35\penalty0 (4):\penalty0 529--558, 2009.

\bibitem[Reynolds \& McDonell(2021)Reynolds and McDonell]{reynolds2021prompt}
Laria Reynolds and Kyle McDonell.
\newblock Prompt programming for large language models: Beyond the few-shot paradigm.
\newblock In \emph{Extended Abstracts of the 2021 CHI Conference on Human Factors in Computing Systems}, pp.\  1--7, 2021.

\bibitem[Ribeiro et~al.(2020)Ribeiro, Wu, Guestrin, and Singh]{ribeiro2020beyond}
Marco~Tulio Ribeiro, Tongshuang Wu, Carlos Guestrin, and Sameer Singh.
\newblock Beyond accuracy: Behavioral testing of nlp models with checklist.
\newblock In \emph{Proceedings of the 58th Annual Meeting of the Association for Computational Linguistics}. Association for Computational Linguistics, 2020.

\bibitem[Rosati et~al.(2024)Rosati, Gonzales, Chen, Yu, Kayani, Rudzicz, and Sajjad]{rosati2024long}
Domenic Rosati, Robie Gonzales, Jinkun Chen, Xuemin Yu, Yahya Kayani, Frank Rudzicz, and Hassan Sajjad.
\newblock Long-form evaluation of model editing.
\newblock In \emph{Proceedings of the 2024 Conference of the North American Chapter of the Association for Computational Linguistics: Human Language Technologies (Volume 1: Long Papers)}, pp.\  3749--3780, 2024.

\bibitem[Ross et~al.(2022)Ross, Wu, Peng, Peters, and Gardner]{ross2022tailor}
Alexis Ross, Tongshuang Wu, Hao Peng, Matthew~E Peters, and Matt Gardner.
\newblock Tailor: Generating and perturbing text with semantic controls.
\newblock In \emph{60th Annual Meeting of the Association for Computational Linguistics, ACL 2022}, pp.\  3194--3213. Association for Computational Linguistics (ACL), 2022.

\bibitem[Ruan et~al.(2024)Ruan, Pu, Gao, Wan, and Zhu]{ruan2024better}
Jie Ruan, Xiao Pu, Mingqi Gao, Xiaojun Wan, and Yuesheng Zhu.
\newblock Better than random: Reliable nlg human evaluation with constrained active sampling.
\newblock In \emph{Proceedings of the AAAI Conference on Artificial Intelligence}, volume~38, pp.\  18915--18923, 2024.

\bibitem[Rush et~al.(2015)Rush, Chopra, and Weston]{rush2015neural}
Alexander~M Rush, Sumit Chopra, and Jason Weston.
\newblock A neural attention model for sentence summarization.
\newblock In \emph{Conference on Empirical Methods in Natural Language Processing, EMNLP 2015}, pp.\  379--389. Association for Computational Linguistics (ACL), 2015.

\bibitem[Sajjadi et~al.(2018)Sajjadi, Bachem, Lucic, Bousquet, and Gelly]{sajjadi2018assessing}
Mehdi~SM Sajjadi, Olivier Bachem, Mario Lucic, Olivier Bousquet, and Sylvain Gelly.
\newblock Assessing generative models via precision and recall.
\newblock In \emph{Advances in Neural Information Processing Systems}, pp.\  5228--5237, 2018.

\bibitem[Saleh et~al.(2020)Saleh, Jaques, Ghandeharioun, Shen, and Picard]{saleh2019hierarchical}
Abdelrhman Saleh, Natasha Jaques, Asma Ghandeharioun, Judy Shen, and Rosalind Picard.
\newblock Hierarchical reinforcement learning for open-domain dialog.
\newblock In \emph{Proceedings of the AAAI conference on artificial intelligence}, volume~34, pp.\  8741--8748, 2020.

\bibitem[Santurkar et~al.(2023)Santurkar, Durmus, Ladhak, Lee, Liang, and Hashimoto]{santurkar2023whose}
Shibani Santurkar, Esin Durmus, Faisal Ladhak, Cinoo Lee, Percy Liang, and Tatsunori Hashimoto.
\newblock Whose opinions do language models reflect?
\newblock In \emph{International Conference on Machine Learning}, pp.\  29971--30004. PMLR, 2023.

\bibitem[Schaeffer et~al.(2024)Schaeffer, Miranda, and Koyejo]{schaeffer2024emergent}
Rylan Schaeffer, Brando Miranda, and Sanmi Koyejo.
\newblock Are emergent abilities of large language models a mirage?
\newblock \emph{Advances in Neural Information Processing Systems}, 36, 2024.

\bibitem[Schoch et~al.(2020)Schoch, Yang, and Ji]{schoch2020problem}
Stephanie Schoch, Diyi Yang, and Yangfeng Ji.
\newblock “this is a problem, don’t you agree?” framing and bias in human evaluation for natural language generation.
\newblock In \emph{Proceedings of the 1st Workshop on Evaluating NLG Evaluation}, pp.\  10--16, 2020.

\bibitem[Schubotz et~al.(2018)Schubotz, Scharpf, Dudhat, Nagar, Hamborg, and Gipp]{schubotz2018introducing}
Moritz Schubotz, Philipp Scharpf, Kaushal Dudhat, Yash Nagar, Felix Hamborg, and Bela Gipp.
\newblock Introducing mathqa: a math-aware question answering system.
\newblock \emph{Information Discovery and Delivery}, 2018.

\bibitem[See et~al.(2017)See, Liu, and Manning]{see2017get}
Abigail See, Peter~J Liu, and Christopher~D Manning.
\newblock Get to the point: Summarization with pointer-generator networks.
\newblock In \emph{Proceedings of the 55th Annual Meeting of the Association for Computational Linguistics (Volume 1: Long Papers)}, pp.\  1073--1083, 2017.

\bibitem[See et~al.(2019)See, Roller, Kiela, and Weston]{see2019makes}
Abigail See, Stephen Roller, Douwe Kiela, and Jason Weston.
\newblock What makes a good conversation? how controllable attributes affect human judgments.
\newblock In \emph{Proceedings of NAACL-HLT}, pp.\  1702--1723, 2019.

\bibitem[Sellam et~al.(2020)Sellam, Das, and Parikh]{sellam2020bleurt}
Thibault Sellam, Dipanjan Das, and Ankur Parikh.
\newblock Bleurt: Learning robust metrics for text generation.
\newblock In \emph{Proceedings of the 58th Annual Meeting of the Association for Computational Linguistics}, pp.\  7881--7892, 2020.

\bibitem[Sennrich et~al.(2016)Sennrich, Haddow, and Birch]{sennrich2016controlling}
Rico Sennrich, Barry Haddow, and Alexandra Birch.
\newblock Controlling politeness in neural machine translation via side constraints.
\newblock In \emph{Proceedings of the 2016 Conference of the North American Chapter of the Association for Computational Linguistics: Human Language Technologies}, pp.\  35--40, 2016.

\bibitem[Serban et~al.(2016)Serban, Sordoni, Bengio, Courville, and Pineau]{serban2016building}
Iulian~V Serban, Alessandro Sordoni, Yoshua Bengio, Aaron Courville, and Joelle Pineau.
\newblock Building end-to-end dialogue systems using generative hierarchical neural network models.
\newblock In \emph{Thirtieth AAAI Conference on Artificial Intelligence}, 2016.

\bibitem[Serban et~al.(2015)Serban, Lowe, Henderson, Charlin, and Pineau]{serban2015survey}
Iulian~Vlad Serban, Ryan Lowe, Peter Henderson, Laurent Charlin, and Joelle Pineau.
\newblock A survey of available corpora for building data-driven dialogue systems.
\newblock \emph{arXiv preprint arXiv:1512.05742}, 2015.

\bibitem[Serban et~al.(2017)Serban, Sordoni, Lowe, Charlin, Pineau, Courville, and Bengio]{serban2017hierarchical}
Iulian~Vlad Serban, Alessandro Sordoni, Ryan Lowe, Laurent Charlin, Joelle Pineau, Aaron Courville, and Yoshua Bengio.
\newblock A hierarchical latent variable encoder-decoder model for generating dialogues.
\newblock In \emph{Thirty-First AAAI Conference on Artificial Intelligence}, 2017.

\bibitem[Sha(2020)]{sha2020gradient}
Lei Sha.
\newblock Gradient-guided unsupervised lexically constrained text generation.
\newblock In \emph{Proceedings of the 2020 Conference on Empirical Methods in Natural Language Processing (EMNLP)}, pp.\  8692--8703, 2020.

\bibitem[Shah et~al.(2020)Shah, Schuster, and Barzilay]{shah2020automatic}
Darsh~J Shah, Tal Schuster, and Regina Barzilay.
\newblock Automatic fact-guided sentence modification.
\newblock In \emph{AAAI}, pp.\  8791--8798, 2020.

\bibitem[Shayegani et~al.(2023)Shayegani, Mamun, Fu, Zaree, Dong, and Abu-Ghazaleh]{shayegani2023survey}
Erfan Shayegani, Md~Abdullah~Al Mamun, Yu~Fu, Pedram Zaree, Yue Dong, and Nael Abu-Ghazaleh.
\newblock Survey of vulnerabilities in large language models revealed by adversarial attacks.
\newblock \emph{arXiv preprint arXiv:2310.10844}, 2023.

\bibitem[Shen et~al.(2023)Shen, Jin, Huang, Liu, Dong, Guo, Wu, Liu, and Xiong]{shen2023large}
Tianhao Shen, Renren Jin, Yufei Huang, Chuang Liu, Weilong Dong, Zishan Guo, Xinwei Wu, Yan Liu, and Deyi Xiong.
\newblock Large language model alignment: A survey.
\newblock \emph{arXiv preprint arXiv:2309.15025}, 2023.

\bibitem[Shen et~al.(2017)Shen, Lei, Barzilay, and Jaakkola]{shen2017style}
Tianxiao Shen, Tao Lei, Regina Barzilay, and Tommi Jaakkola.
\newblock Style transfer from non-parallel text by cross-alignment.
\newblock In \emph{Advances in neural information processing systems}, pp.\  6830--6841, 2017.

\bibitem[Sheng et~al.(2019)Sheng, Chang, Natarajan, and Peng]{sheng2019woman}
Emily Sheng, Kai-Wei Chang, Prem Natarajan, and Nanyun Peng.
\newblock The woman worked as a babysitter: On biases in language generation.
\newblock In \emph{Proceedings of the 2019 Conference on Empirical Methods in Natural Language Processing and the 9th International Joint Conference on Natural Language Processing (EMNLP-IJCNLP)}, pp.\  3407--3412, 2019.

\bibitem[Sheng et~al.(2020)Sheng, Chang, Natarajan, and Peng]{sheng2020towards}
Emily Sheng, Kai-Wei Chang, Prem Natarajan, and Nanyun Peng.
\newblock Towards controllable biases in language generation.
\newblock In \emph{Proceedings of the 2020 Conference on Empirical Methods in Natural Language Processing: Findings}, pp.\  3239--3254, 2020.

\bibitem[Shi et~al.(2018)Shi, Chen, Qiu, and Huang]{shi2018toward}
Zhan Shi, Xinchi Chen, Xipeng Qiu, and Xuanjing Huang.
\newblock Toward diverse text generation with inverse reinforcement learning.
\newblock In \emph{Proceedings of the 27th International Joint Conference on Artificial Intelligence}, pp.\  4361--4367. AAAI Press, 2018.

\bibitem[Shimanaka et~al.(2018)Shimanaka, Kajiwara, and Komachi]{shimanaka2018ruse}
Hiroki Shimanaka, Tomoyuki Kajiwara, and Mamoru Komachi.
\newblock Ruse: Regressor using sentence embeddings for automatic machine translation evaluation.
\newblock In \emph{Proceedings of the Third Conference on Machine Translation: Shared Task Papers}, pp.\  751--758, 2018.

\bibitem[Shin et~al.(2022)Shin, Lee, Ahn, Kim, Kim, Kim, Cho, Lee, Park, Ha, et~al.]{shin2022effect}
Seongjin Shin, Sang-Woo Lee, Hwijeen Ahn, Sungdong Kim, HyoungSeok Kim, Boseop Kim, Kyunghyun Cho, Gichang Lee, Woomyoung Park, Jung-Woo Ha, et~al.
\newblock On the effect of pretraining corpora on in-context learning by a large-scale language model.
\newblock In \emph{Proceedings of the 2022 Conference of the North American Chapter of the Association for Computational Linguistics: Human Language Technologies}, pp.\  5168--5186, 2022.

\bibitem[Shin et~al.(2020)Shin, Razeghi, Logan~IV, Wallace, and Singh]{shin2020autoprompt}
Taylor Shin, Yasaman Razeghi, Robert~L Logan~IV, Eric Wallace, and Sameer Singh.
\newblock Autoprompt: Eliciting knowledge from language models with automatically generated prompts.
\newblock In \emph{Proceedings of the 2020 Conference on Empirical Methods in Natural Language Processing (EMNLP)}, pp.\  4222--4235, 2020.

\bibitem[Si et~al.(2023)Si, Gan, Yang, Wang, Wang, Boyd-Graber, and Wang]{si2022prompting}
Chenglei Si, Zhe Gan, Zhengyuan Yang, Shuohang Wang, Jianfeng Wang, Jordan~Lee Boyd-Graber, and Lijuan Wang.
\newblock Prompting gpt-3 to be reliable.
\newblock In \emph{The Eleventh International Conference on Learning Representations}, 2023.

\bibitem[Singhal et~al.(2024)Singhal, Goyal, Xu, and Durrett]{singhal2023long}
Prasann Singhal, Tanya Goyal, Jiacheng Xu, and Greg Durrett.
\newblock A long way to go: Investigating length correlations in rlhf.
\newblock In \emph{First Conference on Language Modeling}, 2024.

\bibitem[Sinha et~al.(2020)Sinha, Parthasarathi, Wang, Lowe, Hamilton, and Pineau]{sinha2020learning}
Koustuv Sinha, Prasanna Parthasarathi, Jasmine Wang, Ryan Lowe, William~L Hamilton, and Joelle Pineau.
\newblock Learning an unreferenced metric for online dialogue evaluation.
\newblock In \emph{Proceedings of the 58th Annual Meeting of the Association for Computational Linguistics}, pp.\  2430--2441, 2020.

\bibitem[Sorensen et~al.(2024)Sorensen, Moore, Fisher, Gordon, Mireshghallah, Rytting, Ye, Jiang, Lu, Dziri, et~al.]{sorensenposition}
Taylor Sorensen, Jared Moore, Jillian Fisher, Mitchell~L Gordon, Niloofar Mireshghallah, Christopher~Michael Rytting, Andre Ye, Liwei Jiang, Ximing Lu, Nouha Dziri, et~al.
\newblock Position: A roadmap to pluralistic alignment.
\newblock In \emph{Forty-first International Conference on Machine Learning}, 2024.

\bibitem[Sprague et~al.(2024)Sprague, Ye, Bostrom, Chaudhuri, and Durrett]{spraguemusr}
Zayne~Rea Sprague, Xi~Ye, Kaj Bostrom, Swarat Chaudhuri, and Greg Durrett.
\newblock Musr: Testing the limits of chain-of-thought with multistep soft reasoning.
\newblock In \emph{The Twelfth International Conference on Learning Representations}, 2024.

\bibitem[Stent et~al.(2005)Stent, Marge, and Singhai]{stent2005evaluating}
Amanda Stent, Matthew Marge, and Mohit Singhai.
\newblock Evaluating evaluation methods for generation in the presence of variation.
\newblock In \emph{international conference on intelligent text processing and computational linguistics}, pp.\  341--351. Springer, 2005.

\bibitem[Stern et~al.(2019)Stern, Chan, Kiros, and Uszkoreit]{stern2019insertion}
Mitchell Stern, William Chan, Jamie Kiros, and Jakob Uszkoreit.
\newblock Insertion transformer: Flexible sequence generation via insertion operations.
\newblock In \emph{International Conference on Machine Learning}, pp.\  5976--5985, 2019.

\bibitem[Stiennon et~al.(2020)Stiennon, Ouyang, Wu, Ziegler, Lowe, Voss, Radford, Amodei, and Christiano]{stiennon2020learning}
Nisan Stiennon, Long Ouyang, Jeffrey Wu, Daniel Ziegler, Ryan Lowe, Chelsea Voss, Alec Radford, Dario Amodei, and Paul~F Christiano.
\newblock Learning to summarize with human feedback.
\newblock \emph{Advances in Neural Information Processing Systems}, 33:\penalty0 3008--3021, 2020.

\bibitem[Subramani et~al.(2022)Subramani, Suresh, and Peters]{subramani2022extracting}
Nishant Subramani, Nivedita Suresh, and Matthew~E Peters.
\newblock Extracting latent steering vectors from pretrained language models.
\newblock In \emph{Findings of the Association for Computational Linguistics: ACL 2022}, pp.\  566--581, 2022.

\bibitem[Sudhakar et~al.(2019)Sudhakar, Upadhyay, and Maheswaran]{sudhakar2019transforming}
Akhilesh Sudhakar, Bhargav Upadhyay, and Arjun Maheswaran.
\newblock “transforming” delete, retrieve, generate approach for controlled text style transfer.
\newblock In \emph{Proceedings of the 2019 Conference on Empirical Methods in Natural Language Processing and the 9th International Joint Conference on Natural Language Processing (EMNLP-IJCNLP)}, pp.\  3269--3279, 2019.

\bibitem[Sulem et~al.(2018)Sulem, Abend, and Rappoport]{sulem2018bleu}
Elior Sulem, Omri Abend, and Ari Rappoport.
\newblock Bleu is not suitable for the evaluation of text simplification.
\newblock In \emph{Proceedings of the 2018 Conference on Empirical Methods in Natural Language Processing}, pp.\  738--744, 2018.

\bibitem[Sun et~al.(2023{\natexlab{a}})Sun, Tian, Zhou, Xu, Hu, Gupta, Wieting, Peng, and Ma]{sun2023evaluating}
Jiao Sun, Yufei Tian, Wangchunshu Zhou, Nan Xu, Qian Hu, Rahul Gupta, John Wieting, Nanyun Peng, and Xuezhe Ma.
\newblock Evaluating large language models on controlled generation tasks.
\newblock In \emph{Proceedings of the 2023 Conference on Empirical Methods in Natural Language Processing}, pp.\  3155--3168, 2023{\natexlab{a}}.

\bibitem[Sun et~al.(2023{\natexlab{b}})Sun, Gupta, and Iyyer]{sun2023exploring}
Simeng Sun, Dhawal Gupta, and Mohit Iyyer.
\newblock Exploring the impact of low-rank adaptation on the performance, efficiency, and regularization of rlhf.
\newblock \emph{arXiv preprint arXiv:2309.09055}, 2023{\natexlab{b}}.

\bibitem[Talmor et~al.(2019)Talmor, Herzig, Lourie, and Berant]{talmor2019commonsenseqa}
Alon Talmor, Jonathan Herzig, Nicholas Lourie, and Jonathan Berant.
\newblock Commonsenseqa: A question answering challenge targeting commonsense knowledge.
\newblock In \emph{Proceedings of the 2019 Conference of the North American Chapter of the Association for Computational Linguistics: Human Language Technologies, Volume 1 (Long and Short Papers)}, pp.\  4149--4158, 2019.

\bibitem[Tam et~al.(2022)Tam, Mascarenhas, Zhang, Kwan, Bansal, and Raffel]{tam2022evaluating}
Derek Tam, Anisha Mascarenhas, Shiyue Zhang, Sarah Kwan, Mohit Bansal, and Colin Raffel.
\newblock Evaluating the factual consistency of large language models through summarization.
\newblock \emph{arXiv preprint arXiv:2211.08412}, 2022.

\bibitem[Tang et~al.(2019)Tang, Zhao, Xiong, Liang, Xing, and Hu]{tang2019target}
Jianheng Tang, Tiancheng Zhao, Chenyan Xiong, Xiaodan Liang, Eric Xing, and Zhiting Hu.
\newblock Target-guided open-domain conversation.
\newblock In \emph{Proceedings of the 57th Annual Meeting of the Association for Computational Linguistics}, pp.\  5624--5634, 2019.

\bibitem[Tang et~al.(2023)Tang, Wang, Zhou, Li, Cao, and Zhang]{tang2023can}
Zecheng Tang, Pinzheng Wang, Keyan Zhou, Juntao Li, Ziqiang Cao, and Min Zhang.
\newblock Can diffusion model achieve better performance in text generation? bridging the gap between training and inference!
\newblock In \emph{Findings of the Association for Computational Linguistics: ACL 2023}, pp.\  11359--11386, 2023.

\bibitem[Theis et~al.(2016)Theis, van~den Oord, and Bethge]{theis2016note}
L~Theis, A~van~den Oord, and M~Bethge.
\newblock A note on the evaluation of generative models.
\newblock In \emph{International Conference on Learning Representations (ICLR 2016)}, pp.\  1--10, 2016.

\bibitem[Tigges et~al.(2023)Tigges, Hollinsworth, Geiger, and Nanda]{tigges2023linear}
Curt Tigges, Oskar~John Hollinsworth, Atticus Geiger, and Neel Nanda.
\newblock Linear representations of sentiment in large language models.
\newblock \emph{arXiv preprint arXiv:2310.15154}, 2023.

\bibitem[Touvron et~al.(2023)Touvron, Martin, Stone, Albert, Almahairi, Babaei, Bashlykov, Batra, Bhargava, Bhosale, et~al.]{touvron2023llama}
Hugo Touvron, Louis Martin, Kevin Stone, Peter Albert, Amjad Almahairi, Yasmine Babaei, Nikolay Bashlykov, Soumya Batra, Prajjwal Bhargava, Shruti Bhosale, et~al.
\newblock Llama 2: Open foundation and fine-tuned chat models.
\newblock \emph{arXiv preprint arXiv:2307.09288}, 2023.

\bibitem[Turner et~al.(2023)Turner, Monte, Udell, Thiergart, and Mini]{turner2023steering}
Alex Turner, M~Monte, David Udell, Lisa Thiergart, and Ulisse Mini.
\newblock Steering gpt-2-xl by adding an activation vector.
\newblock In \emph{AI Alignment Forum}, 2023.

\bibitem[van~der Lee et~al.(2019)van~der Lee, Gatt, van Miltenburg, Wubben, and Krahmer]{van2019best}
Chris van~der Lee, Albert Gatt, Emiel van Miltenburg, Sander Wubben, and Emiel Krahmer.
\newblock Best practices for the human evaluation of automatically generated text.
\newblock In \emph{Proceedings of the 12th International Conference on Natural Language Generation}, pp.\  355--368, 2019.

\bibitem[van Stegeren \& Theune(2019)van Stegeren and Theune]{van2019narrative}
Judith van Stegeren and Mari{\"e}t Theune.
\newblock Narrative generation in the wild: Methods from nanogenmo.
\newblock In \emph{Proceedings of the Second Workshop on Storytelling}, pp.\  65--74, 2019.

\bibitem[Vaswani et~al.(2017)Vaswani, Shazeer, Parmar, Uszkoreit, Jones, Gomez, Kaiser, and Polosukhin]{vaswani2017attention}
Ashish Vaswani, Noam Shazeer, Niki Parmar, Jakob Uszkoreit, Llion Jones, Aidan~N Gomez, {\L}ukasz Kaiser, and Illia Polosukhin.
\newblock Attention is all you need.
\newblock \emph{Advances in neural information processing systems}, 30, 2017.

\bibitem[Vedantam et~al.(2015)Vedantam, Lawrence~Zitnick, and Parikh]{vedantam2015cider}
Ramakrishna Vedantam, C~Lawrence~Zitnick, and Devi Parikh.
\newblock Cider: Consensus-based image description evaluation.
\newblock In \emph{Proceedings of the IEEE conference on computer vision and pattern recognition}, pp.\  4566--4575, 2015.

\bibitem[Vijayakumar et~al.(2016)Vijayakumar, Cogswell, Selvaraju, Sun, Lee, Crandall, and Batra]{vijayakumar2016diverse}
Ashwin~K Vijayakumar, Michael Cogswell, Ramprasath~R Selvaraju, Qing Sun, Stefan Lee, David Crandall, and Dhruv Batra.
\newblock Diverse beam search: Decoding diverse solutions from neural sequence models.
\newblock \emph{arXiv preprint arXiv:1610.02424}, 2016.

\bibitem[Wallace et~al.(2019)Wallace, Feng, Kandpal, Gardner, and Singh]{wallace2019universal}
Eric Wallace, Shi Feng, Nikhil Kandpal, Matt Gardner, and Sameer Singh.
\newblock Universal adversarial triggers for attacking and analyzing nlp.
\newblock In \emph{Proceedings of the 2019 Conference on Empirical Methods in Natural Language Processing and the 9th International Joint Conference on Natural Language Processing (EMNLP-IJCNLP)}, 2019.

\bibitem[Wang et~al.(2017)Wang, Jojic, Brockett, and Nyberg]{wang2017steering}
Di~Wang, Nebojsa Jojic, Chris Brockett, and Eric Nyberg.
\newblock Steering output style and topic in neural response generation.
\newblock In \emph{Proceedings of the 2017 Conference on Empirical Methods in Natural Language Processing}, pp.\  2140--2150, 2017.

\bibitem[Wang et~al.(2024{\natexlab{a}})Wang, Lin, Xiong, Yang, Diao, Qiu, Zhao, and Zhang]{wang2024arithmetic}
Haoxiang Wang, Yong Lin, Wei Xiong, Rui Yang, Shizhe Diao, Shuang Qiu, Han Zhao, and Tong Zhang.
\newblock Arithmetic control of llms for diverse user preferences: Directional preference alignment with multi-objective rewards.
\newblock In \emph{62nd Annual Meeting of the Association for Computational Linguistics, ACL 2024}, pp.\  8642--8655. Association for Computational Linguistics (ACL), 2024{\natexlab{a}}.

\bibitem[Wang et~al.(2023{\natexlab{a}})Wang, Liang, Meng, Sun, Shi, Li, Xu, Qu, and Zhou]{wang2023chatgpt}
Jiaan Wang, Yunlong Liang, Fandong Meng, Zengkui Sun, Haoxiang Shi, Zhixu Li, Jinan Xu, Jianfeng Qu, and Jie Zhou.
\newblock Is chatgpt a good nlg evaluator? a preliminary study.
\newblock In \emph{Proceedings of the 4th New Frontiers in Summarization Workshop}. Association for Computational Linguistics, 2023{\natexlab{a}}.

\bibitem[Wang et~al.(2019)Wang, Gan, Xu, Zhang, Wang, Shen, Chen, and Carin]{wang2019topic}
Wenlin Wang, Zhe Gan, Hongteng Xu, Ruiyi Zhang, Guoyin Wang, Dinghan Shen, Changyou Chen, and Lawrence Carin.
\newblock Topic-guided variational auto-encoder for text generation.
\newblock In \emph{Proceedings of the 2019 Conference of the North American Chapter of the Association for Computational Linguistics: Human Language Technologies, Volume 1 (Long and Short Papers)}, pp.\  166--177, 2019.

\bibitem[Wang et~al.(2023{\natexlab{b}})Wang, Ivison, Dasigi, Hessel, Khot, Chandu, Wadden, MacMillan, Smith, Beltagy, et~al.]{wang2023far}
Yizhong Wang, Hamish Ivison, Pradeep Dasigi, Jack Hessel, Tushar Khot, Khyathi Chandu, David Wadden, Kelsey MacMillan, Noah~A Smith, Iz~Beltagy, et~al.
\newblock How far can camels go? exploring the state of instruction tuning on open resources.
\newblock \emph{Advances in Neural Information Processing Systems}, 36:\penalty0 74764--74786, 2023{\natexlab{b}}.

\bibitem[Wang et~al.(2024{\natexlab{b}})Wang, Bi, Pentyala, Ramnath, Chaudhuri, Mehrotra, Mao, Asur, et~al.]{wang2024comprehensive}
Zhichao Wang, Bin Bi, Shiva~Kumar Pentyala, Kiran Ramnath, Sougata Chaudhuri, Shubham Mehrotra, Xiang-Bo Mao, Sitaram Asur, et~al.
\newblock A comprehensive survey of llm alignment techniques: Rlhf, rlaif, ppo, dpo and more.
\newblock \emph{arXiv preprint arXiv:2407.16216}, 2024{\natexlab{b}}.

\bibitem[Wang et~al.(2024{\natexlab{c}})Wang, Gui, Negrea, and Veitch]{wang2024concept}
Zihao Wang, Lin Gui, Jeffrey Negrea, and Victor Veitch.
\newblock Concept algebra for (score-based) text-controlled generative models.
\newblock \emph{Advances in Neural Information Processing Systems}, 36, 2024{\natexlab{c}}.

\bibitem[Wei et~al.(2022)Wei, Wang, Schuurmans, Bosma, Xia, Chi, Le, Zhou, et~al.]{wei2022chain}
Jason Wei, Xuezhi Wang, Dale Schuurmans, Maarten Bosma, Fei Xia, Ed~Chi, Quoc~V Le, Denny Zhou, et~al.
\newblock Chain-of-thought prompting elicits reasoning in large language models.
\newblock \emph{Advances in neural information processing systems}, 35:\penalty0 24824--24837, 2022.

\bibitem[Wei et~al.(2024)Wei, Miin, and Miin]{wei2024optimizing}
Timothy Wei, Annabelle Miin, and Anastasia Miin.
\newblock Optimizing large language models for dynamic constraints through human-in-the-loop discriminators.
\newblock \emph{arXiv preprint arXiv:2410.15163}, 2024.

\bibitem[Welleck et~al.(2019)Welleck, Brantley, Iii, and Cho]{welleck2019non}
Sean Welleck, Kiant{\'e} Brantley, Hal~Daum{\'e} Iii, and Kyunghyun Cho.
\newblock Non-monotonic sequential text generation.
\newblock In \emph{International Conference on Machine Learning}, pp.\  6716--6726. PMLR, 2019.

\bibitem[Welleck et~al.(2023)Welleck, Lu, West, Brahman, Shen, Khashabi, and Choi]{welleck2023generating}
Sean Welleck, Ximing Lu, Peter West, Faeze Brahman, Tianxiao Shen, Daniel Khashabi, and Yejin Choi.
\newblock Generating sequences by learning to self-correct.
\newblock In \emph{The Eleventh International Conference on Learning Representations}, 2023.

\bibitem[Wiese et~al.(2017)Wiese, Weissenborn, and Neves]{wiese2017neural}
Georg Wiese, Dirk Weissenborn, and Mariana Neves.
\newblock Neural domain adaptation for biomedical question answering.
\newblock In \emph{Proceedings of the 21st Conference on Computational Natural Language Learning (CoNLL 2017)}, pp.\  281--289, 2017.

\bibitem[Wieting et~al.(2019)Wieting, Berg-Kirkpatrick, Gimpel, and Neubig]{wieting2019beyond}
John Wieting, Taylor Berg-Kirkpatrick, Kevin Gimpel, and Graham Neubig.
\newblock Beyond bleu: Training neural machine translation with semantic similarity.
\newblock In \emph{Proceedings of the 57th Annual Meeting of the Association for Computational Linguistics}, pp.\  4344--4355, 2019.

\bibitem[Williams et~al.(2024)Williams, Carroll, Narang, Weisser, Murphy, and Dragan]{williams2024targeted}
Marcus Williams, Micah Carroll, Adhyyan Narang, Constantin Weisser, Brendan Murphy, and Anca Dragan.
\newblock On targeted manipulation and deception when optimizing llms for user feedback.
\newblock \emph{arXiv preprint arXiv:2411.02306}, 2024.

\bibitem[Wiseman et~al.(2017)Wiseman, Shieber, and Rush]{wiseman2017challenges}
Sam Wiseman, Stuart~M Shieber, and Alexander~M Rush.
\newblock Challenges in data-to-document generation.
\newblock In \emph{Proceedings of the 2017 Conference on Empirical Methods in Natural Language Processing}, pp.\  2253--2263, 2017.

\bibitem[Wu \& Hu(2018)Wu and Hu]{wu2018learning}
Yuxiang Wu and Baotian Hu.
\newblock Learning to extract coherent summary via deep reinforcement learning.
\newblock In \emph{Thirty-Second AAAI Conference on Artificial Intelligence}, 2018.

\bibitem[Wu et~al.(2024{\natexlab{a}})Wu, Hu, Shi, Dziri, Suhr, Ammanabrolu, Smith, Ostendorf, and Hajishirzi]{wu2024fine}
Zeqiu Wu, Yushi Hu, Weijia Shi, Nouha Dziri, Alane Suhr, Prithviraj Ammanabrolu, Noah~A Smith, Mari Ostendorf, and Hannaneh Hajishirzi.
\newblock Fine-grained human feedback gives better rewards for language model training.
\newblock \emph{Advances in Neural Information Processing Systems}, 36, 2024{\natexlab{a}}.

\bibitem[Wu et~al.(2024{\natexlab{b}})Wu, Qiu, Ross, Aky{\"u}rek, Chen, Wang, Kim, Andreas, and Kim]{wu2023reasoning}
Zhaofeng Wu, Linlu Qiu, Alexis Ross, Ekin Aky{\"u}rek, Boyuan Chen, Bailin Wang, Najoung Kim, Jacob Andreas, and Yoon Kim.
\newblock Reasoning or reciting? exploring the capabilities and limitations of language models through counterfactual tasks.
\newblock In \emph{Proceedings of the 2024 Conference of the North American Chapter of the Association for Computational Linguistics: Human Language Technologies (Volume 1: Long Papers)}, pp.\  1819--1862, 2024{\natexlab{b}}.

\bibitem[Wubben et~al.(2012)Wubben, Van Den~Bosch, and Krahmer]{wubben2012sentence}
Sander Wubben, Antal Van Den~Bosch, and Emiel Krahmer.
\newblock Sentence simplification by monolingual machine translation.
\newblock In \emph{Proceedings of the 50th Annual Meeting of the Association for Computational Linguistics: Long Papers-Volume 1}, pp.\  1015--1024. Association for Computational Linguistics, 2012.

\bibitem[Wuebker et~al.(2016)Wuebker, Green, DeNero, Hasan, and Luong]{wuebker2016models}
Joern Wuebker, Spence Green, John DeNero, Sa{\v{s}}a Hasan, and Minh-Thang Luong.
\newblock Models and inference for prefix-constrained machine translation.
\newblock In \emph{Proceedings of the 54th Annual Meeting of the Association for Computational Linguistics (Volume 1: Long Papers)}, pp.\  66--75, 2016.

\bibitem[Xing et~al.(2016)Xing, Wu, Wu, Liu, Huang, Zhou, and Ma]{xing2016topic}
Chen Xing, Wei Wu, Yu~Wu, Jie Liu, Yalou Huang, Ming Zhou, and Wei-Ying Ma.
\newblock Topic augmented neural response generation with a joint attention mechanism.
\newblock \emph{arXiv preprint arXiv:1606.08340}, 2\penalty0 (2), 2016.

\bibitem[Xu et~al.(2024)Xu, Wang, Fan, and Liu]{xu2024benchmarking}
Ruijie Xu, Zengzhi Wang, Run-Ze Fan, and Pengfei Liu.
\newblock Benchmarking benchmark leakage in large language models.
\newblock \emph{arXiv preprint arXiv:2404.18824}, 2024.

\bibitem[Yang \& Klein(2021)Yang and Klein]{yang2021fudge}
Kevin Yang and Dan Klein.
\newblock Fudge: Controlled text generation with future discriminators.
\newblock In \emph{Proceedings of the 2021 Conference of the North American Chapter of the Association for Computational Linguistics: Human Language Technologies}, pp.\  3511--3535, 2021.

\bibitem[Yang et~al.(2023)Yang, Zhang, Song, Hong, Xu, Zhao, Zhang, Cui, and Yang]{yang2022diffusion}
Ling Yang, Zhilong Zhang, Yang Song, Shenda Hong, Runsheng Xu, Yue Zhao, Wentao Zhang, Bin Cui, and Ming-Hsuan Yang.
\newblock Diffusion models: A comprehensive survey of methods and applications.
\newblock \emph{ACM Computing Surveys}, 56\penalty0 (4):\penalty0 1--39, 2023.

\bibitem[Yang et~al.(2024{\natexlab{a}})Yang, Ma, and Cheng]{yang2024plug}
Nai-Chi Yang, Wei-Yun Ma, and Pu-Jen Cheng.
\newblock Plug-in language model: Controlling text generation with a simple regression model.
\newblock In \emph{Findings of the Association for Computational Linguistics: NAACL 2024}, pp.\  2165--2181, 2024{\natexlab{a}}.

\bibitem[Yang et~al.(2024{\natexlab{b}})Yang, Pan, Luo, Qiu, Zhong, Yu, and Chen]{yang2024rewards}
Rui Yang, Xiaoman Pan, Feng Luo, Shuang Qiu, Han Zhong, Dong Yu, and Jianshu Chen.
\newblock Rewards-in-context: multi-objective alignment of foundation models with dynamic preference adjustment.
\newblock In \emph{Proceedings of the 41st International Conference on Machine Learning}, pp.\  56276--56297, 2024{\natexlab{b}}.

\bibitem[Yao et~al.(2024)Yao, Yu, Zhao, Shafran, Griffiths, Cao, and Narasimhan]{yao2024tree}
Shunyu Yao, Dian Yu, Jeffrey Zhao, Izhak Shafran, Tom Griffiths, Yuan Cao, and Karthik Narasimhan.
\newblock Tree of thoughts: Deliberate problem solving with large language models.
\newblock \emph{Advances in Neural Information Processing Systems}, 36, 2024.

\bibitem[Yuan et~al.(2021)Yuan, Neubig, and Liu]{yuan2021bartscore}
Weizhe Yuan, Graham Neubig, and Pengfei Liu.
\newblock Bartscore: Evaluating generated text as text generation.
\newblock \emph{Advances in Neural Information Processing Systems}, 34, 2021.

\bibitem[Zellers et~al.(2021)Zellers, Holtzman, Clark, Qin, Farhadi, and Choi]{zellers2021turingadvice}
Rowan Zellers, Ari Holtzman, Elizabeth Clark, Lianhui Qin, Ali Farhadi, and Yejin Choi.
\newblock Turingadvice: A generative and dynamic evaluation of language use.
\newblock In \emph{Proceedings of the 16th Conference of the European Chapter of the Association for Computational Linguistics: Main Volume}, 2021.

\bibitem[Zhang et~al.(2016)Zhang, Xiong, Su, Duan, and Zhang]{zhang2016variational}
Biao Zhang, Deyi Xiong, Jinsong Su, Hong Duan, and Min Zhang.
\newblock Variational neural machine translation.
\newblock In \emph{Proceedings of the 2016 Conference on Empirical Methods in Natural Language Processing}, pp.\  521--530, 2016.

\bibitem[Zhang et~al.(2024{\natexlab{a}})Zhang, Wang, and Dhillon]{zhang2024causal}
Bohan Zhang, Yixin Wang, and Paramveer~S Dhillon.
\newblock Causal inference for human-language model collaboration.
\newblock In \emph{Proceedings of the 2024 Conference of the North American Chapter of the Association for Computational Linguistics: Human Language Technologies (Volume 1: Long Papers)}, pp.\  1630--1647, 2024{\natexlab{a}}.

\bibitem[Zhang \& Song(2022)Zhang and Song]{zhang2022discup}
Hanqing Zhang and Dawei Song.
\newblock Discup: Discriminator cooperative unlikelihood prompt-tuning for controllable text generation.
\newblock In \emph{Proceedings of the 2022 Conference on Empirical Methods in Natural Language Processing}, pp.\  3392--3406, 2022.

\bibitem[Zhang et~al.(2023)Zhang, Song, Li, Zhou, and Song]{zhang2023survey}
Hanqing Zhang, Haolin Song, Shaoyu Li, Ming Zhou, and Dawei Song.
\newblock A survey of controllable text generation using transformer-based pre-trained language models.
\newblock \emph{ACM Computing Surveys}, 56\penalty0 (3):\penalty0 1--37, 2023.

\bibitem[Zhang et~al.(2018{\natexlab{a}})Zhang, Dinan, Urbanek, Szlam, Kiela, and Weston]{zhang2018personalizing}
Saizheng Zhang, Emily Dinan, Jack Urbanek, Arthur Szlam, Douwe Kiela, and Jason Weston.
\newblock Personalizing dialogue agents: I have a dog, do you have pets too?
\newblock In \emph{Proceedings of the 56th Annual Meeting of the Association for Computational Linguistics (Volume 1: Long Papers)}, pp.\  2204--2213, 2018{\natexlab{a}}.

\bibitem[Zhang et~al.(2020{\natexlab{a}})Zhang, Kishore, Wu, Weinberger, and Artzi]{zhang2019bertscore}
Tianyi Zhang, Varsha Kishore, Felix Wu, Kilian~Q Weinberger, and Yoav Artzi.
\newblock Bertscore: Evaluating text generation with bert.
\newblock In \emph{International Conference on Learning Representations}, 2020{\natexlab{a}}.

\bibitem[Zhang \& Lapata(2014)Zhang and Lapata]{zhang2014chinese}
Xingxing Zhang and Mirella Lapata.
\newblock Chinese poetry generation with recurrent neural networks.
\newblock In \emph{Proceedings of the 2014 Conference on Empirical Methods in Natural Language Processing (EMNLP)}, pp.\  670--680, 2014.

\bibitem[Zhang et~al.(2018{\natexlab{b}})Zhang, Galley, Gao, Gan, Li, Brockett, and Dolan]{zhang2018generating}
Yizhe Zhang, Michel Galley, Jianfeng Gao, Zhe Gan, Xiujun Li, Chris Brockett, and Bill Dolan.
\newblock Generating informative and diverse conversational responses via adversarial information maximization.
\newblock In \emph{Advances in Neural Information Processing Systems}, pp.\  1810--1820, 2018{\natexlab{b}}.

\bibitem[Zhang et~al.(2020{\natexlab{b}})Zhang, Wang, Li, Gan, Brockett, and Dolan]{zhang2020pointer}
Yizhe Zhang, Guoyin Wang, Chunyuan Li, Zhe Gan, Chris Brockett, and William~B Dolan.
\newblock Pointer: Constrained progressive text generation via insertion-based generative pre-training.
\newblock In \emph{Proceedings of the 2020 Conference on Empirical Methods in Natural Language Processing (EMNLP)}, pp.\  8649--8670, 2020{\natexlab{b}}.

\bibitem[Zhang et~al.(2024{\natexlab{b}})Zhang, Rossi, Kveton, Shao, Yang, Zamani, Dernoncourt, Barrow, Yu, Kim, et~al.]{zhang2024personalization}
Zhehao Zhang, Ryan~A Rossi, Branislav Kveton, Yijia Shao, Diyi Yang, Hamed Zamani, Franck Dernoncourt, Joe Barrow, Tong Yu, Sungchul Kim, et~al.
\newblock Personalization of large language models: A survey.
\newblock \emph{arXiv preprint arXiv:2411.00027}, 2024{\natexlab{b}}.

\bibitem[Zhang et~al.(2018{\natexlab{c}})Zhang, Ren, Liu, Wang, Chen, Li, Zhou, and Chen]{zhang2018style}
Zhirui Zhang, Shuo Ren, Shujie Liu, Jianyong Wang, Peng Chen, Mu~Li, Ming Zhou, and Enhong Chen.
\newblock Style transfer as unsupervised machine translation.
\newblock \emph{arXiv preprint arXiv:1808.07894}, 2018{\natexlab{c}}.

\bibitem[Zhao et~al.(2018)Zhao, Kim, Zhang, Rush, and LeCun]{zhao2018adversarially}
Jake Zhao, Yoon Kim, Kelly Zhang, Alexander~M Rush, and Yann LeCun.
\newblock Adversarially regularized autoencoders.
\newblock In \emph{35th International Conference on Machine Learning, ICML 2018}, pp.\  9405--9420. International Machine Learning Society (IMLS), 2018.

\bibitem[Zhao et~al.(2019)Zhao, Peyrard, Liu, Gao, Meyer, and Eger]{zhao2019moverscore}
Wei Zhao, Maxime Peyrard, Fei Liu, Yang Gao, Christian~M Meyer, and Steffen Eger.
\newblock Moverscore: Text generation evaluating with contextualized embeddings and earth mover distance.
\newblock In \emph{Proceedings of the 2019 Conference on Empirical Methods in Natural Language Processing and the 9th International Joint Conference on Natural Language Processing (EMNLP-IJCNLP)}, pp.\  563--578, 2019.

\bibitem[Zheng et~al.(2019)Zheng, Chen, Huang, Liu, and Zhu]{zheng2019personalized}
Yinhe Zheng, Guanyi Chen, Minlie Huang, Song Liu, and Xuan Zhu.
\newblock Personalized dialogue generation with diversified traits.
\newblock \emph{arXiv preprint arXiv:1901.09672}, 2019.

\bibitem[Zhong et~al.(2023)Zhong, Wu, Manning, Potts, and Chen]{zhong2023mquake}
Zexuan Zhong, Zhengxuan Wu, Christopher~D Manning, Christopher Potts, and Danqi Chen.
\newblock Mquake: Assessing knowledge editing in language models via multi-hop questions.
\newblock In \emph{Proceedings of the 2023 Conference on Empirical Methods in Natural Language Processing}, pp.\  15686--15702, 2023.

\bibitem[Zhou \& Neubig(2017)Zhou and Neubig]{zhou2017multi}
Chunting Zhou and Graham Neubig.
\newblock Multi-space variational encoder-decoders for semi-supervised labeled sequence transduction.
\newblock In \emph{Proceedings of the 55th Annual Meeting of the Association for Computational Linguistics (Volume 1: Long Papers)}, pp.\  310--320, 2017.

\bibitem[Zhou et~al.(2021)Zhou, Neubig, Gu, Diab, Guzm{\'a}n, Zettlemoyer, and Ghazvininejad]{zhou2021detecting}
Chunting Zhou, Graham Neubig, Jiatao Gu, Mona Diab, Francisco Guzm{\'a}n, Luke Zettlemoyer, and Marjan Ghazvininejad.
\newblock Detecting hallucinated content in conditional neural sequence generation.
\newblock In \emph{Findings of the Association for Computational Linguistics: ACL-IJCNLP 2021}, pp.\  1393--1404, 2021.

\bibitem[Zhou et~al.(2024)Zhou, Liu, Xu, Iyer, Sun, Mao, Ma, Efrat, Yu, Yu, et~al.]{zhou2024lima}
Chunting Zhou, Pengfei Liu, Puxin Xu, Srinivasan Iyer, Jiao Sun, Yuning Mao, Xuezhe Ma, Avia Efrat, Ping Yu, Lili Yu, et~al.
\newblock Lima: Less is more for alignment.
\newblock \emph{Advances in Neural Information Processing Systems}, 36, 2024.

\bibitem[Zhou et~al.(2018)Zhou, Huang, Zhang, Zhu, and Liu]{zhou2018emotional}
Hao Zhou, Minlie Huang, Tianyang Zhang, Xiaoyan Zhu, and Bing Liu.
\newblock Emotional chatting machine: Emotional conversation generation with internal and external memory.
\newblock In \emph{Thirty-Second AAAI Conference on Artificial Intelligence}, 2018.

\bibitem[Zhou et~al.(2023{\natexlab{a}})Zhou, Lu, Mishra, Brahma, Basu, Luan, Zhou, and Hou]{zhou2023instruction}
Jeffrey Zhou, Tianjian Lu, Swaroop Mishra, Siddhartha Brahma, Sujoy Basu, Yi~Luan, Denny Zhou, and Le~Hou.
\newblock Instruction-following evaluation for large language models.
\newblock \emph{arXiv preprint arXiv:2311.07911}, 2023{\natexlab{a}}.

\bibitem[Zhou et~al.(2022{\natexlab{a}})Zhou, Blodgett, Trischler, Daum{\'e}~III, Suleman, and Olteanu]{zhou2022deconstructing}
Kaitlyn Zhou, Su~Lin Blodgett, Adam Trischler, Hal Daum{\'e}~III, Kaheer Suleman, and Alexandra Olteanu.
\newblock Deconstructing nlg evaluation: Evaluation practices, assumptions, and their implications.
\newblock In \emph{Proceedings of the 2022 Conference of the North American Chapter of the Association for Computational Linguistics: Human Language Technologies}, pp.\  314--324, 2022{\natexlab{a}}.

\bibitem[Zhou et~al.(2023{\natexlab{b}})Zhou, Jiang, Wilcox, Cotterell, and Sachan]{zhou2023controlled}
Wangchunshu Zhou, Yuchen~Eleanor Jiang, Ethan Wilcox, Ryan Cotterell, and Mrinmaya Sachan.
\newblock Controlled text generation with natural language instructions.
\newblock In \emph{International Conference on Machine Learning}, pp.\  42602--42613. PMLR, 2023{\natexlab{b}}.

\bibitem[Zhou et~al.(2022{\natexlab{b}})Zhou, Muresanu, Han, Paster, Pitis, Chan, and Ba]{zhou2022large}
Yongchao Zhou, Andrei~Ioan Muresanu, Ziwen Han, Keiran Paster, Silviu Pitis, Harris Chan, and Jimmy Ba.
\newblock Large language models are human-level prompt engineers.
\newblock In \emph{The Eleventh International Conference on Learning Representations}, 2022{\natexlab{b}}.

\bibitem[Zhu et~al.(2018)Zhu, Lu, Zheng, Guo, Zhang, Wang, and Yu]{zhu2018texygen}
Yaoming Zhu, Sidi Lu, Lei Zheng, Jiaxian Guo, Weinan Zhang, Jun Wang, and Yong Yu.
\newblock Texygen: A benchmarking platform for text generation models.
\newblock In \emph{The 41st International ACM SIGIR Conference on Research \& Development in Information Retrieval}, pp.\  1097--1100. ACM, 2018.

\bibitem[Ziegler et~al.(2019)Ziegler, Stiennon, Wu, Brown, Radford, Amodei, Christiano, and Irving]{ziegler2019fine}
Daniel~M Ziegler, Nisan Stiennon, Jeffrey Wu, Tom~B Brown, Alec Radford, Dario Amodei, Paul Christiano, and Geoffrey Irving.
\newblock Fine-tuning language models from human preferences.
\newblock \emph{arXiv preprint arXiv:1909.08593}, 2019.

\bibitem[Ziyu et~al.(2023)Ziyu, Qiguang, Longxuan, Mingda, Yi, Yushan, Haopeng, Weinan, and Liu]{ziyu2023through}
Zhuang Ziyu, Chen Qiguang, Ma~Longxuan, Li~Mingda, Han Yi, Qian Yushan, Bai Haopeng, Zhang Weinan, and Ting Liu.
\newblock Through the lens of core competency: Survey on evaluation of large language models.
\newblock In \emph{Proceedings of the 22nd Chinese National Conference on Computational Linguistics (Volume 2: Frontier Forum)}, pp.\  88--109, 2023.

\bibitem[Zou et~al.(2021)Zou, Yin, Zhong, Yang, Yang, and Tang]{zou2021controllable}
Xu~Zou, Da~Yin, Qingyang Zhong, Hongxia Yang, Zhilin Yang, and Jie Tang.
\newblock Controllable generation from pre-trained language models via inverse prompting.
\newblock In \emph{Proceedings of the 27th ACM SIGKDD Conference on Knowledge Discovery \& Data Mining}, pp.\  2450--2460, 2021.

\end{thebibliography}
\bibliographystyle{tmlr}

\end{document}